\documentclass[10pt]{article}
\newcommand{\arxivpath}{}
\IfFileExists{arxiv/arxiv.sty}{\renewcommand{\arxivpath}{arxiv/}}{}
\makeatletter
\edef\input@path{{\arxivpath}}
\makeatother
\usepackage{arxiv}
\usepackage{natbib}
\setcitestyle{authoryear,round,citesep={;},aysep={,},yysep={;}}
\date{}
\renewcommand{\headeright}{ }
\renewcommand{\shorttitle}{ }

\usepackage[utf8]{inputenc} 
\usepackage[T1]{fontenc}    
\usepackage{hyperref}       
\usepackage{url}            
\usepackage{booktabs}       
\usepackage{amsfonts}       
\usepackage{nicefrac}       
\usepackage{microtype}      
\usepackage{xcolor}         

\usepackage{amsmath}
\allowdisplaybreaks
\usepackage{amssymb}
\usepackage{mathtools}
\usepackage{amsthm}

\usepackage{graphicx}
\usepackage{subcaption}

\newcommand{\pr}[1]{ \left( #1 \right) }

\newcommand{\piold}[2]{\pi_{\mathrm{old}} (#1 \mid #2)}
\newcommand{\pik}{\pi_{\theta_k}}

\newcommand{\epshigh}{\varepsilon_{\mathrm{high}}}
\newcommand{\epslow}{\varepsilon_{\mathrm{low}}}

\DeclareMathOperator*{\argmax}{arg\,max}

\theoremstyle{plain}
\newtheorem{theorem}{Theorem}[section]

\newtheorem{lemma}[theorem]{Lemma}
\newtheorem{corollary}[theorem]{Corollary}
\theoremstyle{definition}

\newtheorem{assumption}[theorem]{Assumption}
\theoremstyle{remark}

\hypersetup{hidelinks}

\title{Clip-Low Increases Entropy and Clip-High Decreases Entropy in Reinforcement Learning of Large Language Models}

\author{%
\parbox{\dimexpr\textwidth-2\tabcolsep\relax}{\centering
\bfseries
Jaesung R. Park$^{1}$ \quad Junsu Kim$^{2}$ \quad Gyeongman Kim$^{3}$\\
Jinyoung Jo$^{4}$ \quad Sean Choi$^{5}$ \quad Jaewoong Cho$^{3}$ \quad Ernest K. Ryu$^{1}$\\[0.5em]
\normalfont
$^{1}$Department of Mathematics, UCLA\\
$^{2}$Department of Mathematical Sciences, Seoul National University\\
$^{3}$KRAFTON\\
$^{4}$Department of Linguistics, Stanford University\\
$^{5}$Department of Computer Science and Engineering, Santa Clara University
}}

\begin{document}

\maketitle

\begin{abstract}
Reinforcement learning with verifiable rewards (RLVR) is a leading approach for post-training large language models (LLMs) on reasoning tasks. However, it frequently suffers from entropy collapse, where the policy becomes nearly deterministic and exploration degrades. We show that this phenomenon is driven not only by the reward signal, but also by the clipping rule used in PPO- and GRPO-style objectives. In this work, we provide a theoretical explanation of entropy change by separating the effects of the two clipping components: clipping negative-advantage updates (clip-low) increases policy entropy, while clipping positive-advantage updates (clip-high) decreases it. We find that the sign of this bias is not universal for arbitrary policies, but follows from the heavy-head structure of pretrained LLM token distributions. We then validate our theory with extensive experiments, and further demonstrate that tuning clipping hyperparameters provides a lightweight entropy-control knob for counteracting entropy collapse and improving exploration.
\end{abstract}

\section{Introduction}

Reinforcement learning with verifiable rewards (RLVR) has recently emerged as the leading approach for enhancing the reasoning capabilities of large language models (LLMs), especially in the domain of mathematical reasoning \citep{lambert2024tulu, luong2024reft, guo2025deepseek, yang2025qwen3}. However, RLVR is prone to entropy collapse: a phenomenon where the LLM quickly converges to a near-deterministic form, hindering exploration and progress during prolonged RL training  \citep{Yu2025_dapo}.

Recent studies have reported this effect and continue to debate whether it is an inevitable byproduct of improved performance \citep{yue2025does, Cui2025_entropy, wu2025invisible}. A number of works have proposed empirical interventions to mitigate entropy collapse, such as tuning training hyperparameters \citep{Yu2025_dapo} or explicitly incorporating a KL-divergence loss term \citep{Liu2025_prorl}. 
However, these methods are largely heuristic, leaving open a basic mechanistic question: \emph{What parts of the RLVR objective systematically push entropy up or down, and why?} 
Answering this question is of central importance since controlling entropy directly affects the diversity of reasoning trajectories the LLM could produce.  

\paragraph{Contribution.}
In this paper, we show that the clipping mechanism used in PPO \citep{schulman2017proximal} or GRPO \citep{shao2024deepseekmath} 
induces a bias on the entropy dynamics. 
We first derive general first-order decompositions of entropy change for action-dependent advantages. To study the clipping effect in isolation, we then analyze the setting where rewards are \emph{random}, and theoretically prove that the lower clip (`clip-low') on negative advantages increases entropy, while the upper clip (`clip-high') on positive advantages decreases entropy under the stated conditions. 
We demonstrate that these effects crucially depend on the specific structural properties of LLM policies where most of the probability mass resides in the few `head' actions.

Next, we empirically demonstrate that the theoretical results extend to general RLVR settings for mathematical reasoning tasks. 
By simply tuning the clipping hyperparameters, we can effectively control the entropy dynamics during RLVR, 
thereby preventing entropy collapse, improving exploration (\texttt{pass@k}) throughout training.

\subsection{Related works} 
\label{subsec:relate}
\paragraph{Mitigating entropy collapse in RLVR.}
A growing line of work has studied the entropy collapse phenomenon. 
ProRL \citep{Liu2025_prorl} adopts clip-higher and further emphasizes the use of KL divergence loss for stabilizing entropy. 
Another popular approach is to use reward shaping to promote exploration 
\citep{cheng2025reasoning, gao2025navigate}, which could be understood largely as methods motivated by conventional reinforcement learning algorithms \citep{haarnoja2018soft, burda2019rnd}. 
Most closely related to our focus on clipping, DAPO \citep{Yu2025_dapo} empirically found that reducing the clip-high probability by setting higher $\epshigh$ values for the GRPO \citep{shao2024deepseekmath} algorithm mitigates entropy collapse. While this `clip-higher' method has been extensively studied and adopted in various works \citep{liu2025part, yue2025vapo, zhu2025proximal, qi2025defeating, qiu2025training, tang2025mind, chen2025nudging}, these works only tune $\epshigh$, leaving the effect of clip-low elusive. 
Our work complements these findings by giving a mechanistic account of why clipping induces directional bias on entropy, as well as elucidating the role of clip-low in exploration of LLMs.

\textbf{Theoretical analysis of clipping.}
Recent theory analyzes the clipped objective of PPO, including global optimality and hinge-loss interpretations of clipping (\citet{huang2021neural, huang2024ppo}), stationary-point convergence results (\citet{jin2023stationary}), and non-asymptotic global convergence guarantees (\citet{liu2025non}).
These works study clipping mainly through an optimization or convergence lens, whereas we characterize the entropy bias in a RL-LLM specific setup. 
We later show that this distinction is important as policy distribution of pretrained LLMs plays an integral role in elucidating the mechanism for entropy bias.

\paragraph{Exploration of LLMs during RLVR.} 
There are active debates about whether RLVR elicits genuinely novel reasoning or merely reweights reasoning paths already latent in the base model. 
\citet{Cui2025_entropy} conducted an extensive search and provided a different viewpoint that the decreasing entropy during training could actually be understood as a tradeoff with performance~\citep{deng2025decomposing, he2025rewarding, yue2025does, wu2025invisible}. 
On the other hand, conflicting evidence indicates that RLVR can induce capabilities not present in base models \citep{wen2025reinforcement, chen2025pass, song2025outcome, Liu2025_prorl}. 
Our findings strengthen this latter perspective: we show that deliberately maintaining higher entropy through controlled clipping could improve exploration, suggesting it is not an inherent limitation of RLVR.

\paragraph{Random reward for RL.}
Recent studies study whether RL could improve LLM benchmark scores even with weak, noisy, or entirely random rewards \citep{chen2025exploration, wang2025onetraining, lv2025climb, zhu2025surprising}. 
The work most closely related to ours is \citep{shao2025spurious}, where the authors train with purely random rewards and observe gains primarily for models in the \texttt{Qwen} family \citep{yang2025qwen3}. We show that entropy minimization is the consistent driver when training with random rewards, and that this mechanism appears across a broad set of model families. This reframes ``random-reward improvements'' as a predictable consequence of how the clipped RLVR objectives bias policies toward lower-entropy~\citep{zhao2025learning, agarwal2025unreasonable, gao2025one}.

\subsection{Notation and preliminaries} \label{subsec:prelim}

Consider the setup where given a prompt $x$, an LLM $\pi_\theta$ generates a response  $y = (y_1,\ldots,y_T)$ and a reward function $r(y)$ evaluates it. The objective is to maximize expected reward:
\begin{equation}
\begin{array}{ll}
\underset{\theta}{\mbox{maximize}}&
\displaystyle{
 \mathcal{J}(\theta) :=
\mathop{\mathbb{E}}_{\substack{x\sim \mathcal{D}\\ y\sim \pi_\theta(\cdot \mid x)}}[ r(y) ]
},
\end{array}
\label{eq:objective}
\end{equation}
where $\mathcal{D}$ denotes the training distribution of prompts.

We formulate this optimization problem into an RL problem. Specifically, consider the MDP with a discrete, finite state space $\mathcal{S}$ and a discrete, finite action space $\mathcal{A}$. The state is defined as $s_t = (x, y_1, \dots, y_{t-1})$ and action $a_t$ is the next token to generate, and the transition dynamics is deterministic where the generated token is appended to the state. 
We drop the $t$ subscript when there is no ambiguity. 
Finally, the language model $\pi_\theta$ is regarded as the policy, and we refer to this as the reinforcement learning of large language models (RL-LLM) setup. 

Given a policy $\pi$, we define its state visitation measure as
  \[
d^\pi(s) = \sum_{t=0}^\infty \mathbb{P}
\big(s_t = s \big)
= 
\mathbb{E}
\Big[\sum^T_{t=0}\mathbf{1}_{s_t = s}\Big],
\]
where the probability and expectation is with respect to $s_0=x\sim \mathcal{D}$ and $a_{t}\sim \pi(\cdot\,|\,s_t)$ for $t=0,1,\dots$.

\paragraph{REINFORCE.}
The classical REINFORCE policy gradient estimator \citep{williams1992simple} is given by
\begin{equation} 
\nabla_\theta \mathcal{J}(\theta) 
= \mathop{\mathbb{E}}_{\substack{x\sim \mathcal{D}\\ y\sim \pi_\theta(\cdot \mid x)}}\!\Bigg[
\sum_{t=1}^{T} \nabla_\theta \log \pi_\theta\!\big(y_t \,|\, y_{<t}, x\big)\, A_t
\Bigg],
\label{eq:pg}
\end{equation}
where $y_{<t} := (y_1, \dots, y_{t-1})$ and $A_t$ is an advantage estimate derived from the trajectory-level rewards, such as $A_t=r(y_{1:T})$.

Although it is possible to perform stochastic gradient descent (ascent) using the stochastic gradients from Equation~\eqref{eq:pg} (and doing so would avoid the clipping bias that we identify in this work), such an approach is typically less sample-efficient and less stable. Therefore, methods such as PPO \citep{schulman2017proximal} and GRPO \citep{shao2024deepseekmath} are preferred in the RL-LLM setting.

\paragraph{Group Relative Policy Optimization (GRPO).}
GRPO \citep{shao2024deepseekmath} is a variant of proximal policy optimization (PPO) \citep{schulman2017proximal} adapted for trajectory-level rewards.
The algorithm samples
a prompt $x\sim\mathcal{D}$ and $K$ responses $y^{(1)},\dots,y^{(K)}\sim \pi_{{\mathrm{old}}}(\cdot \mid x)$ from the current policy.
Then, the parameter update to $\theta$ is obtained by performing stochastic gradient steps to solve the subproblem
\[
\begin{array}{ll}
\underset{\theta}{\mbox{maximize}}&
\displaystyle{
\sum_{i=1}^K \frac{1}{T^{(i)}}
\sum^{T^{(i)}}_{t=1}
C_{\varepsilon} 
\pr{r_t^{(i)} (\theta), A_t^{(i)}}

}
\end{array}
\]
with
\begin{align*}
C_{\varepsilon} (r, A) &= 
\min\!\left(
r\,A,\;
\operatorname{clip}\!\big(r,\, 1-\varepsilon_{\mathrm{low}},\, 1+\varepsilon_{\mathrm{high}}\big)\,A
\right) \\
r_t^{(i)}(\theta) &=
\frac{\pi_\theta\big(y_t^{(i)} \,\big|\,y_{<t}^{(i)}, x\big)}
{\pi_{\mathrm{old}}\big(y_t^{(i)}\,\big|\, y_{<t}^{(i)}, x\big)}
\\
A_t^{(i)} &=
r (y^{(i)}) - \operatorname{mean}\big(r (y^{(1)}) ,\dots,r (y^{(K)}) \big)
\end{align*}
for $t=1,\dots,T^{(i)}$ and $i=1,\dots,K$.
The clipping mechanism, whose strength is controlled by the hyperparameters $\varepsilon_{\mathrm{low}} \in (0,1)$ and $\varepsilon_{\mathrm{high}} \in (0, \infty) $, originates from trust-region policy optimization (TRPO) \citep{schulman2015trust}. Its purpose is to prevent the optimization for the subproblem from deviating too far from the reference policy $\pi_{\theta_{\mathrm{old}}}$ that generated the responses. Concretely, the importance sampling ratio $r_t^{(i)}(\theta)$ is clipped to lie within the range $[1-\varepsilon_{\mathrm{low}}, 1+\varepsilon_{\mathrm{high}}]$ depending on the sign of $A_t^{(i)}$. The main thesis of this paper is that the two clipping mechanisms induce biases on entropy.

While the original GRPO formulation \citep{shao2024deepseekmath} normalizes $A_t^{(i)}$ by the standard deviation of the rewards, we follow the prescription of Dr.\ GRPO \citep{Liu2025_understanding} and omit this normalization. In addition, whereas the original PPO and GRPO employ a symmetric clipping parameter with $\varepsilon_{\mathrm{low}} = \varepsilon_{\mathrm{high}}$, DAPO~\citep{Yu2025_dapo} introduces asymmetric clipping with $\varepsilon_{\mathrm{low}}<\varepsilon_{\mathrm{high}}$.

\paragraph{Policy entropy.}
For any state $s \in \mathcal{S}$, the token-level (state-conditional) Shannon entropy of the policy $\pi_{\theta}$ is defined as
\begin{equation}
\mathcal{H}(\pi_{\theta} \,|\, s ) = - \sum_{a \in \mathcal{A}} \pi_{\theta}(a\,|\,s) \log \pi_{\theta}(a \,|\, s),
\end{equation}
where $\mathcal{A}$ (note $|\mathcal{A}| < \infty$) is the LLM vocabulary. 
In practice, we report the average token entropy over responses, evaluated over states encountered under the old policy distribution $\pi_{\mathrm{old}}$. 
For a minibatch of size $N$, we estimate the entropy with the following formula
\begin{equation}
\hat{\mathcal{H}}(\pi_\theta ) =  \frac{1}{N} \sum_{i=1}^{N} 
\Bigg[
  \frac{1}{T^{(i)}} \sum_{t=1}^{T^{(i)}} 
  \mathcal{H}(\pi_{\theta} \,|\, s_t^{(i)} ) 
\Bigg].
\end{equation}

\section{Theoretical analysis of clipping} \label{sec:rand_reward}

In this section, we theoretically analyze how the clipping mechanism affects policy entropy. We first derive general decompositions for action-dependent advantages, and then consider random rewards as a special case to study the clipping effect in isolation.

\subsection{Setup for the theoretical analysis}
\label{subsec:random_reward_formulation}
For simplicity, we assume that all responses have a common fixed length $T$. We omit the constant normalization $1/T$ from the GRPO objective, absorbing it into the learning rate:
\[
\mathcal{J}(\pi_{\theta};\pi_\mathrm{old})=
\!\!\!\!\!\!
\mathop{\mathbb{E}}_{\substack{x \sim \mathcal{D} \\ y \sim \pi_{\mathrm{old}}(\cdot\,|\, x) \\ A }}
\Bigg[
\sum^{T}_{t=1}
C_{\varepsilon} \pr{ \frac{\pi_{\theta} (y_t\,|\, y_{<t}, x)}{\pi_{\mathrm{old}}(y_t\,|\, y_{<t}, x)}, A}
\Bigg].
\]

We assume that $A$ is bounded almost surely: there exists some $A_{\max} > 0$ such that
\[
|A| \leq A_{\max}, \qquad \text{a.s.}
\]
In particular, a finite reward range and finite group size give bounded advantages in GRPO.

We parameterize the policy $\pik$ as a \emph{tabular softmax policy}:
\[
\pik (a \mid s) = \frac{\exp(\theta^k_{s,a})}{\sum_{a' \in \mathcal{A}} \exp(\theta^k_{s,a'})}
    \qquad\text{for } s\in \mathcal{S},\,a\in \mathcal{A}
\]
with state space $\mathcal{S}$,  finite action space $\mathcal{A}$, and trainable parameter $\theta^k \in \mathbb{R}^{|\mathcal{S}|\times |\mathcal{A}|}$. The $k$ sub- and superscripts denote the optimization step. 

The actual GRPO algorithm performs a limited number of optimization steps on the objective $\mathcal{J}$, typically using AdamW, which is difficult to model and analyze directly. 
For the sake of analytical tractability,
we consider two simplified formulations: the policy gradient and natural policy gradient algorithms applied to $\mathcal{J}$.
Namely, the first algorithm is the policy gradient algorithm 
\begin{equation}
\theta^{k+1}=\theta^k+\eta \nabla_\theta \mathcal{J}(\pi_{\theta_k};\pi_\mathrm{old}),
\label{eq:PG}
\end{equation}
\leavevmode{ where $\eta$ is the learning rate, and $\pi_\mathrm{old}$ is an older version of $\pi_{\theta_k}$ that is updated by the outer loop of GRPO. }

The second algorithm we consider is the \emph{natural} policy gradient algorithm  \citep{kakade2001natural} 
\begin{equation}
\pi_{\theta_{k+1}}\propto \pi_{\theta_{k}}\circ \exp\big(\eta \nabla_\pi \mathcal{J}(\pik;\pi_\mathrm{old})\big),
\label{eq:NPG}
\end{equation}
where again $\pi_\mathrm{old}$ is an older version of $\pik$ that is updated by the outer loop of GRPO and $\circ $ denotes element-wise multiplication. 
As we will see, our analysis of the two algorithms yields results that differ slightly but are qualitatively aligned. Since the two algorithms are considered models of the true GRPO update, this consistency lends further credibility to the qualitative conclusions drawn from our analysis.

Finally, we define the following \emph{clipping sets}
\begin{alignat*}    {3}
X_k(s)
= \Big\{ a \in \mathcal{A} : \frac{\pik ( a \mid s) }{ \piold{a}{s}}  < 1 - \epslow \Big\},  \quad 
Y_k(s)
= \Big\{ a \in \mathcal{A} : \frac{\pik ( a \mid s) }{ \piold{a}{s}} > 1 + \epshigh \Big\} 
\end{alignat*}
and \emph{clipping events} when sampling $a \sim \pik ( \cdot \mid s)$,
\begin{align*}
    \mathcal{X}_k (s) = \{ a \in X_k(s) \}, \quad
    \mathcal{Y}_k (s) = \{ a \in Y_k(s) \}.
\end{align*}

\subsection{First-order analysis of entropy change} \label{subsec:first_order_taylor}

We first present our analysis of the entropy change of the policy gradient algorithm. 
\begin{theorem}\label{thm:entropy_change_clip_general}
Consider the setup described in Section~\ref{subsec:random_reward_formulation} and the policy gradient algorithm given by Equation \eqref{eq:PG}.
Then, the change in entropy at state $s \in \mathcal{S}$ admits the first-order approximation
\begin{align*}
&\mathcal{H}(\pi_{\theta_{k+1}} \mid s)-\mathcal{H}(\pik \mid s)
= \eta\; d^{\pi_{\mathrm{old}}} 
\Big(
\underbrace{\mathbb{E}[A]\mathbb{E}[Q_k]-\mathbb{E}[AQ_k]}_{\text{unclipped reward contribution}}\\
&\quad+\underbrace{p_k\mathbb{E}\big[\max(-A,0)(\mathbb{E}[Q_k]-Q_k)\mid\mathcal{X}_k\big]}_{\text{clip-low contribution}} -\underbrace{q_k\mathbb{E}\big[\max(A,0)(\mathbb{E}[Q_k]-Q_k)\mid\mathcal{Y}_k\big]}_{\text{clip-high contribution}}
\Big)+\mathcal{O}(\eta^2),
\end{align*}
where $Q_k ( a \mid s) = \pik (a\,|\,s) (\log \pik (a\,|\,s) + \mathcal{H}(\pik \,|\,s))$, $p_k = \mathbb{P}(\mathcal{X}_k)$, $q_k = \mathbb{P}(\mathcal{Y}_k)$, $d^{\pi_{old}}$ is the state visitation measure, and the expectation $\mathbb{E}$ is taken with respect to $a \sim \pik(\cdot \mid s)$ and the fixed rollout's conditional distribution of $A$ given $s,a$. To clarify, all the terms on the right-hand side depend on $s$, and it would be more precise to write them as $d^{\pi_{old}}(s)$, $Q_k(\cdot \mid s)$, $\mathcal{X}_k(s)$, $\mathcal{Y}_k(s)$, $p_k(s)$, and $q_k(s)$. However, we suppress the dependence on $s$ for notational simplicity.
\end{theorem}

We defer the proof to Appendix~\ref{appendix:entropy_change_clip_proof}.

Theorem~\ref{thm:entropy_change_clip_general} decomposes the entropy change into the unclipped contribution and the contributions of clip-low and clip-high.
For fixed policies, decreasing $\varepsilon_{\mathrm{low}}$ expands the clip-low set and weakly increases $p_k=\mathbb{P}(\mathcal{X}_k)$; the same holds for $\varepsilon_{\mathrm{high}}$ and $q_k=\mathbb{P}(\mathcal{Y}_k)$ in the clip-high term.
If clip-low or clip-high is turned off, the corresponding probability $p_k$ or $q_k$ is zero, so that clipping contribution vanishes.
We use the convention $p_k\mathbb{E}[\cdot\mid\mathcal{X}_k]=0$ when $p_k=0$, and analogously for $q_k=0$.

Next, we present our analysis of the entropy change of the \emph{natural} policy gradient algorithm.

\begin{theorem}\label{thm:entropy_change_clip2_general}
Consider the setup described in Section~\ref{subsec:random_reward_formulation} and the natural policy gradient algorithm given by Equation~\eqref{eq:NPG}.
Then, the change in entropy at state $s \in \mathcal{S}$ admits the first-order approximation
\begin{align*}
&\mathcal{H}(\pi_{\theta_{k+1}} \mid s)-\mathcal{H}(\pik \mid s) =\eta\; d^{\pi_{\mathrm{old}}}\Big(
\underbrace{\mathbb{E}\big[A(-\log\pik-\mathcal{H}(\pik\mid s))\big]}_{\text{unclipped reward contribution}}\\
&\quad\!\!\!+\underbrace{p_k\mathbb{E}\big[\max(-A,0)(-\log\pik-\mathcal{H}(\pik\mid s))\mid\mathcal{X}_k\big]}_{\text{clip-low contribution}} 
-\underbrace{q_k\mathbb{E}\big[\max(A,0)(-\log\pik-\mathcal{H}(\pik\mid s))\mid\mathcal{Y}_k\big]}_{\text{clip-high contribution}}
\Big)+\mathcal{O}(\eta^2),
\end{align*}
where $p_k = \mathbb{P}(\mathcal{X}_k)$, $q_k = \mathbb{P}(\mathcal{Y}_k)$, $d^{\pi_{old}}$ is the state visitation measure, and the expectation $\mathbb{E}$ is taken with respect to $a \sim \pik(\cdot \mid s)$ and the fixed rollout's conditional distribution of $A$ given $s,a$. Here $\log\pik$ denotes $\log\pik(a\mid s)$.
To clarify, all the terms on the right-hand side depend on $s$, and it would be more precise to write them as $d^{\pi_{old}}(s)$, $\mathcal{X}_k(s)$, $\mathcal{Y}_k(s)$, $p_k(s)$, and $q_k(s)$. However, we suppress the dependence on $s$ for notational simplicity.
\end{theorem}
We defer the proof to Appendix~\ref{appendix:proof2}.

Importantly, 
if the following conditions
\begin{equation}\label{eq:Q-sign}
\begin{aligned}
p_k\mathbb{E}\big[\max(-A,0)(\mathbb{E}[Q_k]-Q_k)\mid\mathcal{X}_k\big]\geq 0, \quad 
q_k\mathbb{E}\big[\max(A,0)(\mathbb{E}[Q_k]-Q_k)\mid\mathcal{Y}_k\big]\geq 0
\end{aligned}
\end{equation}
for policy gradient, and
\begin{equation}\label{eq:log-sign}
\begin{aligned}
p_k\mathbb{E}\big[\max(-A,0)(-\log\pik-\mathcal{H}(\pik\mid s))\mid\mathcal{X}_k\big]\geq 0, \quad
q_k\mathbb{E}\big[\max(A,0)(-\log\pik-\mathcal{H}(\pik\mid s))\mid\mathcal{Y}_k\big]\geq 0
\end{aligned}
\end{equation}
hold for natural policy gradient, then the claim that the clip-low term increases entropy and clip-high term decreases entropy is substantiated.

However, Inequalities~\eqref{eq:Q-sign} and \eqref{eq:log-sign} are not guaranteed to hold universally, requiring further analysis of the statistical properties of clipping events. 
Theoretically understanding these inequalities requires careful analysis of the clipping events, which are closely related to the properties of policy distributions $\pi_{\mathrm{old}}(a \mid s)$ and $\pik (a \mid s)$. 
In Section~\ref{subsec:clip_characterize}, we construct a stylized statistical framework that enables this analysis, elucidating the fact that the specific structure of the pretrained LLM policy plays a crucial role in determining the direction of the given inequalities.

\paragraph{Remark (Random rewards).}
Following \cite{shao2025spurious}, consider random rewards for which $A$ is independent of the prompt and response and symmetric about zero. Let $\mu$ and $\nu$ satisfy
\[
\mathbb{E}[A]=0,\qquad
\mathbb{P}(A>0)=\mathbb{P}(A<0)=\nu,\qquad
\mathbb{E}[A\mid A>0]=\mu.
\]
Then $\mathbb{E}[\max(A,0)]=\mathbb{E}[\max(-A,0)]=\mu\nu$, so the unclipped contribution vanishes and the two decompositions reduce to the following corollaries.

\begin{corollary}\label{thm:entropy_change_clip}
Consider the setup in Theorem~\ref{thm:entropy_change_clip_general} and the random-reward conditions above. Then,
\begin{align*}
  &\mathcal{H}(\pi_{\theta_{k+1}} | s) \! - \! \mathcal{H}(\pi_{\theta_k} | s) \! =
    \! \mu \nu \eta \; d^{\pi_{\mathrm{old}}} \! \big ( 
    \underbrace{p_k ( \mathbb{E}[Q_k]-\mathbb{E}[Q_k \mid \mathcal{X}_k] ) }_{\text{clip-low contribution}}
    -
     \underbrace{
     q_k (\mathbb{E}[Q_k]-\mathbb{E} [Q_k \mid \mathcal{Y}_k])}_{\text{clip-high contribution}} \! \big)
     \! + \! \mathcal{O}(\eta^2)
\end{align*}
\end{corollary}

\begin{corollary}\label{thm:entropy_change_clip2}
Consider the setup in Theorem~\ref{thm:entropy_change_clip2_general} and the random-reward conditions above. Then,
\begin{align*}    
  \mathcal{H}(\pi_{\theta_{k+1}} \mid s)  -   \mathcal{H}(\pik \mid s)
   &=
    \mu \nu \eta \; d^{\pi_{\mathrm{old}}} \big ( 
    \underbrace{p_k (\mathbb{E}[-\log \pik \mid \mathcal{X}_k]-\mathcal{H}(\pik \mid s) ) }_{\text{clip-low contribution}}  \\*
     & \phantom{=} - 
     \underbrace{
       q_k (\mathbb{E}[-\log \pik \mid \mathcal{Y}_k]-\mathcal{H}(\pik \mid s) ) 
     }_{\text{clip-high contribution}}
\big)
+ \mathcal{O}(\eta^2),
\end{align*}
\end{corollary}
The derivations are provided in Appendices~\ref{appendix:entropy_change_clip_proof} and~\ref{appendix:proof2}.

Under the random-reward conditions, independence of $A$ from the action implies that the left-hand sides of Inequalities~\eqref{eq:Q-sign} and~\eqref{eq:log-sign} reduce to $\mu\nu$ times the corresponding unweighted expressions in Corollaries~\ref{thm:entropy_change_clip} and~\ref{thm:entropy_change_clip2}. Thus, the same inequalities establish the opposing effects of clip-low and clip-high as in Theorems~\ref{thm:entropy_change_clip_general} and~\ref{thm:entropy_change_clip2_general}. Since the unclipped contribution vanishes, these corollaries isolate the entropy bias induced by clipping even when the rewards contain no learning signal.

\subsection{Statistical characterization of clipping} \label{subsec:clip_characterize}

In this section, we analyze the inequalities~\eqref{eq:Q-sign} and \eqref{eq:log-sign}. 
Note that the given inequalities depend on probabilistic events $X_k(s)$ and $Y_k(s)$, which in turn depend on $r_k(a \mid s) = \frac{ \pik (a \mid s)}{\piold{a}{s}}$. 
For simplicity, we assume that $\pik ( a \mid s)$ is obtained by performing \emph{one-step} of policy gradient or natural policy gradient with a size-$1$ minibatch. 
(size-$1$ minibatch assumption could be substituted with size-$N$ minibatch, assuming that there is no state revisit) Note that this is different from the setting in Section~\ref{subsec:first_order_taylor}, where we considered full-batch gradient updates.

As stated earlier, the inequalities~\eqref{eq:Q-sign} and \eqref{eq:log-sign} do not hold for all probability distributions $\piold{a}{s}$. 
Indeed, if we perform RL training on a base model with all weights initialized (for which the policy is close to a uniform distribution) with random rewards, the clipping mechanism does not act as a bias on the entropy, and entropy stays roughly constant throughout RL training; see Appendix~\ref{appendix:subsec:random_init_model}. 

Recall that for LLMs, the action space $\mathcal{A}$ is the whole token vocabulary, which is very large compared to the number of tokens with non-negligible probabilities given a certain context. In other words, the conditional probability distribution of the LLM policy are highly non-uniform.
Moreover, prior works observed that LLM-generated text mostly consists of low-entropy tokens~\citep{wang2025beyond, qian2025demystifying}. 
Motivated by these observations, we impose the following assumption on the conditional probability distribution $\piold{a}{s}$.

\begin{assumption} [Heavy-head assumption]
    For each state $s \in \mathcal{S}$ visited by the LLM policy, there exists a dominant ``head" action $a^*$ such that $\piold{a^*}{s} \geq 1- \delta_0,$
    where $\delta_0 \in [0,1]$ depends on $\eta$, $A_{\max}$, $\epslow$, $\epshigh$,  and $|\mathcal{A}|$.  
\end{assumption}

Table~\ref{tab:heavy_head_table} displays the proportion of tokens satisfying the heavy-head assumption out of the whole response. We experimentally tested for various $\delta_0$ values, throughout the RL training of \texttt{Qwen2.5-3B-Instruct} model with the \texttt{GSM8K} dataset. 
Further validation of the heavy-head assumption on larger models and various datasets are provided in Section~\ref{appendix:subsec:heavy_head_validation}.

\begin{table}[hbtp]
    \centering
    \caption{ Proportion of visited states that satisfy the heavy head assumption $\max_{a \in \mathcal{A}} \pi_{\theta} (a \mid s) \geq 1 - \delta_0$ throughout RL training.}
    \begin{tabular}{ccccc}
    \toprule
       Train steps & $\delta_0 = 0.3$ & $\delta_0 = 0.1$ & $\delta_0 = 0.01$ \\
    \midrule
        0  & 0.88 & 0.78  & 0.63 \\
        120  & 0.95 & 0.89 & 0.80 \\
        240  & 0.96 & 0.91 & 0.84 \\
        360  & 0.97 & 0.92 & 0.86 \\
    \bottomrule
    \end{tabular}
    \label{tab:heavy_head_table}
\end{table}

The following theorems show that inequalities~\eqref{eq:Q-sign} and \eqref{eq:log-sign} hold with high probability, given small enough $\delta_0$.

\begin{theorem} \label{thm:cond_uncond_Q_inequality}
    Consider the setup described in Section~\ref{subsec:random_reward_formulation}, and fix a state $s \in \mathcal{S}$ and a reference distribution $\piold {\cdot}{s}$. 
    Assume a single stochastic policy-gradient step of Equation~\eqref{eq:PG} using one sample $\hat{a} \sim \piold{\cdot} {s}$, producing $\pik (\cdot \mid s)$. 
    Assume that the heavy-head assumption holds with
    \[
    \delta_0 \! = \!  \min \Big\{
    \frac{\epshigh}{1 \! + \! \epshigh}, 
    \frac{\epslow}{1 \! + \! 2\eta A_{\max}\epslow}, \frac{u_0}{1 \! + \! 2 \eta A_{\max}u_0} 
    \Big\}
    \]
    
    where $u_0 = \frac{\log 2}{4 \log (e(|\mathcal{A}|-1))}$.
    Then, on the event $\{ \hat{a} = a^*\} $ (which happens with probability at least $1- \delta_0$), 
    \begin{align*}
    p_k\mathbb{E}\big[\max(-A,0)(\mathbb{E}[Q_k]-Q_k)\mid\mathcal{X}_k\big]\geq 0, \quad 
    q_k\mathbb{E}\big[\max(A,0)(\mathbb{E}[Q_k]-Q_k)\mid\mathcal{Y}_k\big]\geq 0
    \end{align*}
    holds, where $Q_k ( a \mid s) = \pik (a \mid s) (\log \pik (a \mid s) + \mathcal{H}(\pik \mid s))$, $p_k = \mathbb{P}(\mathcal{X}_k)$, $q_k = \mathbb{P}(\mathcal{Y}_k)$,
    and expectation $\mathbb{E}$ is taken with respect to $a \sim \pik (\cdot \mid s)$. For terms involving $A$, the expectation also includes the fixed rollout's conditional distribution of $A$ given $s,a$. 
\end{theorem}

The proof is provided in Appendix~\ref{appendix:proof3}.

Along with Theorem~\ref{thm:entropy_change_clip_general}, we conclude that (with high probability) clip-high decreases entropy and clip-low increases entropy relative to the unclipped update.
We also prove the sign for inequalities~\eqref{eq:log-sign}, deriving analogous results for natural policy gradient updates.

\begin{theorem} \label{thm:cond_uncond_inequality}
Consider the setup described in Section~\ref{subsec:random_reward_formulation}, and fix a state $s \in \mathcal{S}$ and a reference distribution $\piold {\cdot}{s}$. 
    Assume a single stochastic natural policy gradient step of Equation~\eqref{eq:NPG} using one sample $\hat{a} \sim \piold{\cdot} {s}$, producing $\pik(\cdot \mid s)$. 
    Assume that the heavy-head assumption holds with
    \[
    \delta_0 = \min\left\{ 
    \frac{\varepsilon_{\mathrm{high}}}{1+\varepsilon_{\mathrm{high}}},
    u_0e^{-2\eta A_{\max}},
    \frac{\varepsilon_{\mathrm{low}}}{1-\varepsilon_{\mathrm{low}}}\,e^{-2\eta A_{\max}}
    \right\}
    \]
    where $u_0$ is defined in Theorem~\ref{thm:cond_uncond_Q_inequality}.
    Then, on the event $\{ \hat{a} = a^*\} $ (which happens with probability at least $1- \delta_0$), 
    \begin{align*}
    p_k\mathbb{E}\big[\max(-A,0)(-\log\pik-\mathcal{H}(\pik\mid s))\mid\mathcal{X}_k\big]\geq 0, \quad
    q_k\mathbb{E}\big[\max(A,0)(-\log\pik-\mathcal{H}(\pik\mid s))\mid\mathcal{Y}_k\big]\geq 0
    \end{align*}
    holds, where $p_k = \mathbb{P}(\mathcal{X}_k)$, $q_k = \mathbb{P}(\mathcal{Y}_k)$, and the expectation $\mathbb{E}$ is taken with respect to $a \sim \pik (\cdot \mid s)$. For terms involving $A$, the expectation also includes the fixed rollout's conditional distribution of $A$ given $s,a$. 
\end{theorem}

The proof is provided in appendix~\ref{appendix:proof4}.

\section{Experimental results}

In this section, we perform extensive experiments on the RL training with both random and nonrandom (true) rewards. 
First, in Section~\ref{subsec:validation}, we verify our theoretical results for the random reward setting. 
Next,
in Section~\ref{subsec:experiments}, we examine the clipping effects in the general (true reward) RLVR setting.
By controlling $\epshigh$ and $\epslow$, we verify the effect of the clipping mechanism on the entropy dynamics and reasoning capability of large language models during RL training. Full implementation details are in Appendix~\ref{appendix:experimental_settings}.

\begin{figure*}[hbtp]
  \centering
  \begin{subfigure}[t]{0.48\textwidth}
    \centering
    \includegraphics[width=\linewidth]{\arxivpath 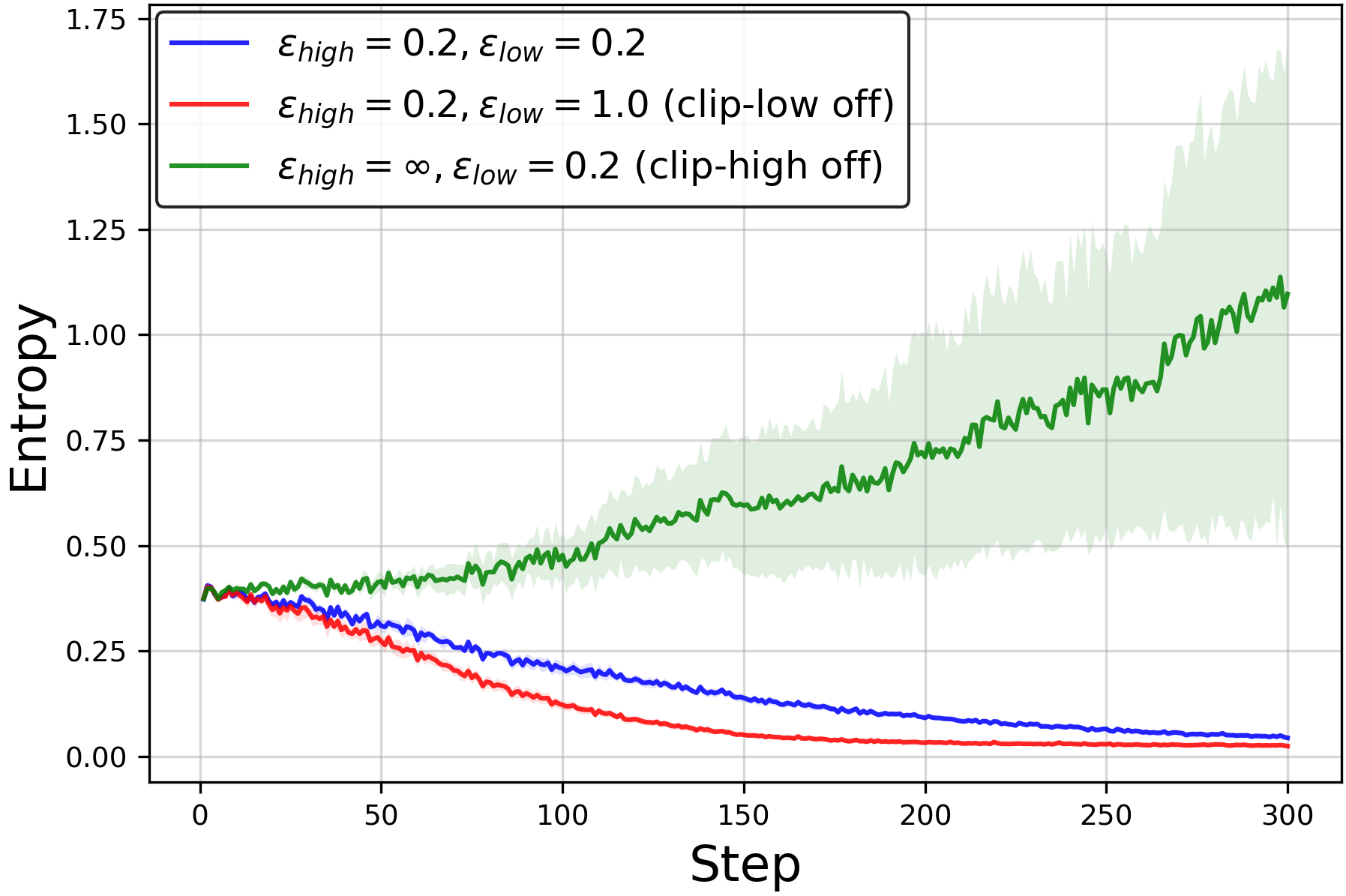}
  \end{subfigure}\hfill
  \begin{subfigure}[t]{0.48\textwidth}
    \centering
    \includegraphics[width=\linewidth]{\arxivpath 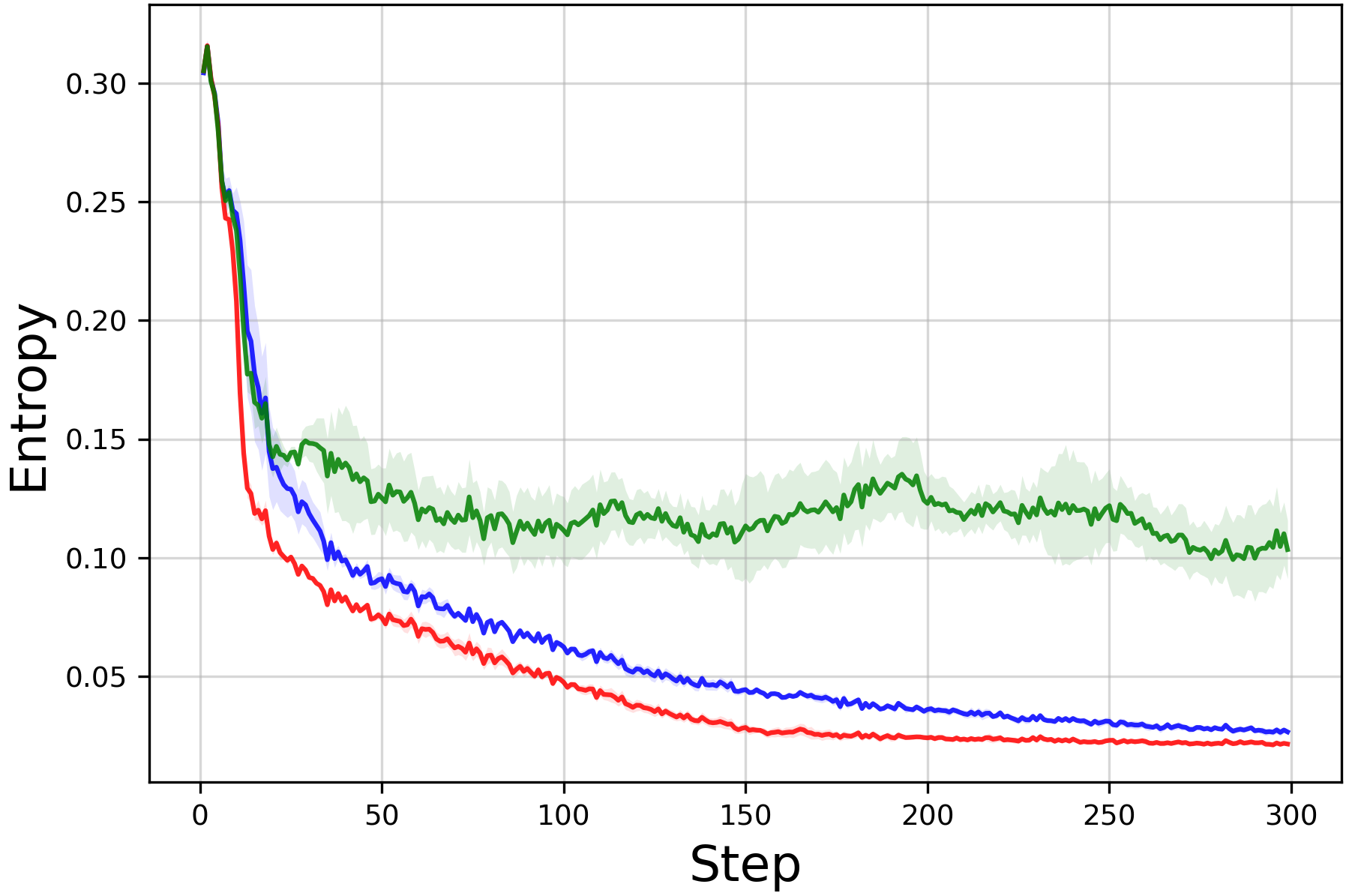}
  \end{subfigure}
  \caption{Change of policy entropy during RL training  with \textbf{(Left)} random rewards on the \texttt{Qwen2.5-1.5B-Instruct} model and \textbf{(Right)} nonrandom rewards on the \texttt{Qwen2.5-3B-Instruct} model with different clipping settings. We observe that both clip-high and clip-low influence the entropy, consistent with our theoretical predictions. We average over three runs with different random seeds; shades denote pointwise uncertainty bands across seeds.
  }
  \label{fig:entropy_ablation}
\end{figure*}

\subsection{Empirical validation of theory}
\label{subsec:validation}

\paragraph{Clip-high decreases entropy and clip-low increases entropy for random reward RL training.}
We find that the experimental results are consistent with our theoretical predictions. The left panel of Figure~\ref{fig:entropy_ablation} shows that, when RL trained with random rewards, decreasing/increasing $\varepsilon_{\mathrm{low}}$ (making clip-low stronger/weaker) increases/decreases entropy, and decreasing/increasing $\varepsilon_{\mathrm{high}}$ (making clip-high stronger/weaker) decreases/increases entropy.
Moreover, we find that with symmetric clipping parameters ($\varepsilon_{\mathrm{low}}=\varepsilon_{\mathrm{high}}=0.2$), the effect of clip-high dominates that of clip-low, leading to a reduction in entropy. 
Note that this result also holds with KL divergence term being present in the loss function; refer to Appendix~\ref{appendix:kl_divergence}.

\paragraph{Noisy and spurious rewards reduce entropy.} 
Prior work has investigated whether RLVR can enhance LLM reasoning even in the presence of noisy rewards~\citep{wang2025onetraining, lv2025climb} or random (spurious) rewards~\citep{shao2025spurious}. In particular, \citet{shao2025spurious} find that GRPO-based training with clipping yields clear improvements for \texttt{Qwen}-based models, but little to no benefit for \texttt{Llama}- or \texttt{Olmo}-based models. By contrast, Figure~\ref{fig:random_reward_model_ablation} shows that training with random rewards consistently reduces policy entropy across \texttt{Qwen}, \texttt{Llama}, and \texttt{Olmo}. This pattern suggests that the primary effect of spurious reward training may be entropy minimization, which in turn influences reasoning behavior as recently suggested in \citep{agarwal2025unreasonable, gao2025one}. For further details on ablation studies on random reward training, refer to Appendix~\ref{appendix:random_reward_ablation}.

\paragraph{Clipping directions under Transformer--AdamW updates.}
To directly test the clipping mechanism under the practical optimizer, we conduct a matched $2\times2$ intervention with \texttt{Qwen2.5-3B} and true GRPO advantages. After a common first AdamW update, we clone the model parameters, AdamW moments, scheduler state, training batch, minibatch order, and RNG state. We change only the clipping configuration in the second update: neither clipping mechanism, clip-low only, clip-high only, or both. All four updated models are evaluated on the same fixed rollout prefixes.

In Table~\ref{tab:adamw_intervention}, we compare these effects with the clip-low and clip-high contributions in Theorems~\ref{thm:entropy_change_clip_general} and~\ref{thm:entropy_change_clip2_general}, computed from the stored GRPO advantages and full-vocabulary policy probabilities. For AdamW, each entry measures the entropy difference caused by enabling one clipping mechanism, with the other mechanism disabled or enabled. The PG/NPG contributions and the Transformer--AdamW intervention show the same directional result: clip-low increases entropy, while clip-high decreases it. In the AdamW intervention, these directions persist whether the other clipping mechanism is enabled or disabled. Thus, the opposing effects of clip-low and clip-high extend to shared Transformer parameters and accumulated AdamW state. Further experimental details are provided in Appendix~\ref{appendix:adamw_intervention}.

\begin{table}[t]
\centering
\small
\setlength{\tabcolsep}{7pt}
\renewcommand{\arraystretch}{1.15}
\caption{Clipping effects under PG, NPG, and Transformer--AdamW. All values are in units of $10^{-3}$. PG/NPG entries are learning-rate-free first-order coefficients; AdamW entries are post-update entropy differences evaluated on the same fixed rollout prefixes.}
\label{tab:adamw_intervention}
\begin{tabular}{lcc}
\toprule
\textbf{Update} & \textbf{Clip-low effect} & \textbf{Clip-high effect} \\
\midrule
Regular PG & $+0.0725$ & $-0.1650$ \\
NPG & $+0.6170$ & $-1.0591$ \\
AdamW, other clipping mechanism disabled & $+0.5409$ & $-1.5727$ \\
AdamW, other clipping mechanism enabled & $+0.6360$ & $-1.4776$ \\
\bottomrule
\end{tabular}
\end{table}

\subsection{Experiments: Math reasoning tasks} \label{subsec:experiments}

\paragraph{Clip-high decreases entropy and clip-low increases entropy for general RLVR training.} 

Next, we conduct experiments with true nonrandom rewards for mathematical reasoning tasks. 
We begin with an ablation study of the clipping mechanisms. Specifically, we disable the clip-low mechanism (by setting $\varepsilon_{\mathrm{low}}=1.0$) and the clip-high mechanism (by setting $\varepsilon_{\mathrm{high}}=\infty$). As shown in the right panel of Figure~\ref{fig:entropy_ablation}, removing clip-high increases entropy, while removing clip-low decreases it, in qualitative agreement with the clipping contributions characterized in Section~\ref{sec:rand_reward}.

It is worth noting that unlike the random reward setting, RLVR training with true rewards has an entropy-reduction effect, which can be attributed to RLVR's suppression of incorrect reasoning paths. For example, while the configuration $\varepsilon_{\mathrm{high}} = \infty$ (clip-high off) and $\varepsilon_{\mathrm{low}} = 0.2$ increased entropy in the random reward setting (Figure~\ref{fig:entropy_ablation}, left), the same configuration leads to reduced entropy in the true reward RLVR setup (Figure~\ref{fig:entropy_ablation}, right).

\begin{table*}[t]
\centering
\small
\setlength{\tabcolsep}{6pt}
\renewcommand{\arraystretch}{1.15}
\caption{Clipping ablation results on math benchmarks. For the \texttt{Qwen2.5-7B-Instruct} model, we perform RL training with clip-high off ($\epshigh=\infty$) and clip-low off ($\epslow=1.0$), with the symmetric clipping baseline ($\epshigh=\epslow=0.2$). Reported numbers are \texttt{pass@32} (\%) and \texttt{mean@32} (\%) scores. Within each run, we average the common checkpoints at steps 190, 200, and 210, and report the mean and sample standard deviation across three independent matched training runs. Bold indicates the best result per metric.}
\label{tab:clip_ablations_math_benchmarks}
\begin{tabular}{lccccc}
\toprule
\textbf{Qwen2.5-7B-Instruct} & \textbf{AIME24} & \textbf{AIME25} & \textbf{MATH500} & \textbf{AMC23} & \textbf{Avg.} \\
\midrule
\multicolumn{6}{l}{\texttt{pass@32} (\%)}\\
$\epshigh=0.2,\epslow=0.2$ & $26.3\pm4.4$ & $22.8\pm4.8$ & $\mathbf{88.3\pm1.2}$ & $76.4\pm1.0$ & $53.5\pm2.7$ \\
\quad + $\epslow=1.0$ & $24.9\pm7.0$ & $20.6\pm2.3$ & $80.0\pm14.4$ & $75.2\pm2.8$ & $50.2\pm4.7$ \\
\quad + $\epshigh=\infty$ & $\mathbf{38.2\pm2.5}$ & $\mathbf{29.1\pm4.6}$ & $85.7\pm11.3$ & $\mathbf{81.2\pm1.7}$ & $\mathbf{58.5\pm2.6}$ \\
\midrule
\multicolumn{6}{l}{\texttt{mean@32} (\%)}\\
$\epshigh=0.2,\epslow=0.2$ & $14.3\pm1.3$ & $7.7\pm1.8$ & $\mathbf{75.6\pm2.0}$ & $54.2\pm0.7$ & $38.0\pm1.0$ \\
\quad + $\epslow=1.0$ & $12.6\pm1.4$ & $6.7\pm1.8$ & $69.3\pm12.5$ & $53.7\pm7.6$ & $35.6\pm1.8$ \\
\quad + $\epshigh=\infty$ & $\mathbf{14.6\pm1.5}$ & $\mathbf{11.0\pm0.2}$ & $71.8\pm11.0$ & $\mathbf{55.6\pm1.2}$ & $\mathbf{38.2\pm3.1}$ \\
\bottomrule
\end{tabular}
\end{table*}

\paragraph{Entropy control leads to improved exploration.}
While RLVR enhances the reasoning performance of LLMs, prior work \citep{yue2025does, song2025outcome} has shown that it also narrows the range of reasoning trajectories the model can explore, also referred to as the \emph{reasoning boundary}. 
This was often measured by the decline of \texttt{pass@k} metric throughout RL training. 

Our empirical results suggest that the clipping mechanism could be a central factor to understanding this phenomenon.
In Table~\ref{tab:clip_ablations_math_benchmarks}, we report the result of clipping ablation experiments on various math benchmarks. 
Overall, we observe that the clip-high off setting $(\epshigh = \infty)$ yields the highest average \texttt{pass@32} score, with a comparable average \texttt{mean@32} score to the symmetric clipping baseline, while the clip-low off setting $(\epslow = 1.0)$ has the lowest average scores. 
This result suggests that clip-low also plays a central role in the exploration. Refer to Appendix~\ref{appendix:dapo_experiments} for the full experimental results throughout training.

To further probe the effect of entropy control via clipping, we conducted a hyperparameter search on $\epshigh$ and $\epslow$. 
We train the \texttt{Qwen2.5-3B-Instruct} model with 15 different configurations, where $\epshigh \in \{0.2, 0.25, 0.3 \} $ and $\epslow \in \{ 0.1, 0.15, 0.2, 0.25, 0.3\}$. 
For each of these configurations, we measured the entropy, \texttt{mean@32}, and \texttt{pass@32} score on the \texttt{GSM8K} test dataset. 
The result is summarized in the left panel of Figure~\ref{fig:epsilon_grid} as a $5 \times 3$ grid, with full details in Appendix~\ref{appendix:clip_param_search}. 
The results show that simply controlling $\epshigh$ and $\epslow$ offers a means to continuously adjust the policy entropy. 


In the right panel of Figure~\ref{fig:epsilon_grid}, we quantify the correlation between entropy and \texttt{pass@32} score, and between entropy and \texttt{mean@32} score, 
using the four complementary correlation measures
:
Pearson's $r$ (linear), Spearman's $\rho$ and Kendall's $\tau$ (rank-based monotone), and distance correlation (dCor; general dependence)~\citep{szekely2007measuring}. 
We observe consistent strong association between entropy and the \texttt{pass@32} across all measures, while \texttt{mean@32} and entropy do not exhibit clear monotone correlation.
Together, these findings suggest that appropriate clipping choices can prevent entropy collapse even without explicit entropy regularization or a KL penalty, and that the resulting entropy changes reflect genuine exploration behavior.

\begin{figure*}[hbtp]
  \centering
  \begin{subfigure}[t]{0.48\textwidth}
    \centering
    \includegraphics[width=\linewidth]{\arxivpath 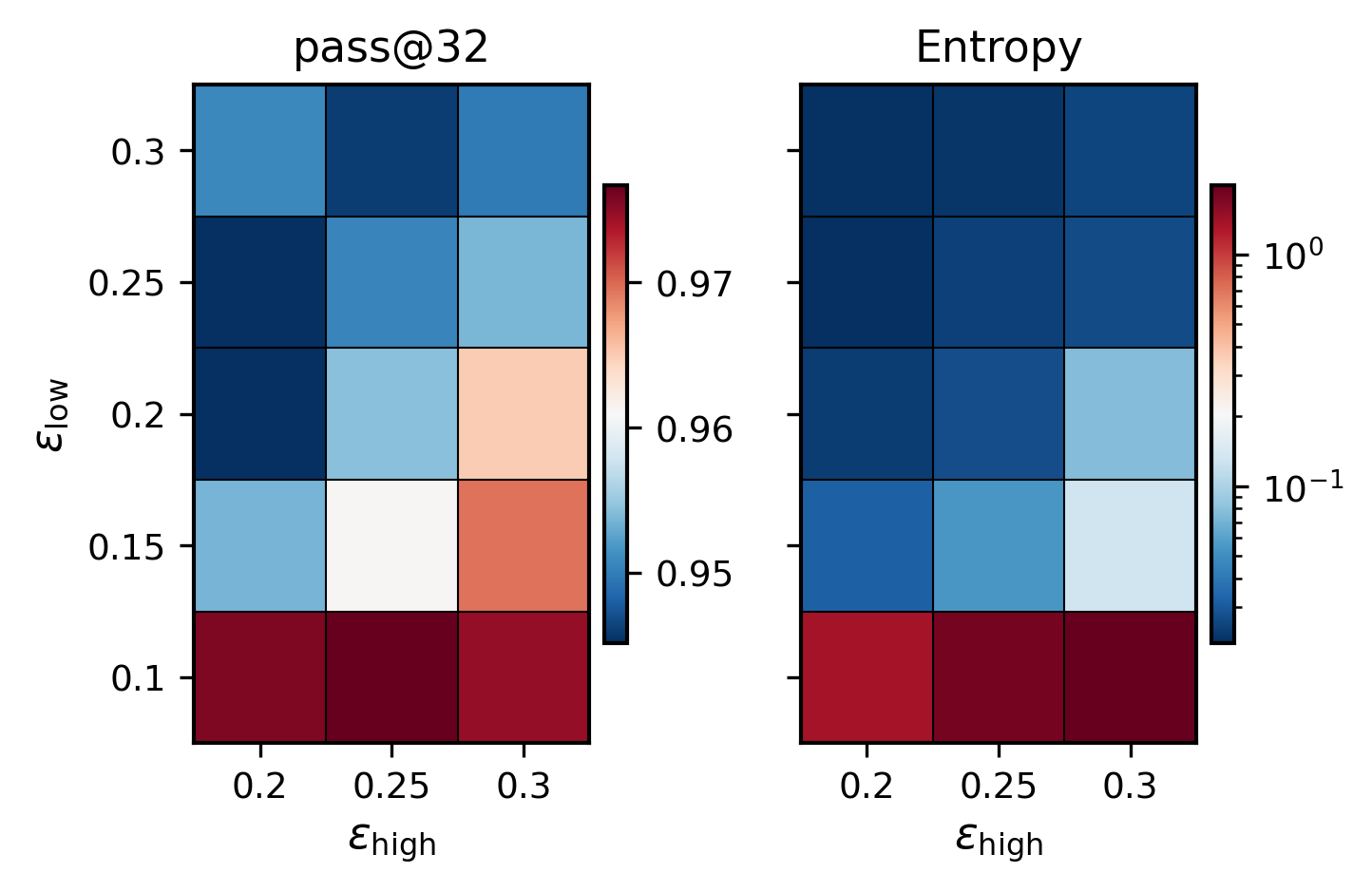}
  \end{subfigure}\hfill
  \begin{subfigure}[t]{0.48\textwidth}
    \centering
    \includegraphics[width=\linewidth]{\arxivpath 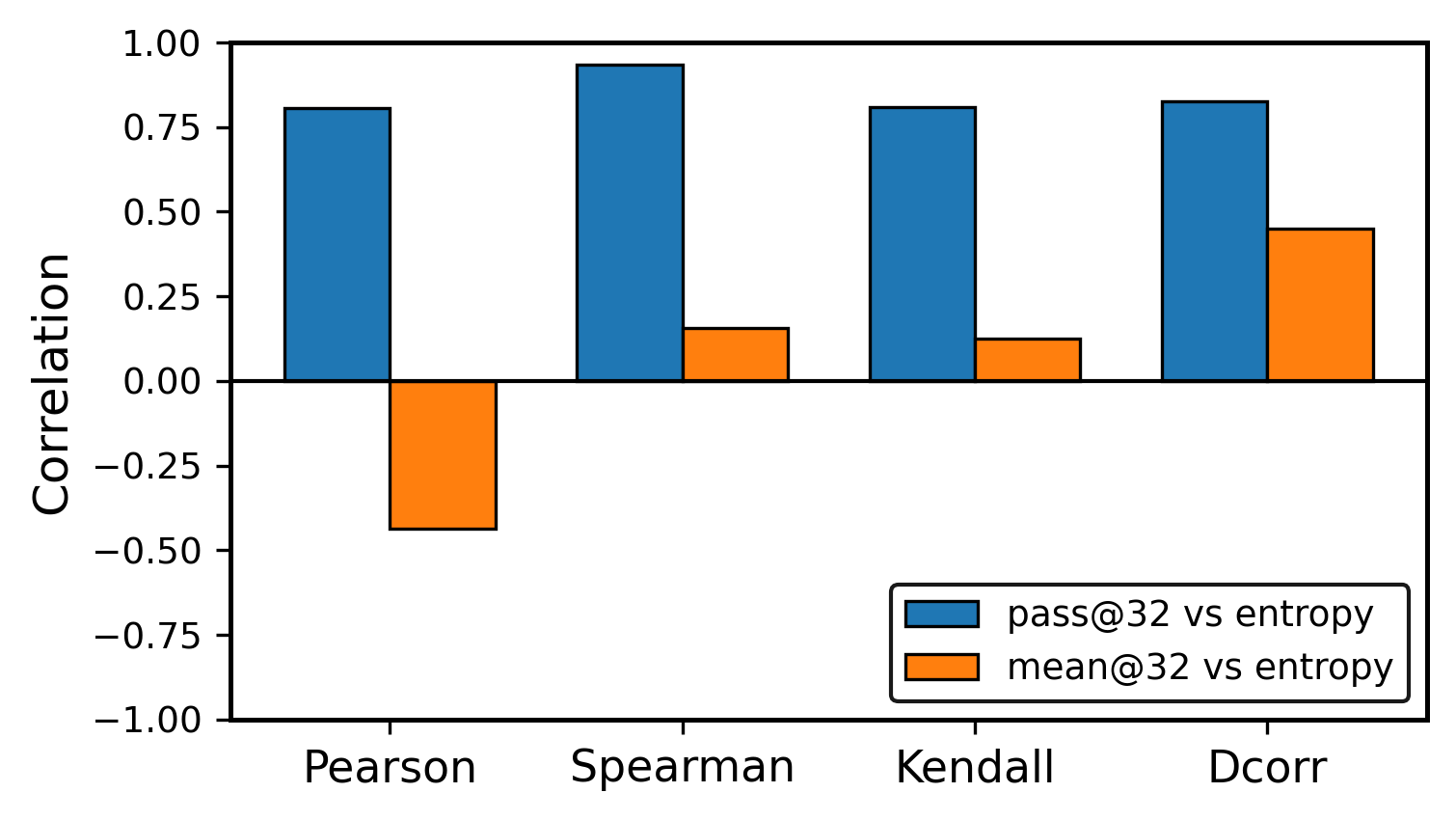}
  \end{subfigure}
  \caption{ \textbf{(Left)} Averaged \texttt{pass@32} score and entropy for various $\epshigh$ and $\epslow$ settings. We color entropy in log scale for visibility.
  \textbf{(Right)} Correlation between entropy and \texttt{pass@32} compared with correlation between entropy and \texttt{mean@32}.
  }
  \label{fig:epsilon_grid}
\end{figure*}

\section{Conclusion} \label{sec:conclusion}

In this work, we theoretically analyzed the clipping bias mechanism on entropy, highlighting an overlooked confounding factor in RLVR. Furthermore, we demonstrated that the entropy can be controlled by appropriately setting the clip-low and clip-high values.
Our work opens up exciting new research directions, including optimizing  $(\epshigh, \epslow)$ settings, or developing an adaptive scheme. 
One limitation of our work is that the formal results analyze stylized model setup such as first-order approximations and  tabular softmax policies, which motivates a future direction of expanding the theory by relaxing the assumptions.

\bibliography{\arxivpath neurips_paper}

\begin{thebibliography}{55}
\providecommand{\natexlab}[1]{#1}
\providecommand{\url}[1]{\texttt{#1}}
\expandafter\ifx\csname urlstyle\endcsname\relax
  \providecommand{\doi}[1]{doi: #1}\else
  \providecommand{\doi}{doi: \begingroup \urlstyle{rm}\Url}\fi

\bibitem[Agarwal et~al.(2025)Agarwal, Zhang, Yuan, Han, and
  Peng]{agarwal2025unreasonable}
Shivam Agarwal, Zimin Zhang, Lifan Yuan, Jiawei Han, and Hao Peng.
\newblock The unreasonable effectiveness of entropy minimization in {LLM}
  reasoning.
\newblock \emph{Neural Information Processing Systems}, 2025.

\bibitem[AI-MO()]{AIMO_validation_AMC}
AI-MO.
\newblock {AI-MO} validation {AMC} ({American Mathematics Competitions})
  dataset.
\newblock Dataset on Hugging Face.
\newblock URL \url{https://huggingface.co/datasets/AI-MO/aimo-validation-amc}.

\bibitem[Burda et~al.(2019)Burda, Edwards, Storkey, and Klimov]{burda2019rnd}
Yuri Burda, Harrison Edwards, Amos Storkey, and Oleg Klimov.
\newblock Exploration by random network distillation.
\newblock In \emph{International Conference on Learning Representations}, 2019.

\bibitem[Chen et~al.(2025{\natexlab{a}})Chen, Peng, Choubey, Huang, Zhang,
  Bansal, and Wu]{chen2025nudging}
Justin Chih-Yao Chen, Becky~Xiangyu Peng, Prafulla~Kumar Choubey, Kung-Hsiang
  Huang, Jiaxin Zhang, Mohit Bansal, and Chien-Sheng Wu.
\newblock Nudging the boundaries of llm reasoning.
\newblock \emph{arXiv:2509.25666}, 2025{\natexlab{a}}.

\bibitem[Chen et~al.(2026)Chen, Li, Li, Yin, Chen, and
  Lin]{chen2025exploration}
Peter Chen, Xiaopeng Li, Ziniu Li, Wotao Yin, Xi~Chen, and Tianyi Lin.
\newblock Exploration vs exploitation: Rethinking rlvr through clipping,
  entropy, and spurious reward.
\newblock \emph{International Conference on Learning Representations}, 2026.

\bibitem[Chen et~al.(2025{\natexlab{b}})Chen, Qin, Wu, Ling, Ye, Zhao, and
  Shi]{chen2025pass}
Zhipeng Chen, Xiaobo Qin, Youbin Wu, Yue Ling, Qinghao Ye, Wayne~Xin Zhao, and
  Guang Shi.
\newblock Pass@k training for adaptively balancing exploration and exploitation
  of large reasoning models.
\newblock \emph{arXiv:2508.10751}, 2025{\natexlab{b}}.

\bibitem[Cheng et~al.(2025)Cheng, Huang, Zhu, Dai, Zhao, Zhang, and
  Wei]{cheng2025reasoning}
Daixuan Cheng, Shaohan Huang, Xuekai Zhu, Bo~Dai, Wayne~Xin Zhao, Zhenliang
  Zhang, and Furu Wei.
\newblock Reasoning with exploration: An entropy perspective.
\newblock \emph{arXiv:2506.14758}, 2025.

\bibitem[Cobbe et~al.(2021)Cobbe, Kosaraju, Bavarian, Chen, Jun, Kaiser,
  Plappert, Tworek, Hilton, Nakano, et~al.]{cobbe2021training}
Karl Cobbe, Vineet Kosaraju, Mohammad Bavarian, Mark Chen, Heewoo Jun, Lukasz
  Kaiser, Matthias Plappert, Jerry Tworek, Jacob Hilton, Reiichiro Nakano,
  et~al.
\newblock Training verifiers to solve math word problems.
\newblock \emph{arXiv:2110.14168}, 2021.

\bibitem[Cui et~al.(2025)Cui, Zhang, Chen, Yuan, Wang, Zuo, Li, Fan, Chen,
  Chen, Liu, Peng, Bai, Ouyang, Cheng, Zhou, and Ding]{Cui2025_entropy}
Ganqu Cui, Yuchen Zhang, Jiacheng Chen, Lifan Yuan, Zhi Wang, Yuxin Zuo,
  Haozhan Li, Yuchen Fan, Huayu Chen, Weize Chen, Zhiyuan Liu, Hao Peng, Lei
  Bai, Wanli Ouyang, Yu~Cheng, Bowen Zhou, and Ning Ding.
\newblock The entropy mechanism of reinforcement learning for reasoning
  language models.
\newblock \emph{arXiv:2505.22617}, 2025.

\bibitem[Deng et~al.(2025)Deng, Chen, Chen, Zhao, and Wen]{deng2025decomposing}
Jia Deng, Jie Chen, Zhipeng Chen, Wayne~Xin Zhao, and Ji{-}Rong Wen.
\newblock Decomposing the entropy{-}performance exchange: The missing keys to
  unlocking effective reinforcement learning.
\newblock \emph{arXiv:2508.02260}, 2025.

\bibitem[Gao et~al.(2025{\natexlab{a}})Gao, Pan, Wang, Zhong, Lu, Cai, Jiang,
  and Zhao]{gao2025navigate}
Jingtong Gao, Ling Pan, Yejing Wang, Rui Zhong, Chi Lu, Qingpeng Cai, Peng
  Jiang, and Xiangyu Zhao.
\newblock Navigate the unknown: Enhancing {LLM} reasoning with intrinsic
  motivation guided exploration.
\newblock \emph{arXiv:2505.17621}, 2025{\natexlab{a}}.

\bibitem[Gao et~al.(2025{\natexlab{b}})Gao, Chen, Luo, Zhou, and
  Dai]{gao2025one}
Zitian Gao, Lynx Chen, Haoming Luo, Joey Zhou, and Bryan Dai.
\newblock One{-}shot entropy minimization.
\newblock \emph{arXiv:2505.20282}, 2025{\natexlab{b}}.

\bibitem[Grattafiori et~al.(2024)Grattafiori, Dubey, Jauhri, Pandey, Kadian,
  Al-Dahle, Letman, Mathur, Schelten, Vaughan, et~al.]{grattafiori2024llama}
Aaron Grattafiori, Abhimanyu Dubey, Abhinav Jauhri, Abhinav Pandey, Abhishek
  Kadian, Ahmad Al-Dahle, Aiesha Letman, Akhil Mathur, Alan Schelten, Alex
  Vaughan, et~al.
\newblock The llama 3 herd of models.
\newblock \emph{arXiv preprint arXiv:2407.21783}, 2024.

\bibitem[Guo et~al.(2025)Guo, Yang, Zhang, Song, Wang, Zhu, Xu, Zhang, Ma, Bi,
  et~al.]{guo2025deepseek}
Daya Guo, Dejian Yang, Haowei Zhang, Junxiao Song, Peiyi Wang, Qihao Zhu,
  Runxin Xu, Ruoyu Zhang, Shirong Ma, Xiao Bi, et~al.
\newblock Deepseek-r1 incentivizes reasoning in llms through reinforcement
  learning.
\newblock \emph{Nature}, 645\penalty0 (8081):\penalty0 633--638, 2025.

\bibitem[Haarnoja et~al.(2018)Haarnoja, Zhou, Abbeel, and
  Levine]{haarnoja2018soft}
Tuomas Haarnoja, Aurick Zhou, Pieter Abbeel, and Sergey Levine.
\newblock Soft actor{-}critic: Off{-}policy maximum entropy deep reinforcement
  learning with a stochastic actor.
\newblock \emph{International Conference on Machine Learning}, 2018.

\bibitem[He et~al.(2025)He, Fried, and Welleck]{he2025rewarding}
Andre He, Daniel Fried, and Sean Welleck.
\newblock Rewarding the unlikely: Lifting grpo beyond distribution sharpening.
\newblock \emph{arXiv:2506.02355}, 2025.

\bibitem[Hendrycks et~al.(2021)Hendrycks, Burns, Kadavath, Arora, Basart, Tang,
  Song, and Steinhardt]{hendrycks2021math}
Dan Hendrycks, Collin Burns, Saurav Kadavath, Akul Arora, Steven Basart, Eric
  Tang, Dawn Song, and Jacob Steinhardt.
\newblock Measuring mathematical problem solving with the {MATH} dataset.
\newblock \emph{Neural Information Processing Systems Track on Datasets and
  Benchmarks}, 2021.

\bibitem[Huang et~al.(2021)Huang, Hsieh, Ho, Yao, Hu, Ouyang, Wu,
  et~al.]{huang2021neural}
Nai-Chieh Huang, Ping-Chun Hsieh, Kuo-Hao Ho, Hsuan-Yu Yao, Kai-Chun Hu,
  Liang-Chun Ouyang, I~Wu, et~al.
\newblock Neural ppo-clip attains global optimality: A hinge loss perspective.
\newblock \emph{arXiv:2110.13799}, 2021.

\bibitem[Huang et~al.(2024)Huang, Hsieh, Ho, and Wu]{huang2024ppo}
Nai-Chieh Huang, Ping-Chun Hsieh, Kuo-Hao Ho, and I-Chen Wu.
\newblock Ppo-clip attains global optimality: Towards deeper understandings of
  clipping.
\newblock \emph{AAAI Conference on Artificial Intelligence}, 2024.

\bibitem[HuggingFace(2025)]{math_verify_github}
HuggingFace.
\newblock Math-verify.
\newblock GitHub repository, 2025.
\newblock URL \url{https://github.com/huggingface/Math-Verify}.

\bibitem[HuggingFaceH4()]{AIME_2024_HF}
HuggingFaceH4.
\newblock Aime 2024 dataset.
\newblock Dataset on Hugging Face.
\newblock URL \url{https://huggingface.co/datasets/HuggingFaceH4/aime_2024}.

\bibitem[Jin et~al.(2023)Jin, Li, and Wang]{jin2023stationary}
Ruinan Jin, Shuai Li, and Baoxiang Wang.
\newblock On stationary point convergence of ppo-clip.
\newblock \emph{International Conference on Learning Representations}, 2023.

\bibitem[Kakade(2001)]{kakade2001natural}
Sham~M. Kakade.
\newblock A natural policy gradient.
\newblock \emph{Advances in Neural Information Processing Systems}, 2001.

\bibitem[Lambert et~al.(2024)Lambert, Morrison, Pyatkin, Huang, Ivison,
  Brahman, Miranda, Liu, Dziri, Lyu, et~al.]{lambert2024tulu}
Nathan Lambert, Jacob Morrison, Valentina Pyatkin, Shengyi Huang, Hamish
  Ivison, Faeze Brahman, Lester James~V Miranda, Alisa Liu, Nouha Dziri, Shane
  Lyu, et~al.
\newblock Tulu 3: Pushing frontiers in open language model post{-}training.
\newblock \emph{arXiv:2411.15124}, 2024.

\bibitem[Liu(2025)]{liu2025_entropy_iteration}
Jiacai Liu.
\newblock How does {RL} policy entropy converge during iteration?
\newblock Zhihu Zhuanlan, 2025.
\newblock URL \url{https://zhuanlan.zhihu.com/p/28476703733}.

\bibitem[Liu et~al.(2025{\natexlab{a}})Liu, Diao, Lu, Hu, Dong, Choi, Kautz,
  and Dong]{Liu2025_prorl}
Mingjie Liu, Shizhe Diao, Ximing Lu, Jian Hu, Xin Dong, Yejin Choi, Jan Kautz,
  and Yi~Dong.
\newblock {ProRL}: Prolonged reinforcement learning expands reasoning
  boundaries in large language models.
\newblock \emph{Neural Information Processing Systems}, 2025{\natexlab{a}}.

\bibitem[Liu et~al.(2025{\natexlab{b}})Liu, Dai, Zhang, and Wen]{liu2025non}
Yin Liu, Qiming Dai, Junyu Zhang, and Zaiwen Wen.
\newblock Non-asymptotic global convergence of ppo-clip.
\newblock \emph{arXiv:2512.16565}, 2025{\natexlab{b}}.

\bibitem[Liu et~al.(2025{\natexlab{c}})Liu, Chen, Li, Qi, Pang, Du, Lee, and
  Lin]{Liu2025_understanding}
Zichen Liu, Changyu Chen, Wenjun Li, Penghui Qi, Tianyu Pang, Chao Du, Wee~Sun
  Lee, and Min Lin.
\newblock Understanding {R1-Zero}-like training: A critical perspective.
\newblock \emph{Conference on Language Modeling}, 2025{\natexlab{c}}.

\bibitem[Liu et~al.(2025{\natexlab{d}})Liu, Liu, He, Wang, Liu, Pan, Hu, Xiong,
  Huang, Hu, et~al.]{liu2025part}
Zihe Liu, Jiashun Liu, Yancheng He, Weixun Wang, Jiaheng Liu, Ling Pan, Xinyu
  Hu, Shaopan Xiong, Ju~Huang, Jian Hu, et~al.
\newblock Part i: Tricks or traps? a deep dive into rl for llm reasoning.
\newblock \emph{arXiv:2508.08221}, 2025{\natexlab{d}}.

\bibitem[Luong et~al.(2024)Luong, Zhang, Jie, Sun, Jin, and Li]{luong2024reft}
Trung~Quoc Luong, Xinbo Zhang, Zhanming Jie, Peng Sun, Xiaoran Jin, and Hang
  Li.
\newblock {ReFT}: Reasoning with reinforced fine-tuning.
\newblock \emph{Association for Computational Linguistics}, 2024.

\bibitem[Lv et~al.(2025)Lv, Xie, Sun, Kang, and Yan]{lv2025climb}
Ang Lv, Ruobing Xie, Xingwu Sun, Zhanhui Kang, and Rui Yan.
\newblock The climb carves wisdom deeper than the summit: On the noisy rewards
  in learning to reason.
\newblock \emph{arXiv:2505.22653}, 2025.

\bibitem[Qi et~al.(2025)Qi, Liu, Zhou, Pang, Du, Lee, and Lin]{qi2025defeating}
Penghui Qi, Zichen Liu, Xiangxin Zhou, Tianyu Pang, Chao Du, Wee~Sun Lee, and
  Min Lin.
\newblock Defeating the training-inference mismatch via fp16.
\newblock \emph{arXiv:2510.26788}, 2025.

\bibitem[Qian et~al.(2025)Qian, Liu, Wen, Bai, Liu, and
  Shao]{qian2025demystifying}
Chen Qian, Dongrui Liu, Haochen Wen, Zhen Bai, Yong Liu, and Jing Shao.
\newblock Demystifying reasoning dynamics with mutual information: Thinking
  tokens are information peaks in llm reasoning.
\newblock \emph{arXiv preprint arXiv:2506.02867}, 2025.

\bibitem[Qiu et~al.(2025)Qiu, Ting, Dong, Chen, and Huang]{qiu2025training}
Yue Qiu, Yujan Ting, Pei Dong, Terrence Chen, and Weijing Huang.
\newblock Training medical qa models based on mixed rewards from
  multiple-choice and open-ended questions.
\newblock \emph{Empirical Methods in Natural Language Processing}, 2025.

\bibitem[Schulman et~al.(2015)Schulman, Levine, Abbeel, Jordan, and
  Moritz]{schulman2015trust}
John Schulman, Sergey Levine, Pieter Abbeel, Michael Jordan, and Philipp
  Moritz.
\newblock Trust region policy optimization.
\newblock \emph{International Conference on Machine Learning}, 2015.

\bibitem[Schulman et~al.(2017)Schulman, Wolski, Dhariwal, Radford, and
  Klimov]{schulman2017proximal}
John Schulman, Filip Wolski, Prafulla Dhariwal, Alec Radford, and Oleg Klimov.
\newblock Proximal policy optimization algorithms.
\newblock \emph{arXiv:1707.06347}, 2017.

\bibitem[Shao et~al.(2025)Shao, Li, Xin, Geng, Wang, Oh, Du, Lambert, Min,
  Krishna, Tsvetkov, Hajishirzi, Koh, and Zettlemoyer]{shao2025spurious}
Rulin Shao, Shuyue~Stella Li, Rui Xin, Scott Geng, Yiping Wang, Sewoong Oh,
  Simon~Shaolei Du, Nathan Lambert, Sewon Min, Ranjay Krishna, Yulia Tsvetkov,
  Hannaneh Hajishirzi, Pang~Wei Koh, and Luke Zettlemoyer.
\newblock Spurious rewards: Rethinking training signals in {RLVR}.
\newblock \emph{arXiv:2506.10947}, 2025.

\bibitem[Shao et~al.(2024)Shao, Wang, Zhu, Xu, Song, Bi, Zhang, Zhang, Li,
  et~al.]{shao2024deepseekmath}
Zhihong Shao, Peiyi Wang, Qihao Zhu, Runxin Xu, Junxiao Song, Xiao Bi, Haowei
  Zhang, Mingchuan Zhang, Y.~K. Li, et~al.
\newblock {DeepSeekMath}: Pushing the limits of mathematical reasoning in open
  language models.
\newblock \emph{arXiv:2402.03300}, 2024.

\bibitem[Sheng et~al.(2025)Sheng, Zhang, Ye, Wu, Zhang, Zhang, Peng, Lin, and
  Wu]{sheng2025hybridflow}
Guangming Sheng, Chi Zhang, Zilingfeng Ye, Xibin Wu, Wang Zhang, Ru~Zhang,
  Yanghua Peng, Haibin Lin, and Chuan Wu.
\newblock {HybridFlow}: A flexible and efficient {RLHF} framework.
\newblock \emph{European Conference on Computer Systems}, 2025.

\bibitem[Song et~al.(2025)Song, Kempe, and Munos]{song2025outcome}
Yuda Song, Julia Kempe, and Remi Munos.
\newblock Outcome-based exploration for {LLM} reasoning.
\newblock \emph{arXiv:2509.06941}, 2025.

\bibitem[Sz{\'e}kely et~al.(2007)Sz{\'e}kely, Rizzo, and
  Bakirov]{szekely2007measuring}
GJ~Sz{\'e}kely, ML~Rizzo, and NK~Bakirov.
\newblock Measuring and testing dependence by correlation of distances.
\newblock \emph{Annals of Statistics}, 35\penalty0 (6):\penalty0 2769--2794,
  2007.

\bibitem[Tang et~al.(2025)Tang, Xie, Sun, Huang, Luo, Yang, Jin, and
  Wang]{tang2025mind}
Peijun Tang, Shangjin Xie, Binyan Sun, Baifu Huang, Kuncheng Luo, Haotian Yang,
  Weiqi Jin, and Jianan Wang.
\newblock Mind to hand: Purposeful robotic control via embodied reasoning.
\newblock \emph{arXiv:2512.08580}, 2025.

\bibitem[Wang et~al.(2025{\natexlab{a}})Wang, Yu, Gao, Zheng, Liu, Lu, Dang,
  Chen, Yang, Zhang, et~al.]{wang2025beyond}
Shenzhi Wang, Le~Yu, Chang Gao, Chujie Zheng, Shixuan Liu, Rui Lu, Kai Dang,
  Xionghui Chen, Jianxin Yang, Zhenru Zhang, et~al.
\newblock High-entropy minority tokens drive effective reinforcement learning
  for {LLM} reasoning.
\newblock \emph{Neural Information Processing Systems}, 2025{\natexlab{a}}.

\bibitem[Wang et~al.(2025{\natexlab{b}})Wang, Yang, Zeng, Ren, Liu, Peng,
  Cheng, He, Wang, Gao, Chen, Wang, Du, and Shen]{wang2025onetraining}
Yiping Wang, Qing Yang, Zhiyuan Zeng, Liliang Ren, Liyuan Liu, Baolin Peng, Hao
  Cheng, Xuehai He, Kuan Wang, Jianfeng Gao, Weizhu Chen, Shuohang Wang,
  Simon~Shaolei Du, and Yelong Shen.
\newblock Reinforcement learning for reasoning in large language models with
  one training example.
\newblock \emph{Neural Information Processing Systems}, 2025{\natexlab{b}}.

\bibitem[Wen et~al.(2025)Wen, Liu, Zheng, Xu, Ye, Wu, Liang, Wang, Li, Miao,
  et~al.]{wen2025reinforcement}
Xumeng Wen, Zihan Liu, Shun Zheng, Zhijian Xu, Shengyu Ye, Zhirong Wu, Xiao
  Liang, Yang Wang, Junjie Li, Ziming Miao, et~al.
\newblock Reinforcement learning with verifiable rewards implicitly
  incentivizes correct reasoning in base {LLMs}.
\newblock \emph{arXiv:2506.14245}, 2025.

\bibitem[Williams(1992)]{williams1992simple}
Ronald~J. Williams.
\newblock Simple statistical gradient{-}following algorithms for connectionist
  reinforcement learning.
\newblock \emph{Machine Learning}, 8\penalty0 (3):\penalty0 229--256, 1992.

\bibitem[Wu et~al.(2025)Wu, Xuan, Lu, Harchaoui, and Choi]{wu2025invisible}
Fang Wu, Weihao Xuan, Ximing Lu, Zaid Harchaoui, and Yejin Choi.
\newblock The invisible leash: Why {RLVR} may not escape its origin.
\newblock \emph{arXiv:2507.14843}, 2025.

\bibitem[Yang et~al.(2024)Yang, Yang, Zhang, Hui, Zheng, Yu, Li, Liu, Huang,
  Dong, Wei, Lin, Yang, Tu, Zhang, Yang, Yang, Zhou, Lin, Dang, Lu, Bao, Yang,
  Yu, Li, Xue, Zhang, Zhu, Men, Lin, Li, Xia, Ren, Ren, Fan, Su, Zhang, Wan,
  Liu, Cui, Zhang, Qiu, Quan, and Wang]{qwen2024_techreport}
An~Yang, Baosong Yang, Beichen Zhang, Binyuan Hui, Bo~Zheng, Bowen Yu,
  Chengyuan Li, Dayiheng Liu, Fei Huang, Guanting Dong, Haoran Wei, Huan Lin,
  Jian Yang, Jianhong Tu, Jianwei Zhang, Jianxin Yang, Jiaxin Yang, Jingren
  Zhou, Junyang Lin, Kai Dang, Keming Lu, Keqin Bao, Kexin Yang, Le~Yu, Mei Li,
  Mingfeng Xue, Pei Zhang, Qin Zhu, Rui Men, Runji Lin, Tianhao Li, Tingyu Xia,
  Xingzhang Ren, Xuancheng Ren, Yang Fan, Yang Su, Yi-Chao Zhang, Yunyang Wan,
  Yuqi Liu, Zeyu Cui, Zhenru Zhang, Zihan Qiu, Shanghaoran Quan, and Zekun
  Wang.
\newblock Qwen2.5 technical report.
\newblock \emph{arXiv:2412.15115}, 2024.

\bibitem[Yang et~al.(2025)Yang, Li, Yang, Zhang, Hui, Zheng, Yu, Gao, Huang,
  Lv, et~al.]{yang2025qwen3}
An~Yang, Anfeng Li, Baosong Yang, Beichen Zhang, Binyuan Hui, Bo~Zheng, Bowen
  Yu, Chang Gao, Chengen Huang, Chenxu Lv, et~al.
\newblock Qwen3 technical report.
\newblock \emph{arXiv:2505.09388}, 2025.

\bibitem[Yu et~al.(2025)Yu, Zhang, Zhu, Yuan, Zuo, Yue, Dai, Fan, Liu, Liu,
  Liu, Lin, Lin, Ma, Sheng, Tong, Zhang, Zhang, Zhang, Zhu, Zhu, Chen, Chen,
  Wang, Yu, Song, Wei, Zhou, Liu, Ma, Zhang, Yan, Qiao, Wu, and
  Wang]{Yu2025_dapo}
Qiying Yu, Zheng Zhang, Ruofei Zhu, Yufeng Yuan, Xiaochen Zuo, Yu~Yue, Weinan
  Dai, Tiantian Fan, Gaohong Liu, Lingjun Liu, Xin Liu, Haibin Lin, Zhiqi Lin,
  Bole Ma, Guangming Sheng, Yuxuan Tong, Chi Zhang, Mofan Zhang, Wang Zhang,
  Hang Zhu, Jinhua Zhu, Jiaze Chen, Jiangjie Chen, Chengyi Wang, Hongli Yu,
  Yuxuan Song, Xiangpeng Wei, Hao Zhou, Jingjing Liu, Wei-Ying Ma, Ya-Qin
  Zhang, Lin Yan, Mu~Qiao, Yonghui Wu, and Mingxuan Wang.
\newblock {DAPO}: An open-source {LLM} reinforcement learning system at scale.
\newblock \emph{Neural Information Processing Systems}, 2025.

\bibitem[Yue et~al.(2025{\natexlab{a}})Yue, Chen, Lu, Zhao, Wang, Song, and
  Huang]{yue2025does}
Yang Yue, Zhiqi Chen, Rui Lu, Andrew Zhao, Zhaokai Wang, Shiji Song, and Gao
  Huang.
\newblock Does reinforcement learning really incentivize reasoning capacity in
  {LLMs} beyond the base model?
\newblock \emph{Neural Information Processing Systems}, 2025{\natexlab{a}}.

\bibitem[Yue et~al.(2025{\natexlab{b}})Yue, Yuan, Yu, Zuo, Zhu, Xu, Chen, Wang,
  Fan, Du, et~al.]{yue2025vapo}
Yu~Yue, Yufeng Yuan, Qiying Yu, Xiaochen Zuo, Ruofei Zhu, Wenyuan Xu, Jiaze
  Chen, Chengyi Wang, TianTian Fan, Zhengyin Du, et~al.
\newblock Vapo: Efficient and reliable reinforcement learning for advanced
  reasoning tasks.
\newblock \emph{arXiv:2504.05118}, 2025{\natexlab{b}}.

\bibitem[Zhao et~al.(2025)Zhao, Kang, Feng, Levine, and Song]{zhao2025learning}
Xuandong Zhao, Zhewei Kang, Aosong Feng, Sergey Levine, and Dawn Song.
\newblock Learning to reason without external rewards.
\newblock \emph{arXiv:2505.19590}, 2025.

\bibitem[Zhu et~al.(2025{\natexlab{a}})Zhu, Xie, Wang, Sun, Wang, and
  Liu]{zhu2025proximal}
Wenhong Zhu, Ruobing Xie, Rui Wang, Xingwu Sun, Di~Wang, and Pengfei Liu.
\newblock Proximal supervised fine-tuning.
\newblock \emph{arXiv:2508.17784}, 2025{\natexlab{a}}.

\bibitem[Zhu et~al.(2025{\natexlab{b}})Zhu, Xia, Wei, Chen, Chen, and
  Meng]{zhu2025surprising}
Xinyu Zhu, Mengzhou Xia, Zhepei Wei, Wei{-}Lin Chen, Danqi Chen, and Yu~Meng.
\newblock The surprising effectiveness of negative reinforcement in {LLM}
  reasoning.
\newblock \emph{Neural Information Processing Systems}, 2025{\natexlab{b}}.

\end{thebibliography}
\bibliographystyle{\arxivpath tmlr}

\newpage
\appendix

\section{Analysis of policy gradient: Proofs of Theorem~\ref{thm:entropy_change_clip_general} and Corollary~\ref{thm:entropy_change_clip}}
\label{appendix:entropy_change_clip_proof}

Here we present the proof for Theorem~\ref{thm:entropy_change_clip_general}, followed by Corollary~\ref{thm:entropy_change_clip}.

\begin{proof}[Proof of Theorem~\ref{thm:entropy_change_clip_general}]

We first analyze the first-order Taylor expansion of entropy relative to logit change ($\Delta \theta_{s,a} = \theta^{k+1}_{s,a} - \theta^{k}_{s,a})$. 
This first step is closely inspired by \cite{liu2025_entropy_iteration}.
We can Taylor expand the entropy with respect to $\Delta \theta$:
\[
\mathcal{H}(\pi_{\theta_{k+1}} \mid s) = \mathcal{H}(\pik \mid s) + \left\langle \nabla_\theta \mathcal{H}(\pik \mid s),\ \Delta \theta \right\rangle + \mathcal{O} ((\Delta \theta)^2  )
\]
The gradient of policy entropy is
\begin{align*}
\nabla_\theta \mathcal{H}(\pik \mid s) 
&= \nabla_\theta \left( -\mathbb{E}_{a \sim \pik (\cdot|s)} [\log \pik (a\;|\;s)] \right) \\
& = -\mathbb{E}_{a \sim \pik (\cdot|s)} \left[ \nabla_\theta \log \pik (a\;|\;s) + \log \pik (a\;|\;s) \nabla_\theta \log \pik (a\;|\;s) \right] \\
&= -\mathbb{E}_{a \sim \pik (\cdot|s)} \left[ \log \pik (a\;|\;s) \nabla_\theta \log \pik (a\;|\;s) \right].
\end{align*}

Therefore, we have
\begin{align*}
& \left\langle \nabla_{\theta}\mathcal{H}(\pik \mid s) , \theta^{k+1}-\theta^k \right\rangle \\
&\phantom{=} = - \left\langle \mathbb{E}_{a \sim \pik(\cdot|s)} \left[ \log \pik(a\;|\;s) \nabla_\theta \log \pik(a\;|\;s) \right],\ \theta^{k+1} - \theta^k \right\rangle \\
&\phantom{=} = - \mathbb{E}_{a \sim \pik(\cdot \mid s)} \left[ \log \pik(a\;|\;s) \left\langle \nabla_\theta \log \pik(a\;|\;s),\ \theta^{k+1} - \theta^k \right\rangle \right] \\
&\phantom{=} = - \mathbb{E}_{a \sim \pik(\cdot \mid s)} \left[ \log \pik(a\;|\;s) \sum_{s' \in \mathcal{S}, a' \in \mathcal{A}} \frac{\partial \log \pik(a\;|\;s)}{\partial \theta_{s', a'}} \cdot \left( \theta^{k+1}_{s',a'} - \theta^k_{s',a'} \right) \right] \\
& \phantom{=} = - \sum_{s' \in \mathcal{S}, a' \in \mathcal{A}} \mathbb{E}_{a \sim \pik(\cdot \mid s)} \left[ \log \pik(a\;|\;s) \cdot \frac{\partial \log \pik(a\;|\;s)}{\partial \theta_{s', a'}} \right] \cdot \left( \theta^{k+1}_{s',a'} - \theta^k_{s',a'} \right) \\
& \phantom{=} = - \sum_{s' \in \mathcal{S}, a' \in \mathcal{A}} \left( \theta^{k+1}_{s',a'} - \theta^k_{s',a'} \right) \cdot \mathbb{E}_{a \sim \pik(\cdot \mid s)} \left[ \log \pik(a\;|\;s) \cdot \frac{\partial \log \pik (a\;|\;s)}{\partial \theta_{s', a'}} \right] \\
&\phantom{=} \overset{(\star)}{=} - \! \! \! \! \! \sum_{s' \in \mathcal{S}, a' \in \mathcal{A}} \! \! \left( \theta^{k+1}_{s',a'} - \theta^k_{s',a'} \right) \cdot \mathbf{1}_{\{s = s'\}} \cdot \pik (a' | s) \left( \log \pik(a' | s) - \mathbb{E}_{a \sim \pik(\cdot | s)} [\log \pik(a\;|\;s)] \right)
\end{align*}
where the final equation $(\star)$ holds from the derivation below.
\begin{align*}
& \mathbb{E}_{a \sim \pik(\cdot \mid s)} \left[ \log \pik(a\;|\;s) \cdot \frac{\partial \log \pik (a\;|\;s)}{\partial \theta_{s',a'}} \right]  \\
&\phantom{=} = \mathbb{E}_{a \sim \pik (\cdot|s)} \left[ \log \pik (a\;|\;s) \cdot \frac{\partial}{\partial \theta_{s',a'}} \left( \theta^k_{s,a} - \log \left( \sum_{a \in \mathcal{A}} \exp (\theta^k_{s,a}) \right) \right) \right] \\
&\phantom{=}= \mathbb{E}_{a \sim \pik(\cdot \mid s)} \left[ \log \pik(a\;|\;s) \cdot \mathbf{1}_{\{s = s'\}} \cdot \left( \mathbf{1}_{\{a = a'\}} - \pik(a' \mid s) \right) \right] \notag \\
&\phantom{=}= \mathbf{1}_{\{s = s'\}} \cdot \mathbb{E}_{a \sim \pik(\cdot \mid s)} \left[ \log \pik(a\;|\;s) \cdot \left( \mathbf{1}_{\{a = a'\}} - \pik(a' \mid s) \right) \right] \\
&\phantom{=}= \mathbf{1}_{\{s = s'\}} \cdot \left[ \pik (a' \mid s) \log \pik (a' \mid s) - \pik(a' \mid s) \cdot \mathbb{E}_{a \sim \pik(\cdot \mid s)} \left[ \log \pik(a\;|\;s) \right] \right].
\end{align*}
Hence, we obtain the first-order Taylor expansion of policy entropy:
\begin{equation} \label{eq:entropy_change_by_logits}
    \mathcal{H}(\pi_{\theta_{k+1}} | s) - \mathcal{H}(\pik | s) = - \mathbb{E}_{a \sim \pik(\cdot \mid s)} \left[ \left( \theta^{k+1}_{s,a} - \theta^k_{s,a} \right) \left( \log \pik (a | s) + \mathcal{H}(\pik | s ) \right) \right] + \mathcal{O}((\Delta \theta)^2).
\end{equation}

For our next step, we express the logit change in terms of clipping events. The rollout distribution and the advantages are fixed when differentiating the clipped surrogate objective. Away from the clipping boundaries, we have
\[
\frac{\partial}{\partial r}C_{\varepsilon}(r,A)
=A+\max(-A,0)\mathbf{1}_{\{r<1-\epslow\}}
-\max(A,0)\mathbf{1}_{\{r>1+\epshigh\}}.
\]
Using the definition of the state visitation measure, the softmax derivative, and the importance ratio, we obtain
\begin{align*}
\frac{\partial}{\partial\theta_{s,a}}\mathcal{J}(\pi_{\theta_k};\pi_{\mathrm{old}})&=d^{\pi_{\mathrm{old}}}(s)
\mathbb{E}_{a'\sim\pi_{\mathrm{old}}(\cdot\mid s),A}
\left[\frac{\partial}{\partial\theta_{s,a}}C_{\varepsilon}(r_k(a'\mid s),A)\right]\\\color{red}
&=d^{\pi_{\mathrm{old}}}(s)
\mathbb{E}_{a'\sim\pik(\cdot\mid s),A}\Big[
\big(A+\max(-A,0)\mathbf{1}_{\{a'\in X_k(s)\}}
-\max(A,0)\mathbf{1}_{\{a'\in Y_k(s)\}}\big)\\\color{red}
&\hspace{15em}\cdot\big(\mathbf{1}_{\{a'=a\}}-\pik(a\mid s)\big)\Big].
\end{align*}
The change from $\pi_{\mathrm{old}}$ to $\pik$ changes only the action marginal; the conditional distribution of $A$ given $s,a'$ remains that of the fixed rollout.

 Since $\partial\mathcal{H}(\pik\mid s)/\partial\theta_{s,a}=-Q_k(a\mid s)$, substituting the update $\eta\nabla_\theta\mathcal{J}$ into Equation~\eqref{eq:entropy_change_by_logits} gives
\begin{align*}
\mathcal{H}(\pi_{\theta_{k+1}}\mid s)-\mathcal{H}(\pik\mid s)&=\eta d^{\pi_{\mathrm{old}}}(s)\Big(
\mathbb{E}\big[A+\max(-A,0)\mathbf{1}_{\mathcal{X}_k}
-\max(A,0)\mathbf{1}_{\mathcal{Y}_k}\big]\mathbb{E}[Q_k]\\
&\quad-\mathbb{E}\big[
\big(A+\max(-A,0)\mathbf{1}_{\mathcal{X}_k}
-\max(A,0)\mathbf{1}_{\mathcal{Y}_k}\big)Q_k\big]
\Big)+\mathcal{O}(\eta^2).
\end{align*}
Separating the three contributions and using $\mathbb{E}[\mathbf{1}_{\mathcal{X}_k}(\cdot)]=p_k\mathbb{E}[\cdot\mid\mathcal{X}_k]$ and $\mathbb{E}[\mathbf{1}_{\mathcal{Y}_k}(\cdot)]=q_k\mathbb{E}[\cdot\mid\mathcal{Y}_k]$ proves Theorem~\ref{thm:entropy_change_clip_general}.
\end{proof}

\begin{proof}[ Proof of Corollary~\ref{thm:entropy_change_clip}]
Under the random-reward conditions, $A$ is independent of the action at the fixed state $s$. Therefore, under the fixed rollout's conditional distribution given $s,a$,
\[
\mathbb{E}[A\mid s,a]=0,
\qquad
\mathbb{E}[\max(-A,0)\mid s,a]
=\mathbb{E}[\max(A,0)\mid s,a]
=\mu\nu.
\]
Substituting these identities into Theorem~\ref{thm:entropy_change_clip_general} eliminates the unclipped contribution. Since $Q_k$ and the clipping events depend only on the action at the fixed state $s$, the two clipping contributions become
\[
\mu\nu\eta d^{\pi_{\mathrm{old}}}(s)\big(
p_k(\mathbb{E}[Q_k]-\mathbb{E}[Q_k\mid\mathcal{X}_k])
-q_k(\mathbb{E}[Q_k]-\mathbb{E}[Q_k\mid\mathcal{Y}_k])
\big),
\]
which yields the claimed first-order approximation.
\end{proof}

\section{Analysis of natural policy gradient: Proofs of Theorem~\ref{thm:entropy_change_clip2_general} and Corollary~\ref{thm:entropy_change_clip2}}
\label{appendix:proof2}

Here we present the proof for Theorem~\ref{thm:entropy_change_clip2_general}, followed by Corollary~\ref{thm:entropy_change_clip2}.
\begin{proof}[Proof of Theorem~\ref{thm:entropy_change_clip2_general}]
We first obtain the first-order Taylor expansion of policy entropy relative to the policy change $\Delta \pi = \pi_{k+1}(s) - \pi_k (s) := (\pi_{k+1}(a|s) -\pi_{k}(a|s))_{a\in \mathcal{A}}$.
\begin{align} 
    &\mathcal{H}(\pi_{\theta_{k+1}} \mid s) - \mathcal{H}(\pik \mid s) \\ 
    &\phantom{=}= \langle \pi_{\theta_{k+1}} (a\;|\;s) - \pik (a\;|\;s), \nabla_{\pi} \mathcal{H} (\pik \mid s) \rangle +  \mathcal{O}(\|\Delta \pi\|^2) \notag \\ 
    &\phantom{=}= \sum_{a\in \mathcal{A}} (\pi_{\theta_{k+1}}(a\;|\;s) - \pik (a\;|\;s) ) \frac{\partial}{\partial \pik (a\;|\;s)} (-\pik(a\;|\;s) \log \pik (a\;|\;s) ) + \mathcal{O}(\|\Delta \pi\|^2) \notag \\
    &\phantom{=}= -\sum_{a\in \mathcal{A}} (\pi_{\theta_{k+1}}(a\;|\;s) - \pik (a\;|\;s) ) (\log \pik (a\;|\;s) +1) + \mathcal{O} (\|\Delta \pi\|^2) \notag
    \\
    &\phantom{=}= -\sum_{a\in \mathcal{A}} (\pi_{\theta_{k+1}}(a\;|\;s) - \pik (a\;|\;s) ) \log \pik (a\;|\;s) - \sum_{a\in \mathcal{A}} (\pi_{\theta_{k+1}}(a\;|\;s) - \pik (a\;|\;s) )  + \mathcal{O} (\|\Delta \pi\|^2) \notag
    \\  
    &\phantom{=}= - \mathbb{E}_{a \sim \pik(\cdot \mid s)} \left[ \left( \frac{\pi_{\theta_{k+1}} (a\;|\;s)}{\pik (a\;|\;s)} - 1  \right)  \log \pik(a\;|\;s)  \right] + \mathcal{O}(\|\Delta \pi\|^2) \label{eq:entropy_change_by_logits_npg}
\end{align}

For our next step, we express the policy change in terms of clipping events. Differentiating the clipped surrogate objective with respect to $\pi(a\mid s)$ gives
\begin{align*}
\frac{\partial}{\partial\pi(a\mid s)}\mathcal{J}(\pi;\pi_{\mathrm{old}})
&=d^{\pi_{\mathrm{old}}}(s)\piold{a}{s}
\mathbb{E}\left[\left.
\frac{\partial}{\partial\pi(a\mid s)}C_{\varepsilon}(r_k(a\mid s),A)
\,\right|\,s,a\right]\\
&=d^{\pi_{\mathrm{old}}}(s)\mathbb{E}\Big[
A+\max(-A,0)\mathbf{1}_{\{a\in X_k(s)\}} -\max(A,0)\mathbf{1}_{\{a\in Y_k(s)\}}
\,\Big|\,s,a\Big].
\end{align*}
Here the conditional expectation is over the advantage from the fixed rollout, and the importance-ratio derivative cancels the factor $\piold{a}{s}$.

The natural policy gradient update in Equation~\eqref{eq:NPG} satisfies
\[
\frac{\pi_{\theta_{k+1}}(a\mid s)}{\pik(a\mid s)}
=\frac{\exp\big(\eta\,\partial_{\pi(a\mid s)}\mathcal{J}\big)}
{\sum_{a'}\pik(a'\mid s)\exp\big(\eta\,\partial_{\pi(a'\mid s)}\mathcal{J}\big)}.
\]
Expanding the numerator and denominator to first order yields
\[
\frac{\pi_{\theta_{k+1}}(a\mid s)}{\pik(a\mid s)}-1
=\eta\Big(\partial_{\pi(a\mid s)}\mathcal{J}
-\mathbb{E}_{a'\sim\pik(\cdot\mid s)}[\partial_{\pi(a'\mid s)}\mathcal{J}]\Big)
+\mathcal{O}(\eta^2),
\]
where the derivatives are evaluated at $\pik$. Substituting into Equation~\eqref{eq:entropy_change_by_logits_npg}, and using $\mathbb{E}_{a\sim\pik}[\log\pik(a\mid s)]=-\mathcal{H}(\pik\mid s)$, gives
\begin{align*}
&\mathcal{H}(\pi_{\theta_{k+1}}\mid s)-\mathcal{H}(\pik\mid s)\\
&\quad=\eta d^{\pi_{\mathrm{old}}}(s)
\mathbb{E}\Big[
\big(A+\max(-A,0)\mathbf{1}_{\mathcal{X}_k}
-\max(A,0)\mathbf{1}_{\mathcal{Y}_k}\big)\\
&\hspace{12em}\cdot\big(-\log\pik(a\mid s)-\mathcal{H}(\pik\mid s)\big)
\Big]+\mathcal{O}(\eta^2).
\end{align*}
Separating the three contributions and using $\mathbb{E}[\mathbf{1}_{\mathcal{X}_k}(\cdot)]=p_k\mathbb{E}[\cdot\mid\mathcal{X}_k]$ and $\mathbb{E}[\mathbf{1}_{\mathcal{Y}_k}(\cdot)]=q_k\mathbb{E}[\cdot\mid\mathcal{Y}_k]$ proves Theorem~\ref{thm:entropy_change_clip2_general}.
\end{proof}

\begin{proof}[Proof of Corollary~\ref{thm:entropy_change_clip2}]
As in the proof of Corollary~\ref{thm:entropy_change_clip}, the random-reward conditions imply
\[
\mathbb{E}[A\mid s,a]=0,
\qquad
\mathbb{E}[\max(-A,0)\mid s,a]
=\mathbb{E}[\max(A,0)\mid s,a]
=\mu\nu,
\]
where the conditional expectations are under the fixed rollout distribution. Substituting into Theorem~\ref{thm:entropy_change_clip2_general} eliminates the unclipped contribution and gives
\begin{align*}
\mu\nu\eta d^{\pi_{\mathrm{old}}}(s)\big(
&p_k(\mathbb{E}[-\log\pik\mid\mathcal{X}_k]-\mathcal{H}(\pik\mid s))-q_k(\mathbb{E}[-\log\pik\mid\mathcal{Y}_k]-\mathcal{H}(\pik\mid s))
\big)
\end{align*}
for the first-order entropy change, proving the corollary.
\end{proof}

\section{Proof of Theorem~\ref{thm:cond_uncond_Q_inequality}} \label{appendix:proof3}


Here we present the proof for Theorem~\ref{thm:cond_uncond_Q_inequality}.

\begin{proof}

    Let $\max_{a \in \mathcal{A}} \piold{a}{s} = \piold{a^*}{s} = 1 - \delta$. Then, by heavy-head assumption $\delta \leq \delta_0$.
For the one-step stochastic policy gradient update, for probability $\piold{a^*}{s} = 1- \delta \geq 1- \delta_0$, the gradient descent update chooses the head action by the heavy-head assumption. Conditioning on this event, the update direction becomes (note that at this point, clipping does not happen because we are updating $\pi_{\mathrm{old}}$)
    \begin{equation} \label{eq:gd_gradient}
    g_a = A ( \mathbf{1}\{ a = a^* \} - \piold{a}{s})
    \end{equation}
    where we omitted the $s$, and the updated policy $\pik$ is
    \[
    \pik (a \mid s) = \frac{\piold{a}{s} e^{\eta g_a} }{\sum_{b \in \mathcal{A}} \piold{b}{s} e^{\eta g_b}}
    \]
    for all $a \in \mathcal{A}$. 
Next, we define the normalization factor $Z$ as
\[
Z := \sum_{b \in \mathcal{A}} \piold{b}{s} e^{\eta g_b} = (1- \delta) e^{\eta A \delta} + \sum_{a \neq a^*} \piold{a}{s} e^{-\eta A \piold{a}{s}}.
\]
Observe from \eqref{eq:gd_gradient} that 
    \begin{align} \label{eq:gd_gradient_cases}
    g_a = \begin{cases} 
A ( 1 - \piold{a^*}{s} ) = A \delta & \text{if } a = a^* \\ 
- A \piold{a}{s} & \text{if } a \neq a^* 
\end{cases}.
    \end{align}
Then, we can write the importance ratio $r_k (a \mid s ) = \frac{ \pik (a \mid s) } { \piold{a}{s}}$ as
\[
    r_k (a \mid s ) =
    \begin{cases}
        \frac{e^{\eta A \delta}}{Z} \quad &\text{ if $a = a^*$} \\
        \frac{e^{-\eta A \piold{a}{s}}}{Z} \quad &\text{ if $a \neq a^*$}
    \end{cases}.
\]

We proceed by showing that the `head' action $a^*$ is not in the \emph{clipping sets}, i.e., $a^* \notin X_k(s)$ and $a^* \notin Y_k(s)$. 

First consider the clip-low case. Note that $A$ has to be negative for $r_k (a^* \mid s)$ to be smaller than $1$. Then, since $\piold{a}{s} \leq \delta$ for all $a \neq a^*$, 
\begin{align*}
r_k (a^* \mid s)
&= \frac{1}{(1-\delta)+\sum_{a \neq a^*}\piold{a}{s}e^{-\eta A(\piold{a}{s}+\delta)}}\\
&\geq \frac{1}{(1-\delta)+\delta e^{2\eta A_{\max}\delta}}
\geq \frac{1}{1+\frac{\delta}{1-2\eta A_{\max}\delta}}\\
&\geq \frac{1}{1+\epslow}
\geq 1-\epslow.
\end{align*}
Here $2\eta A_{\max}\delta<1$ follows from $\delta\leq u_0/(1+2\eta A_{\max}u_0)$.
We used $e^x\leq (1-x)^{-1}$ for $0\leq x<1$ and the constraint
\[
\delta\leq \frac{\epslow}{1+2\eta A_{\max}\epslow}
\quad\Longleftrightarrow\quad
\frac{\delta}{1-2\eta A_{\max}\delta}\leq \epslow.
\]
Hence we have $a^* \notin X_k$. 
For the clip-high case, 
\[
r_k (a^* \mid s) = \frac{1}{ (1-\delta) + \sum_{a \neq a^*} \piold{a}{s}e^{-\eta A (\piold{a}{s}+\delta)}} 
\leq \frac{1}{1-\delta} \leq 1 + \epshigh,
\]
where we used $\delta \leq \frac{\epshigh}{1 + \epshigh}$ for the last inequality. Hence $a^* \notin Y_k(s)$. 





Now, recall that $Q_k (a \mid s) = \pik (a \mid s) ( \log \pik (a \mid s) + \mathcal{H}(\pik \mid s))$.
The following lemma shows that the unconditional expectation of $Q_k$ is always nonnegative.

\begin{lemma} \label{lem:Eq_nonnegative}
    Let $ q \in \Delta (\mathcal{A})$ be any distribution with discrete actions and let $Q(a) = q(a) ( \log q(a) + \mathcal{H} (q))$. Then,
    \[
    \mathbb{E}_{a \sim q(\cdot)} [ Q(a) ] \geq 0. 
    \]
\end{lemma}

\begin{proof}
    Observe that
    \[
    \mathbb{E}_{a \sim q(\cdot)} [ Q(a) ] = 
    \mathbb{E}_{a \sim q(\cdot)} [q(a)\log q(a)] - \mathbb{E}_{a \sim q(\cdot)}[q(a)] \mathbb{E}_{a \sim q(\cdot)} [\log q(a)]
    =
    \mathrm{Cov} ( q, \log q).
    \]
    Let $A, A'$ be i.i.d. random actions sampled from  $q$. Then the covariance identity yields
    \[
    \mathrm{Cov}(q, \log q) = \frac{1}{2} \mathbb{E}  [ ( q(A) - q(A')) ( \log q(A) - \log q(A') ) ],
    \]
    but the integrand is always pointwise nonnegative since $\log(\cdot)$ is an increasing function. Hence $\mathbb{E}_{a \sim q(\cdot)} [ Q(a) ] \geq 0$.
\end{proof}

Using this result, we first establish the corresponding unweighted inequalities by showing that
$\mathbb{E} [ Q_k  \mid \mathcal{X}_k]
= \mathbb{E}_{a \sim \pik(\cdot \mid s)} [ Q_k(a \mid s)  \mid a \in X_k] \leq 0$ and $\mathbb{E} [ Q_k \mid \mathcal{Y}_k ] 
= \mathbb{E}_{a \sim \pik(\cdot \mid s)} [ Q_k (a \mid s) \mid a \in Y_k ]
\leq 0$.
We first prove another lemma to show this. 

\begin{lemma} \label{lem:cond_Q_nonnegative}
    Let $ q \in \Delta (\mathcal{A})$ be any distribution with discrete actions and let $Q(a) = q(a) ( \log q(a) + \mathcal{H} (q))$. Assume
    \[
    q(a) \leq e^{-\mathcal{H}(q)}
    \]
    for some $a \in \mathcal{A}$. Then, 
    \[
    Q(a) \leq 0.
    \]
\end{lemma}

\begin{proof}
    For $a \in \mathcal{A}$ where $q(a) \leq e^{- \mathcal{H}(q)}$, 
    \[
    Q(a) = q(a) ( \log q(a) + \mathcal{H}(q) ) = q(a) \pr{ \log \frac{q(a)}{e^{-\mathcal{H}(q)}} } \leq 0. 
    \]
\end{proof}

Therefore, if we could show that $\pik(a \mid s) \leq e^{-\mathcal{H}(\pik \mid s) } $ for all $a \in X_k$, we could conclude that $\mathbb{E} [ Q_k \mid \mathcal{X}_k  ] \leq 0$. The same argument holds for $Y_k$ as well. 
Since we have already shown that the head action $a^* \notin X, Y$, we aim to show that $\pik(a \mid s) \leq e^{-\mathcal{H}(\pik \mid s)}$ holds for all $a \neq a^*$.

We next bound the total tail probability.
From \eqref{eq:gd_gradient_cases} we observe $|g_a|\leq A_{\max}\delta$ for all $a\in\mathcal{A}$, so $Z\geq e^{-\eta A_{\max}\delta}$.
Let $\delta_q=\sum_{a\neq a^*}\pik(a\mid s)$. Then,
\[
\delta_q
=\frac{\sum_{a\neq a^*}\piold{a}{s}e^{\eta g_a}}{Z}
\leq \delta e^{2\eta A_{\max}\delta}
\leq \frac{\delta}{1-2\eta A_{\max}\delta},
\]
where we again used $e^x\leq (1-x)^{-1}$ for $0\leq x<1$.
Hence, from the constraint $\delta\leq u_0/(1+2\eta A_{\max}u_0)$, we have $\delta_q\leq u_0$.

Now we aim to lower bound $e^{- \mathcal{H} (\pik \mid s) }$. 
Note that with the head action probability $\pik(a^*) = 1 - \delta_q$ fixed, $\mathcal{H} (\pik \mid s)$ is maximized when the remaining probabilities are distributed evenly. 
That is,
\[
\mathcal{H}(\pik \mid s) \leq -(1-\delta_q) \log (1 - \delta_q) - \delta_q \log \pr{ \frac{ \delta_q} {(|\mathcal{A}| -1) }} = h(\delta_q) + \delta_q \log (|\mathcal{A}| -1 ) 
\]
where $h(x) := -x \log x - (1-x) \log (1-x) $ is the binary entropy function. Now we show that for $ 0 < \delta_q \leq u_0$, the following holds:
\begin{equation} \label{eq:tail_ineq}
e^{-\mathcal{H}(\pik \mid s)}
\geq e^{-h(\delta_q)} (|\mathcal{A}|-1)^{-\delta_q}
\overset{(a)}{\geq} \pr{ \frac{\delta_q}{e (|\mathcal{A}|-1)}  }^{\delta_q}
\overset{(b)}{\geq} \delta_q.
\end{equation}

Inequality $(a)$ uses the fact that $h(x) \leq x - x \log x$ for $x \in (0,1)$. 
For $(b)$, we first define function
\[
F(x) := - (1-x) \log x - x \log (e (|\mathcal{A}| -1 ))
\]
on $x \in (0, 1)$. 
Since
\begin{align*}
    F(x) &\geq 0 
    \quad \Longleftrightarrow \quad
    x \log \pr{ \frac{x}{e(|\mathcal{A}| - 1)}} \geq \log x
    \quad \Longleftrightarrow \quad
    \pr{\frac{x}{e(|\mathcal{A}|-1)}}^{x} \geq x,
\end{align*}
showing that $(b)$ holds for $\delta_q \in (0, u_0)$ is equivalent to showing $F(x) \geq 0 $ for $x \in (0, u_0)$.
Differentiating $F$ yields
\[
F'(x) = - \frac{1}{x} +\log x  - \log (|\mathcal{A}|-1) < 0 
\]
for $x \in (0,1)$, since $|\mathcal{A}| \geq 2$. Hence $F$ is decreasing, and it is sufficient to show that $F(u_0) \geq 0$. 
Indeed, 
\[
F(u_0) = - (1 - u_0) \log u_0 - \frac{\log 2} { 4} > 0 
\]
holds due to the fact $u_0 = \frac{\log 2}{ 4 \log(e (|\mathcal{A}|-1))} \leq \frac{\log 2 }{4} < 1/2$. 

Now, since $\pik(a \mid s) \leq \delta_q$ for all $a \neq a^*$, we have $\pik (a \mid s ) \leq e^{-\mathcal{H}(\pik \mid s)}$ for all $a \neq a^*$ by \eqref{eq:tail_ineq}. 
Since we have shown that $ X_k, Y_k \subset \mathcal{A} \setminus \{a^*\} $,
we could apply Lemma~\ref{lem:cond_Q_nonnegative} to conclude $\mathbb{E}[Q_k \mid \mathcal{X}_k ] \leq 0$ and $\mathbb{E}[Q_k \mid \mathcal{Y}_k ] \leq 0$. 
Finally, applying Lemma~\ref{lem:Eq_nonnegative}, we conclude $\mathbb{E}[Q_k ] - \mathbb{E}[ Q_k  \mid \mathcal{X}_k ] \geq 0 $ and $\mathbb{E}[ Q_k ] - \mathbb{E} [ Q_k  \mid \mathcal{Y}_k ] \geq 0 $.

The same argument also proves the advantage-weighted inequalities in the theorem. Indeed, $Q_k(a\mid s)\leq 0$ for every $a\in X_k\cup Y_k$, while $\mathbb{E}[Q_k]\geq 0$. Therefore, $\mathbb{E}[Q_k]-Q_k(a\mid s)\geq 0$ on both clipping sets. Multiplying this pointwise inequality by $\max(-A,0)$ on $X_k$ and by $\max(A,0)$ on $Y_k$, and taking the corresponding conditional expectations, gives
\begin{align*}
p_k\mathbb{E}\big[\max(-A,0)(\mathbb{E}[Q_k]-Q_k)\mid\mathcal{X}_k\big]&\geq 0,\\
q_k\mathbb{E}\big[\max(A,0)(\mathbb{E}[Q_k]-Q_k)\mid\mathcal{Y}_k\big]&\geq 0.
\end{align*}

\end{proof}

\section{Proof of Theorem~\ref{thm:cond_uncond_inequality}}
\label{appendix:proof4}

Here we present the proof for Theorem~\ref{thm:cond_uncond_inequality}.

\begin{proof}

Let $\max_{a \in \mathcal{A}} \piold{a}{s} = \piold{a^*}{s} = 1 - \delta$. Then, by heavy-head assumption $\delta \leq \delta_0$.
For probability $\piold{a^*}{s} = 1- \delta \geq 1 - \delta_0 $, the natural policy gradient update chooses the head action by the heavy-head assumption. Conditioning on this event, the update direction for state $s$ becomes (note that at this point, clipping does not happen because we are updating $\pi_{\mathrm{old}}$)
\[
g_a = A \cdot \frac{ \mathbf{1} \{a = a^*\}} {\piold {a} {s} } 
\]
where we omitted the $s$, and
\[
\pik (a \mid s) = \frac{\piold{a}{s} e^{\eta g_a} }{\sum_{b \in \mathcal{A}} \piold{b}{s} e^{\eta g_b}}
\]
for all $a \in \mathcal{A}$. This defines the importance sampling ratio
\[
r_k (a \mid s ) = \frac{ \pik (a \mid s) } { \piold{a}{s}}
\]
as in appendix~\ref{appendix:proof3}.

Observe that for natural policy gradient updates, $g_{a^*} = \frac{A}{ \piold{a^*}{s}}$ and $g_a = 0$ if $a \neq a^*$. 
Hence the importance sampling ratio takes only \emph{two} possible values, i.e., 
\[
r_k (a \mid s) = \begin{cases}
\frac{e^{\eta A / \piold{a^*}{s}}}{Z} &\text{if } a = a^* \\
\frac{1}{Z} &\text{if } a \neq a^*
\end{cases}
\]
where the normalization factor $Z$ is written as 
\[
Z = \sum_{b \in \mathcal{A}} \piold {b}{s} e^{\eta g_b} = (1 - \piold{a^*}{s}) + \piold {a^*}{s} e^{\eta A / \piold{a^*}{s}}.
\]

    Now, we characterize the random event set $X_k = \{ a : r_k (a \mid s) < 1 - \epslow\}$. We assert that it is either a singleton $\{a^*\}$, complement of a singleton $\mathcal{A} \setminus \{a^*\}$, or an empty set. $X_k$ is an empty set simply when the update was not large enough for $r_k $ to go lower than the threshold for any action $a \in \mathcal{A}$. 

    Consider the case where $A>0$. Then $r_k(a^* \mid s) = \frac{e^{\eta A / \piold{a^*}{s}}}{Z} = \frac{1}{(1-\piold{a^*}{s} )e^{-\eta A / \piold{a^*}{s}} + \piold{a^*}{s}} > 1$, implying $a^* \notin X_k$. 
    Hence if $X_k$ is nonempty, it has to be $\frac{1}{Z} < 1 - \epslow$, which implies $X_k = \mathcal{A} \setminus \{a^*\}$. 
    
    On the other hand, if $A<0$, we have $Z < 1$, which in turn implies that $r_k(a \mid s) > 1 > 1 - \epslow$ for all $a \neq a^*$. 
    Hence $X_k$ is nonempty if and only if $r(a^* \mid s) < 1 - \epslow$, letting $X_k = \{a^*\}$. 

    We characterize $Y_k = \{ a: r_k (a \mid s) > 1 + \epshigh\}$ in a similar manner. If $A > 0$, $Z$ is greater than 1, and hence $a \notin Y_k$ for all $a \neq a^*$. In this case, $Y_k$ is nonempty only if $r_k(a^* \mid s) = \frac{e^{\eta A / \piold{a^*}{s}}}{Z} > 1 + \epshigh$, letting $Y_k = \{a^*\}$. 
    
    In the case of $A < 0$, then $r_k(a^* \mid s) = \frac{e^{\eta A / \piold{a^*}{s}}}{Z} = \frac{1}{(1-\piold{a^*}{s} )e^{-\eta A / \piold{a^*}{s}} + \piold{a^*}{s}} < 1$, implying $a^* \notin Y_k$. 
    So $Y$ is non-empty only if $\frac{1}{Z} > 1 + \epshigh$, which is achievable and lets $Y_k = \mathcal{A} \setminus \{a^*\}$. 
    In conclusion, $Y_k$ can either be an empty set, singleton ($\{a^*\}$), or a complement of a singleton ($\mathcal{A} \setminus \{a^*\}$), as it was the case with $X_k$. 
    
    We proceed to showing that with small enough $\delta > 0 $ satisfying the given constraints in the theorem statement, cases $X_k = \{a^*\}$ or $Y_k = \{a^*\}$ cannot happen. 
    If $X_k = \{a^*\}$, $r_k (a^* \mid s) = \frac{e^{\eta A/ \piold{a^*}{s}}}{Z} < 1 - \epslow$. With some algebra, we can show that this is equivalent to 
    \begin{equation} \label{eq:A_ineq}
    A 
    < 
    - \frac{1-\delta}{\eta} \log \pr{ \frac{\epslow + \delta(1-\epslow)}{\delta (1 -  \epslow)}} 
    \leq - \frac{1}{2\eta} \log \pr{ \frac{ \epslow}{\delta ( 1- \epslow)}}
    \end{equation}
    since $\piold{a^*}{s} = 1- \delta \geq \frac{1}{2}$.
    However, since $\delta \leq \frac{\epslow}{1-\epslow}e^{-2 \eta A_{\max}}$, we have 
    \[
    - A_{\max} \geq -\frac{1}{2 \eta} \log \pr{ \frac{\epslow}{\delta ( 1- \epslow}}
    \]
    which contradicts~\eqref{eq:A_ineq}. 
    
    Now assume that $Y_k = \{ a^* \}$, i.e. $r(a^* \mid s) = \frac{e^{\eta A/ \piold{a^*}{s}}}{Z} > 1 + \epshigh $. 
    Again, this yields a contradiction since by assumption $\delta \leq \frac{\epshigh}{1+\epshigh}$, we obtain
    \[
    r_k ( a^* \mid s) = \frac{ \pik ( a^* \mid s ) }{ \piold {a^*} {s} } \leq \frac{1}{1- \delta} \leq 1 + \epshigh. 
    \]
    
    Hence for both $X_k$ and $Y_k$, they are either empty or equal to $\mathcal{A} \setminus \{ a^* \}$. 

Since the desired inequalities hold trivially when $X_k$ and $Y_k$ are empty sets, it suffices to show that the inequality holds when $X_k = \mathcal{A} \setminus \{ a^* \}$ or $Y_k = \mathcal{A} \setminus \{ a^* \}$. 
In order to prove that, we first prove the following lemma. 
    \begin{lemma}
        Let $q \in \Delta (\mathcal{A})$ be any distribution and let $a^* \in \mathcal{A}$ satisfy $q(a^*) = \max_{a} q(a)$. Let $S = \mathcal{A} \setminus \{a^*\}$. Then,
        \[
        \mathbb{E}_{a \sim q(\cdot)} [ - \log q(a) \mid a \in S ] - \mathcal{H}(q) \geq 0
        \]
        where $\mathcal{H}(q)$ is the entropy of $q$. 
    \end{lemma}

    \begin{proof}
    By direct calculation, 
    \[
    \mathbb{E}_q [ - \log q | S ] = \frac{1}{1- q(a^*) } \sum_{b \neq a^*} q(b) ( - \log q(b) ) 
    = \frac{\mathcal{H}(q) - q(a^*) ( - \log q(a^*))}{1- q(a^*) }
    \]
    leading to 
    \[
    \mathbb{E}_q [ - \log q | S ] - \mathcal{H}(q) = \frac{q(a^*)}{1-q(a^*)} ( \mathcal{H}(q) + \log q(a^*)) \geq 0,
    \]
    where the last inequality is due to
    \[
    \mathcal{H} (q) = \mathbb{E}_{b \sim q(\cdot)} [ - \log q(b) ] 
    \geq - \log q(a^*).
    \]
    \end{proof}

Applying the result of the lemma, it suffices to show that $\pik (a^* \mid s) = \max_{a \in \mathcal{A}} \pik (a \mid s)$. Note that this is not obvious, since by definition $a^* = \argmax_{a \in \mathcal{A}} \piold {a}{s} $. 
Observe that for any $j \neq a^*$, 
\[
\frac{\pik(j \mid s)}{\pik(a^* \mid s)} = \frac{\piold{j}{s}}{\piold{a^*}{s} e^{\eta A / \piold{a^*}{s}  } } = \frac{\piold{j}{s}}{(1- \delta) e^{\eta A / (1- \delta)}} 
\overset{(a)}{\leq} \frac{2 \piold{j}{s}}{e^{-2\eta A_{\max}}} 
\overset{(b)}{\leq} \frac{\piold{j}{s}}{\delta} \overset{(c)}{\leq} 1,
\]
where $(a)$ uses $\delta \leq \frac{1}{2}$, $(b)$ comes from the assumption $\delta \leq \frac{1}{2} e^{-2\eta A_{\max}}$, and $(c)$ is due to the fact that $\piold {j}{s} \leq \delta$ for any $j \neq a^*$.
Hence, we have that $a^* = \argmax_{a \in \mathcal{A}} \pik (a \mid s)$, which proves the corresponding unweighted inequalities.

To prove the advantage-weighted inequalities, we further bound the total tail probability. Since $\delta\leq u_0e^{-2\eta A_{\max}}$ and $\delta\leq 1/2$, the update above gives
\begin{align*}
\sum_{a\neq a^*}\pik(a\mid s)
&=\frac{\delta}{(1-\delta)e^{\eta A/(1-\delta)}+\delta}\\
&\leq\frac{\delta e^{2\eta A_{\max}}}{1+\delta(e^{2\eta A_{\max}}-1)}
\leq\delta e^{2\eta A_{\max}}\leq u_0.
\end{align*}
By the entropy bound in Equation~\eqref{eq:tail_ineq}, this implies $\pik(a\mid s)\leq e^{-\mathcal{H}(\pik\mid s)}$ for every $a\neq a^*$. Hence $-\log\pik(a\mid s)-\mathcal{H}(\pik\mid s)\geq 0$ on both clipping sets. Multiplying by $\max(-A,0)$ on $X_k$ and by $\max(A,0)$ on $Y_k$, and taking conditional expectations, yields
\begin{align*}
p_k\mathbb{E}\big[\max(-A,0)(-\log\pik-\mathcal{H}(\pik\mid s))\mid\mathcal{X}_k\big]&\geq 0,\\
q_k\mathbb{E}\big[\max(A,0)(-\log\pik-\mathcal{H}(\pik\mid s))\mid\mathcal{Y}_k\big]&\geq 0.
\end{align*}

\end{proof}

\section{Experimental Configurations} \label{appendix:experimental_settings}

\begin{table}[h]
\centering
\caption{Training configurations used for \texttt{GSM8K} dataset / \texttt{DAPO-Math-17k} dataset.}
\setlength{\tabcolsep}{10pt}
\renewcommand{\arraystretch}{1.2}
\begin{tabular}{l l}
\hline
\textbf{Hyperparameter} & \textbf{Value} \\
\hline
Optimizer & AdamW \\
Learning rate & $5\times 10^{-7}$ / $1\times10^{-6}$ \\
GRPO batch size & 512 \\
Optimizer batch size & 256 \\
Policy updates per rollout & 16 \\
Group Size & 8 \\
Max response length & 4096 \\
Temperature (train) & 1.0 \\
Temperature (validation) & 1.0 \\
Top p (train) & 1.0 \\
Top p (validation) & 0.95 \\
Dynamic Sampling & None / True \\
Overlong penalty factor & None / 1.0 \\
\hline
\end{tabular}
\label{tab:rl-hparams}
\end{table}

We use the \texttt{verl} framework~\citep{sheng2025hybridflow} for all experiments. We use public models, datasets, and software and cite their original sources. Licenses include Apache-2.0 for \texttt{verl}, DAPO-Math-17k, OLMo-2, Qwen2.5-1.5B/7B, and yentinglin/aime\_2025; MIT for GSM8K; Meta Llama licenses for Llama models; and Qwen Research/Qwen licenses for Qwen2.5-3B/72B. We do not redistribute these assets.
All experiments used NVIDIA H100 GPUs. One full RLVR run used 2 H100s for about 5 days, i.e., roughly 240 H100-GPU hours; two runs were typically run in parallel on a 4-H100 node. The reported training runs used compute on the order of $10^4$ H100-GPU hours.

For the random reward RL training experiments, we used the \texttt{GSM8K} dataset as the traning dataset, and conducted experiments with base models \texttt{Qwen2.5-1.5B-Instruct} \citep{qwen2024_techreport} and \texttt{Llama-3.2-1B-Instruct} \citep{grattafiori2024llama}. 
Rewards were sampled from the Bernoulli distribution with $0.5$ probability.  
For general mathematical reasoning tasks, we train the \texttt{Qwen2.5-7B-Instruct} model with the \texttt{DAPO-Math-17k} \citep{Yu2025_dapo} dataset, and validate it on \texttt{MATH-500} \citep{hendrycks2021math}, \texttt{AMC23} \citep{AIMO_validation_AMC}, \texttt{AIME2024},  and \texttt{AIME2025} datasets \citep{AIME_2024_HF}.
We also train \texttt{Qwen2.5-3B-Instruct} and \texttt{Llama-3-8B-Instruct} model with the \texttt{GSM8K} dataset, and validate it on the \texttt{GSM8K} \citep{cobbe2021training} test dataset.
For validation, we perform string match for the last numerical value for \texttt{GSM8K} test datasets, and use the \texttt{Math-Verify} \citep{math_verify_github} package. 

We use different training configurations for the \texttt{GSM8K} and \texttt{DAPO-MATH-17k} dataset, and separate them with /. In Table~\ref{tab:rl-hparams}, we provide the training and generation details for the experiments in the paper. For all experiments, unless stated otherwise, KL divergence loss or entropy regularization loss were not deployed.

For the \texttt{Qwen2.5-7B-Instruct} clipping ablation, we use three independent matched training runs and the common-checkpoint evaluation protocol described in Appendix~\ref{appendix:dapo_experiments}.

\section{ Heavy head assumption details }
\label{appendix:heavy_head_assumption}

In this section, we provide further details regarding the justification of the heavy-head assumption used in Section~\ref{subsec:clip_characterize}.

\subsection{ RL training of randomly-initialized base models } \label{appendix:subsec:random_init_model}

In Section~\ref{subsec:clip_characterize}, we showed how the policy distribution determines the clipping signs in Inequalities~\eqref{eq:Q-sign} and~\eqref{eq:log-sign}. In the random-reward setting considered here, the unclipped contribution vanishes, so these signs determine the direction of the entropy bias.
To demonstrate this point, we experiment with the \emph{from-scratch} \texttt{Qwen2.5-1.5B} model with random weight initializations. 
Specifically, we call the model from \texttt{huggingface} using the \texttt{Automodel.from\_config()} method which does not call the pretrained weights, and uses randomly initialized weights.
We train with random rewards with the \texttt{GSM8K} training dataset, and set a more aggressive clipping with $\epshigh = \epslow = 0.02$, since with the standard setting with $\epshigh = \epslow = 0.2$, clipping events almost never happened.

\begin{table}[h]
    \centering
    \caption{ Difference of maximum entropy and minimum entropy throughout random reward RL training of \texttt{Qwen2.5-1.5B} model with randomly initialized weights. }
    \begin{tabular}{ccccc}
    \toprule
       $\epshigh$ / $\epslow$ & $ \{ \max_k \mathcal{H}(\pik)$  - $\min_k \mathcal{H}(\pik) \}$ $\times 10^5$ \\
    \midrule
        $0.02 \;$ / $\; 0.02$  & 4.20 \\
        $\infty \;$ / $\;0.02$  & 4.96 \\
        $0.02 \;$ / $\; 1.0$  & 5.63 \\
    \bottomrule
    \end{tabular}
    \label{tab:from_scratch_training}
\end{table}

As shown in Table~\ref{tab:from_scratch_training},
the policy entropy stays nearly constant throughout the whole training process of $30$ epochs, with the fluctuation being in the order of $10^{-5}$. 
This is in stark contrast with the result with pretrained models, where the clipping bias on entropy clearly shows; compare with results in Figure~\ref{fig:entropy_ablation} or Appendix~\ref{appendix:random_reward_ablation}. 
This result strengthens the viewpoint that the directional clipping bias on entropy is due to the shape of the probability distribution of the pretrained model, making it a RL-LLM specific phenomenon.

\subsection{ Empirical validation of heavy-head assumption on larger models.} 
\label{appendix:subsec:heavy_head_validation}

Indeed, the heavy-head assumption is the key bridge from theory to practical RLVR, and hence  we have repeated the assumption verification with various model sizes and datasets: the result is summarized in Table~\ref{tab:heavy_head_extra} and Table~\ref{tab:heavy_head_extra2}. 
We broadly observe the same qualitative pattern: a large fraction of visited states satisfy $\max_a \pi ( a \mid s ) \geq 1 - \delta_0$ with reasonably small  $\delta_0$. 

Importantly, these measurements are taken before RLVR post-training. As shown in the manuscript, this skewness typically becomes even stronger as RL training proceeds. Taken together, these results indicate that the heavy-head phenomenon is an assumption that holds in practice in the regimes and models of interest. 

    \begin{table}[hbtp]
        \centering
        \caption{Heavy-head assumption validation on \texttt{Qwen2.5-72B-Instruct} model.}
        \begin{tabular}{cccc}
        \toprule
            Dataset & $\delta_0 = 0.3$ & $\delta_0 = 0.1$ & $\delta_0 = 0.01 $\\
        \midrule
            GSM8K & 0.96 & 0.91 & 0.84 \\
            MATH500 & 0.96 & 0.92 & 0.86 \\
            AIME2024 & 0.94 & 0.86 & 0.79 \\
            AMC2023 & 0.95 & 0.90 & 0.83  \\
            GPQA diamond & 0.92 & 0.85 & 0.76  \\
            \bottomrule
        \end{tabular}

        \label{tab:heavy_head_extra}
    \end{table}    

    \begin{table}[hbtp]
        \centering
                \caption{Heavy-head assumption validation on \texttt{Llama3-70B-Instruct} model. }
        \begin{tabular}{cccc}
        \toprule
            Dataset & $\delta_0 = 0.3$ & $\delta_0 = 0.1$ & $\delta_0 = 0.01 $\\
            \midrule
            GSM8K & 0.80 & 0.66 & 0.50 \\
            MATH500 & 0.87 & 0.78 & 0.62 \\
            AIME2024 & 0.86 & 0.76 & 0.64 \\
            AMC2023 & 0.85 & 0.76 & 0.64  \\
            GPQA diamond & 0.90 & 0.82 & 0.70  \\
        \bottomrule
        \end{tabular}  

        \label{tab:heavy_head_extra2}
    \end{table}




\section{KL divergence ablation experiment}
\label{appendix:kl_divergence}

\begin{figure*}[hbtp]
  \centering
  \begin{subfigure}[t]{0.48\textwidth}
    \centering
    \includegraphics[width=\linewidth]{\arxivpath 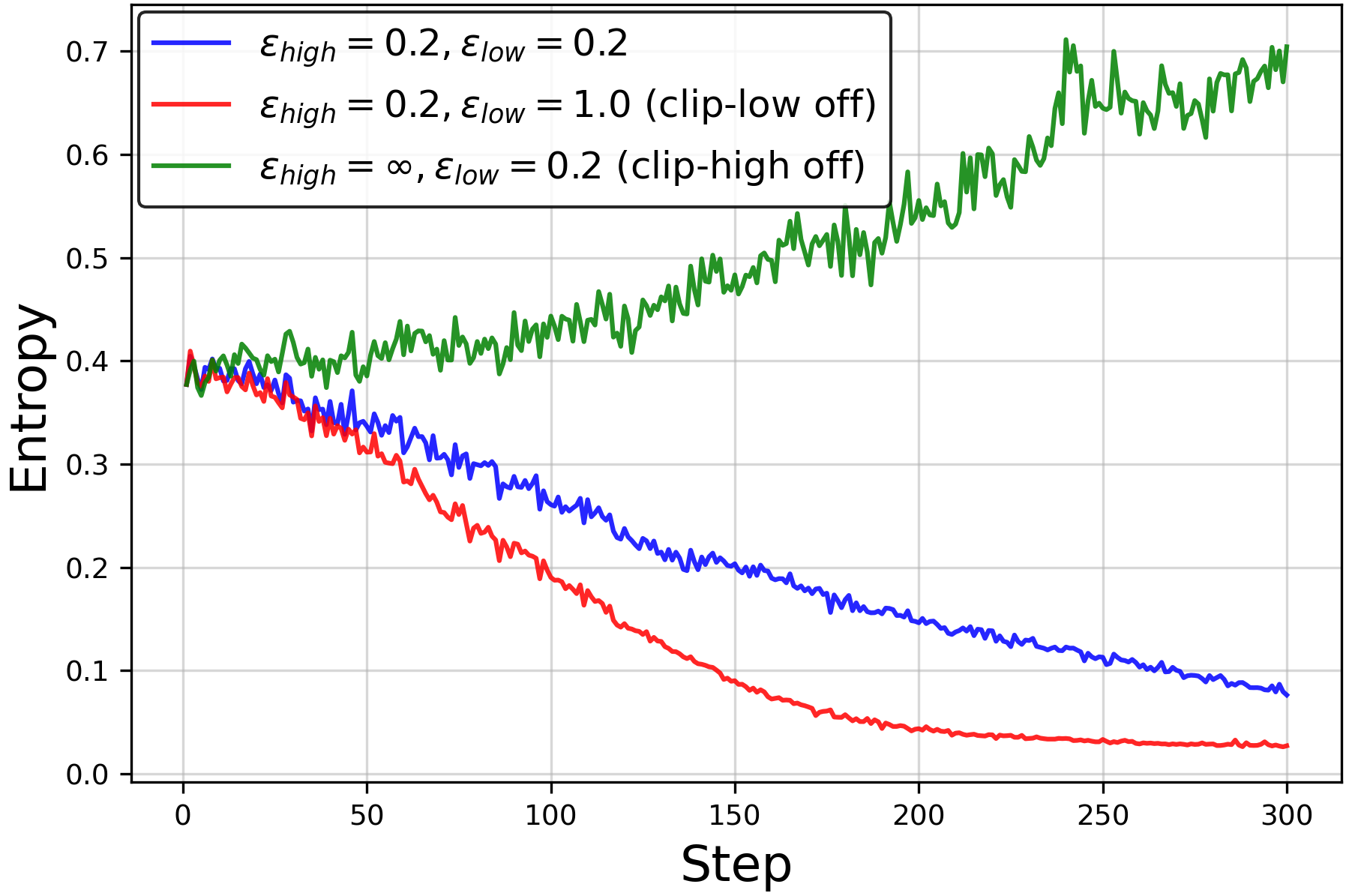}
  \end{subfigure}\hfill
  \begin{subfigure}[t]{0.48\textwidth}
    \centering
    \includegraphics[width=\linewidth]{\arxivpath 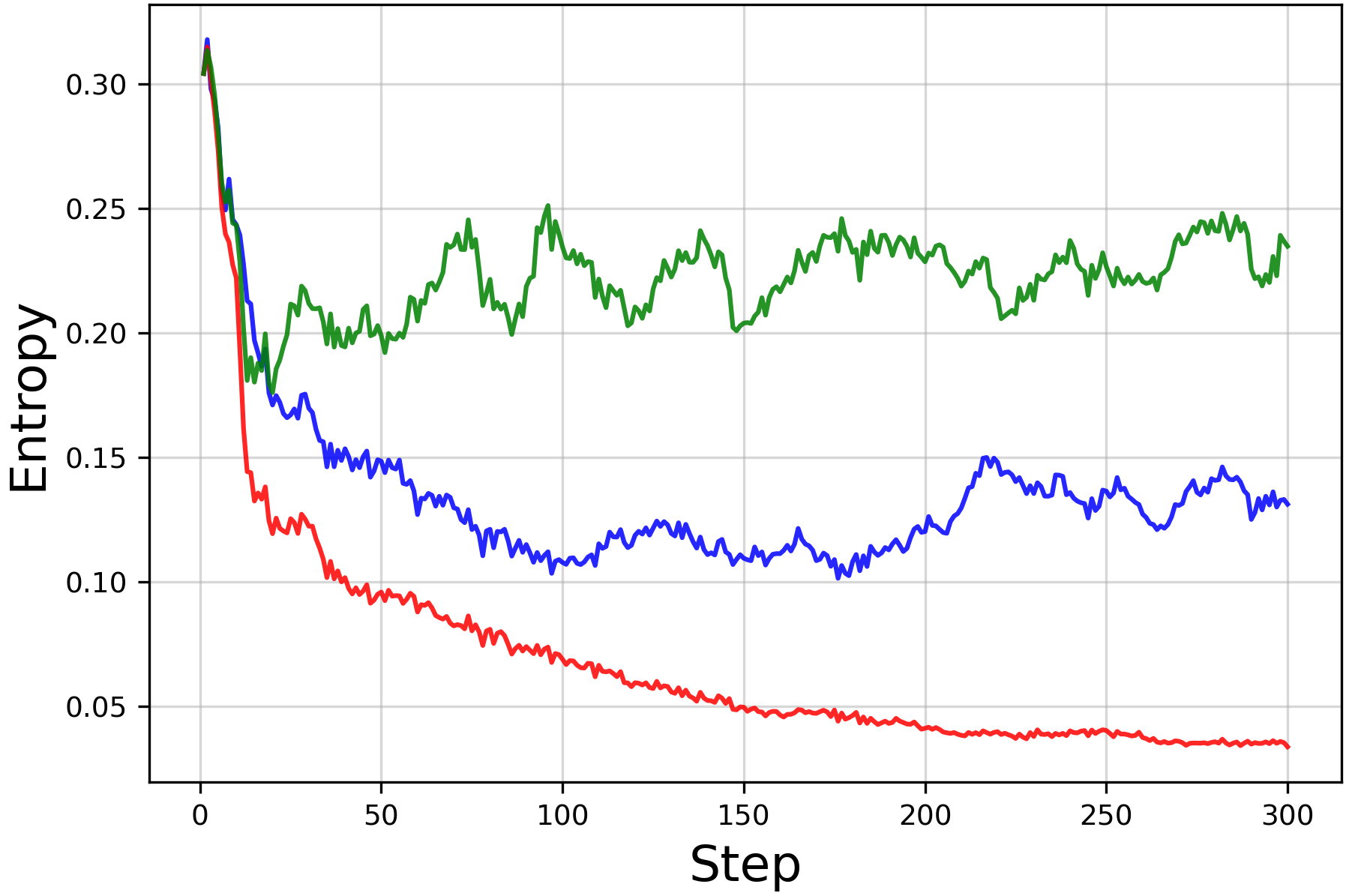}
  \end{subfigure}
  \caption{Change of policy entropy during RL training with \textbf{(Left)} random rewards on the \texttt{Qwen2.5-1.5B-Instruct} model and \textbf{(Right)} nonrandom rewards on the \texttt{Qwen2.5-3B-Instruct} model with KL divergence loss term during training.
  }
  \label{fig:entropy_ablation_kl}
\end{figure*}

We reproduce the results in Figure~\ref{fig:entropy_ablation} with KL divergence term on during training. We use the standard hyperparameter choice of $\beta = 0.001$, with all other configurations being identical. 
As shown in Figure~\ref{fig:entropy_ablation_kl}, we observe that the directional bias of clipping is still clear even with KL divergence term being present.

\section{ Random reward training across different settings }
\label{appendix:random_reward_ablation}

\begin{figure}[hbtp]
  \centering
    \includegraphics[width= 0.6 \linewidth]{\arxivpath 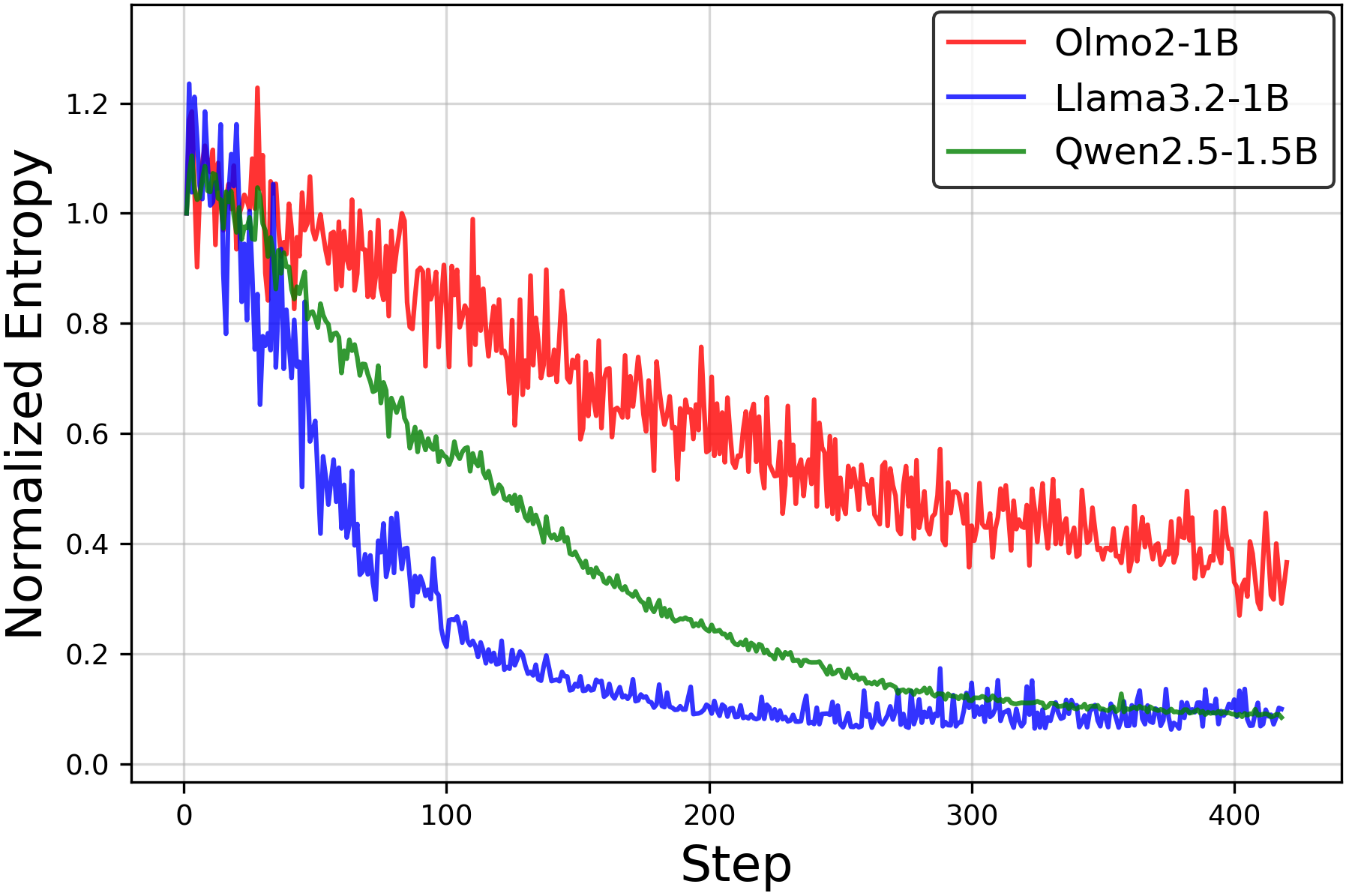}
  \caption{ Entropy change across different model families when trained with random rewards under symmetric clipping $\varepsilon_{\mathrm{low}} = \varepsilon_{\mathrm{high}}$. The entropy is normalized with the corresponding base model's entropy.
  This result substantiates that the clipping bias is not a model-specific phenomenon.  
  }
  \label{fig:random_reward_model_ablation}
\end{figure}

To corroborate that the entropy minimization effect of random rewards with symmetric clipping $\varepsilon_{\mathrm{low}} = \varepsilon_{\mathrm{high}}$ is a model-agnostic result, we conduct the same experiment with three base models from different model families. In Figure~\ref{fig:random_reward_model_ablation}, we present the normalized entropy of models \texttt{Qwen2.5-1.5B-Instruct}, \texttt{Llama3.2-1B-Instruct}, and \texttt{OLMo-2-0425-1B-Instruct} during RL training.
We normalize the entropy of each model by the entropy of the base model. 
Due to slow convergence, we set $\varepsilon_{\mathrm{high}}=\varepsilon_{\mathrm{low}}=0.1$ for \texttt{Olmo2}, and $\varepsilon_{\mathrm{high}}=\varepsilon_{\mathrm{low}}=0.2$ for other models. One can clearly observe a decreasing trend for all three models.

\begin{figure}[hbtp]
  \centering
    \includegraphics[width=0.6\linewidth]{\arxivpath 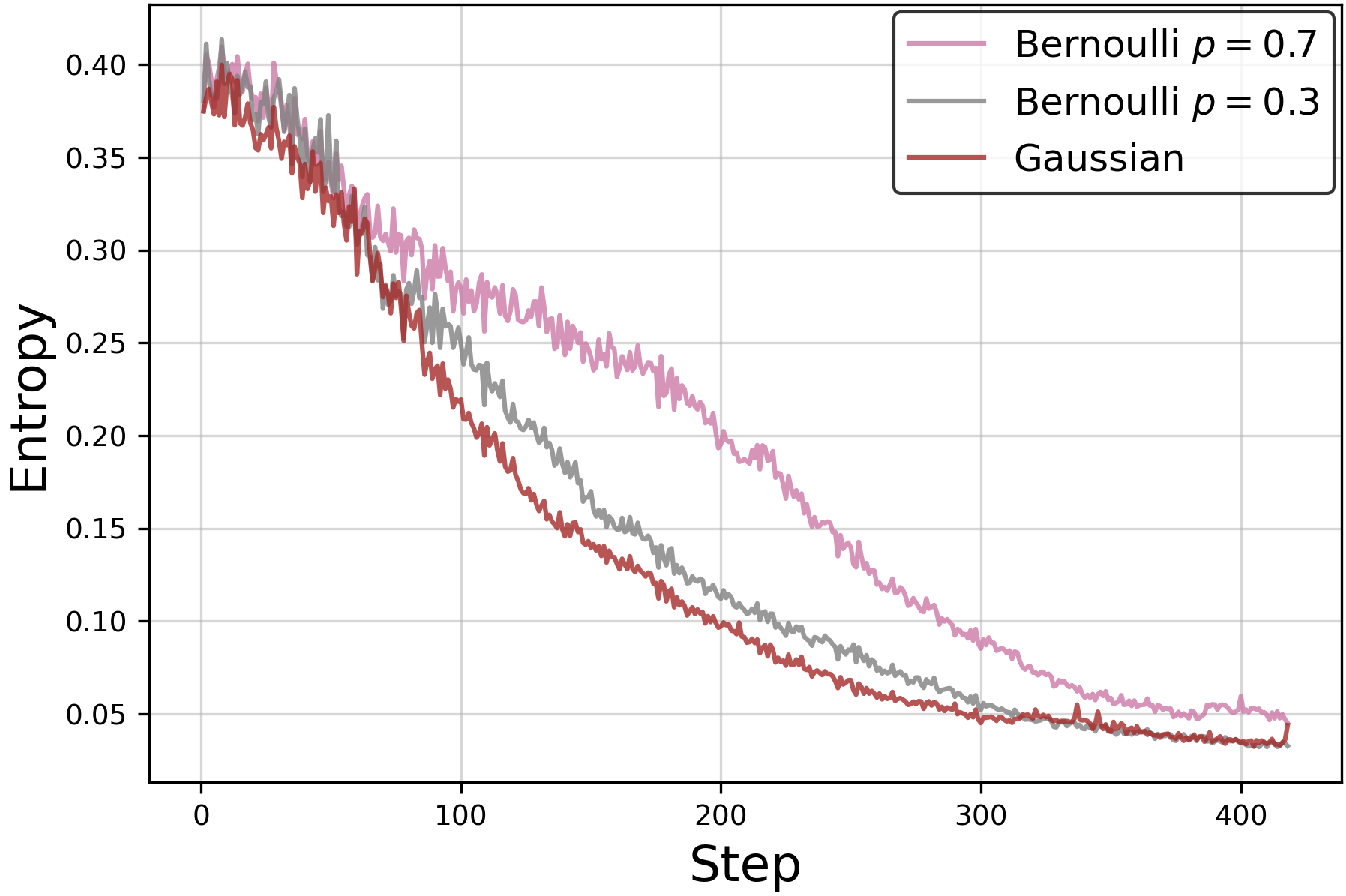}
  \caption{Entropy change of \texttt{Qwen2.5-1.5B-Instruct} model with random rewards sampled from various probability distributions.
  }
  \label{fig:random_reward_shape_ablation}
\end{figure}

Further, we use different random sources for the rewards for RL training of \texttt{Qwen2.5-1.5B-Instruct} model. We test three random sources from which we sample the rewards: Bernoulli random reward with $p=0.3$ (`Bernoulli $p=0.3$) and $p=0.7$ (`Bernoulli $p=0.7$) where reward $1$ is given for probability $p$ and $0$ for probability $1-p$, and standard normal distribution (`Gaussian') so that $r \sim \mathcal{N}(0,1)$. As in other experiments, we use the GRPO algorithm with group size $8$. 
In Figure~\ref{fig:random_reward_shape_ablation}, one can conclude that entropy minimization is implicitly performed during the RL training, regardless of the distribution from which the reward is sampled.

\section{ Clipping hyperparameter search experiment details } \label{appendix:clip_param_search}
\begin{figure}[hbtp]
  \centering
  {
    \begin{subfigure}[t]{\textwidth}
      \begin{minipage}[t]{0.32\textwidth}
        \centering
        \includegraphics[width=\linewidth]{\arxivpath 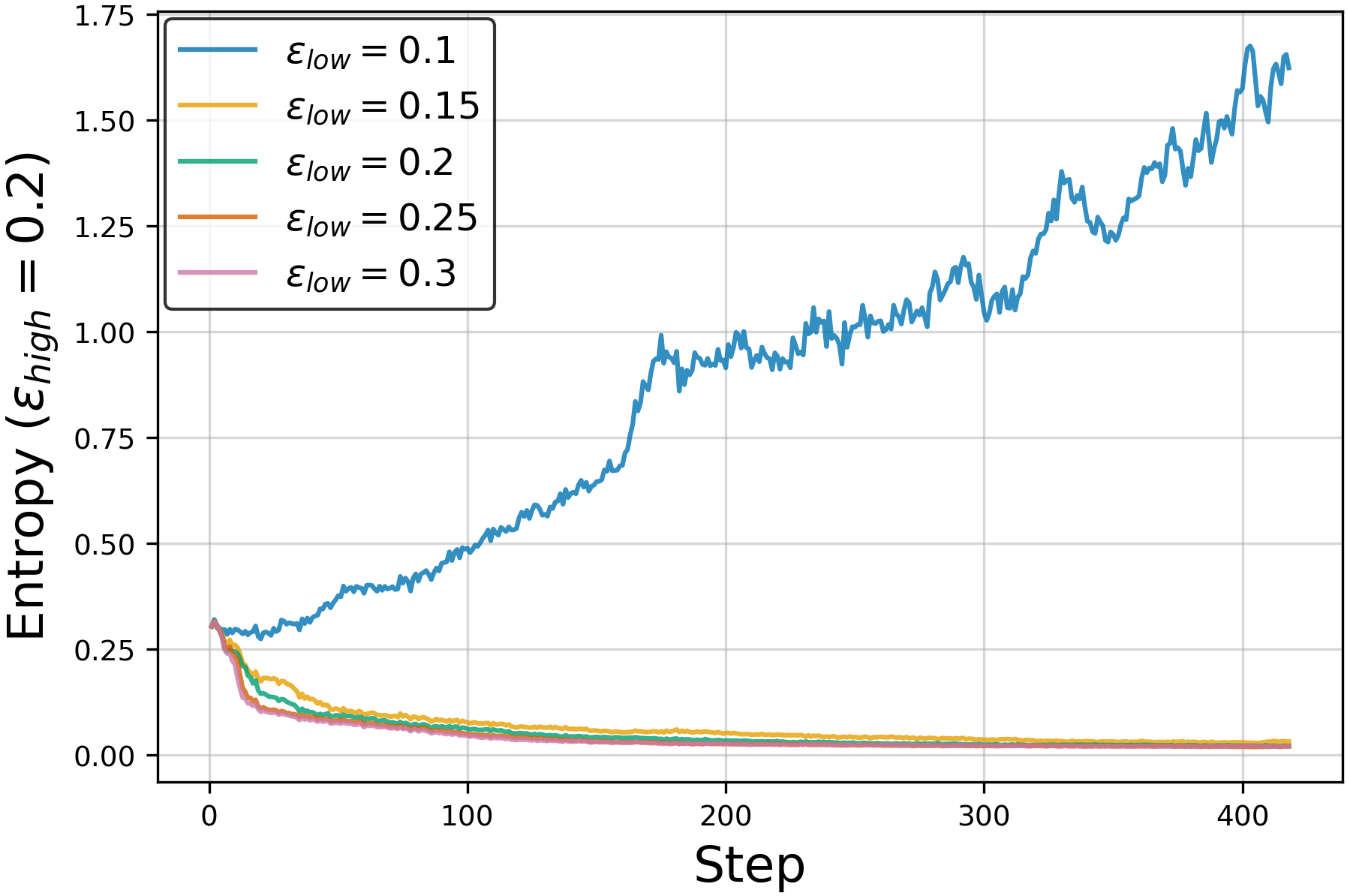}
      \end{minipage}\hfill
      \begin{minipage}[t]{0.32\textwidth}
        \centering
        \includegraphics[width=\linewidth]{\arxivpath 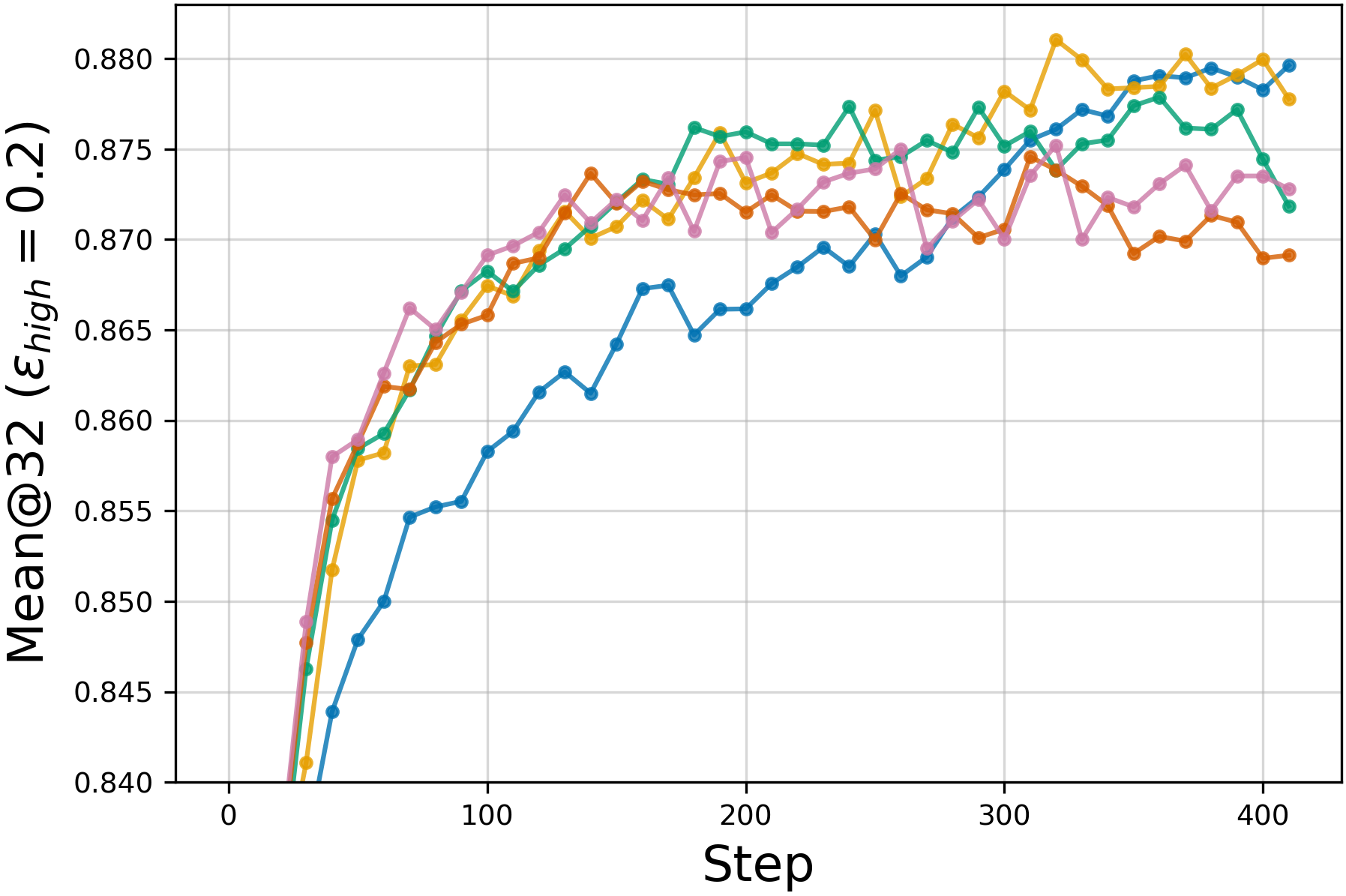}
      \end{minipage}\hfill
      \begin{minipage}[t]{0.32\textwidth}
        \centering
        \includegraphics[width=\linewidth]{\arxivpath 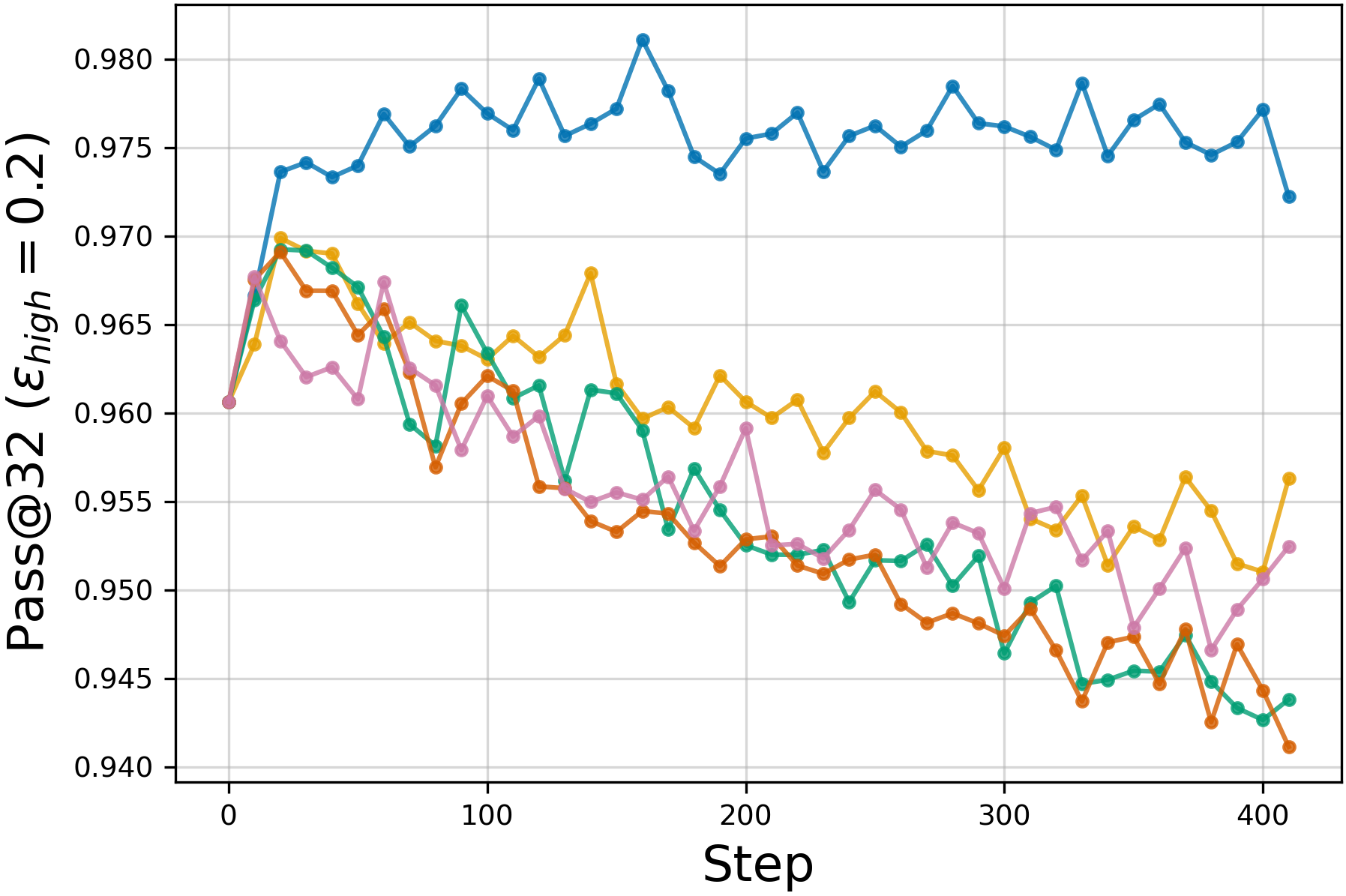}
      \end{minipage}
    \end{subfigure}%
  }
  {
    \begin{subfigure}[t]{\textwidth}
      \begin{minipage}[t]{0.32\textwidth}
        \centering
        \includegraphics[width=\linewidth]{\arxivpath 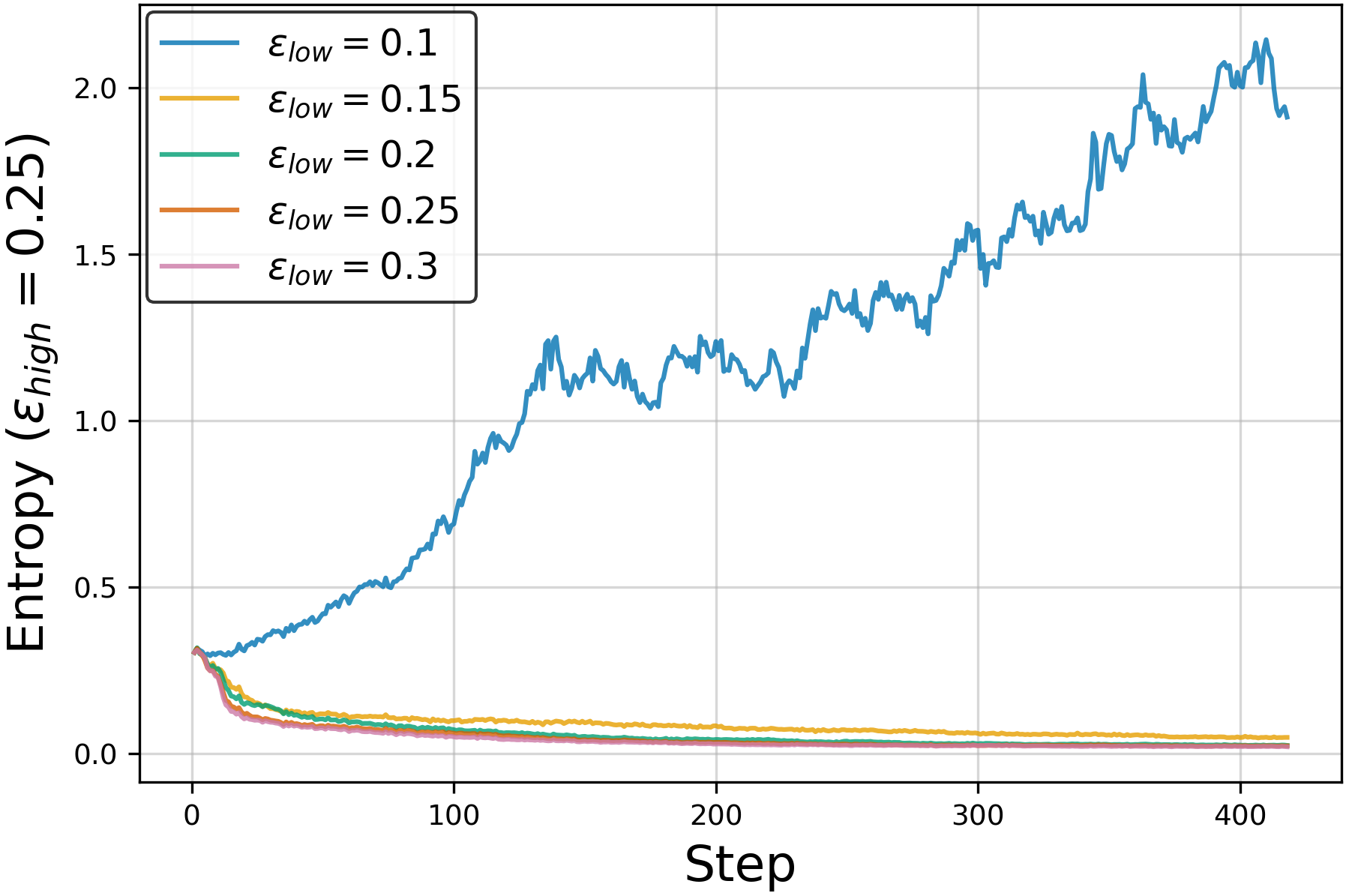}
      \end{minipage}\hfill
      \begin{minipage}[t]{0.32\textwidth}
        \centering
        \includegraphics[width=\linewidth]{\arxivpath 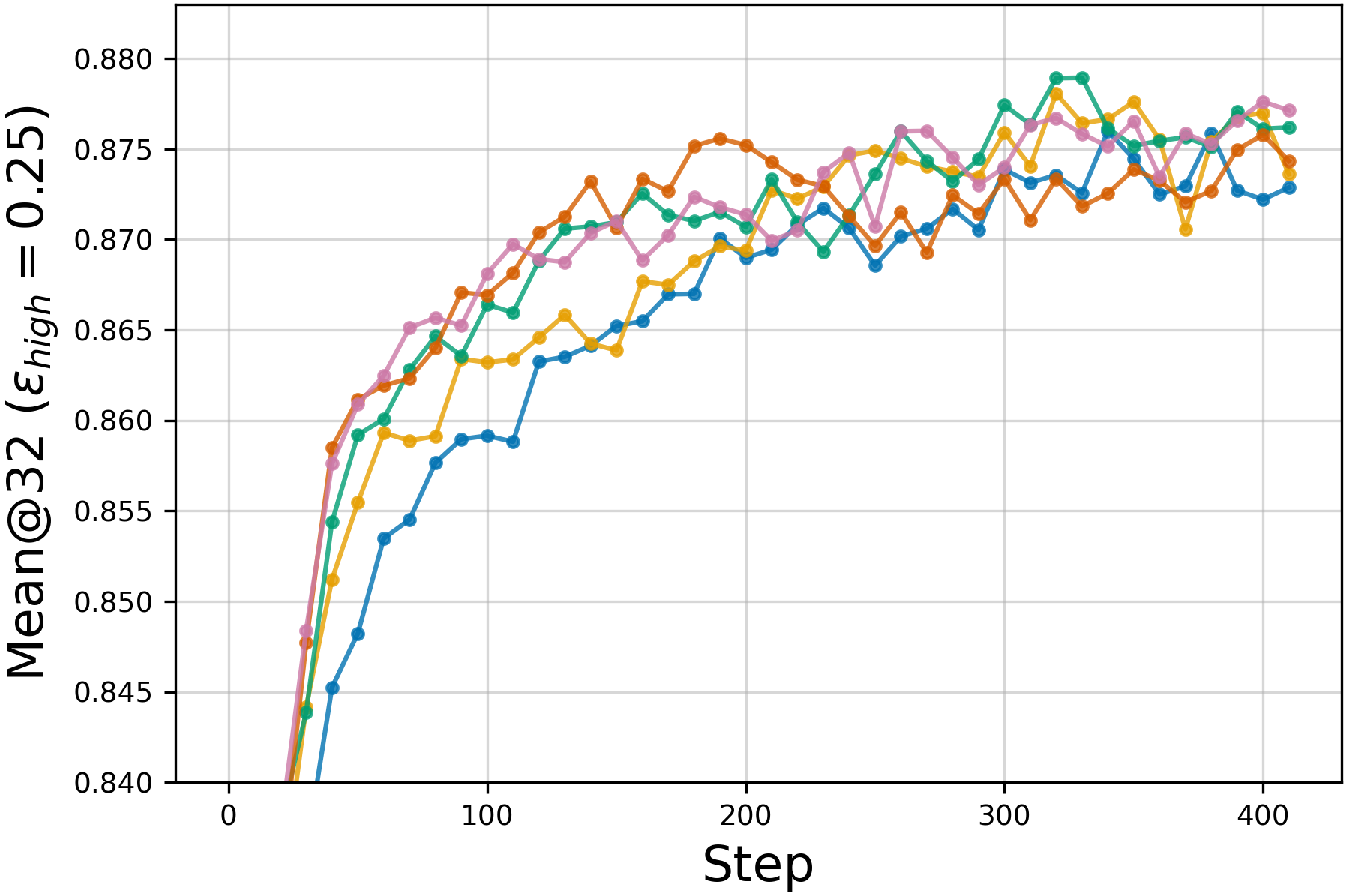}
      \end{minipage}\hfill
      \begin{minipage}[t]{0.32\textwidth}
        \centering
        \includegraphics[width=\linewidth]{\arxivpath 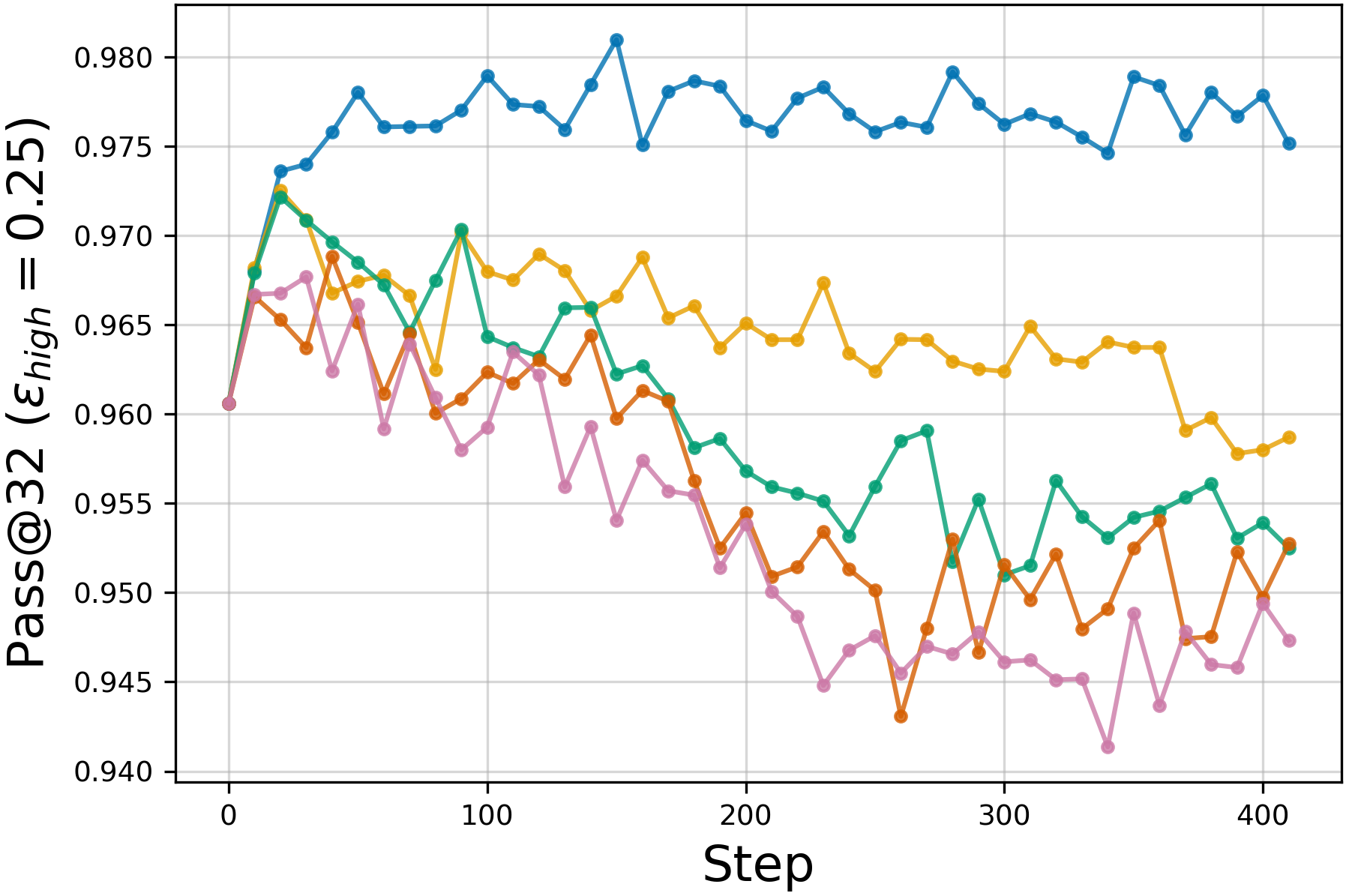}
      \end{minipage}
    \end{subfigure}%
  }
  {
    \begin{subfigure}[t]{\textwidth}
      \begin{minipage}[t]{0.32\textwidth}
        \centering
        \includegraphics[width=\linewidth]{\arxivpath 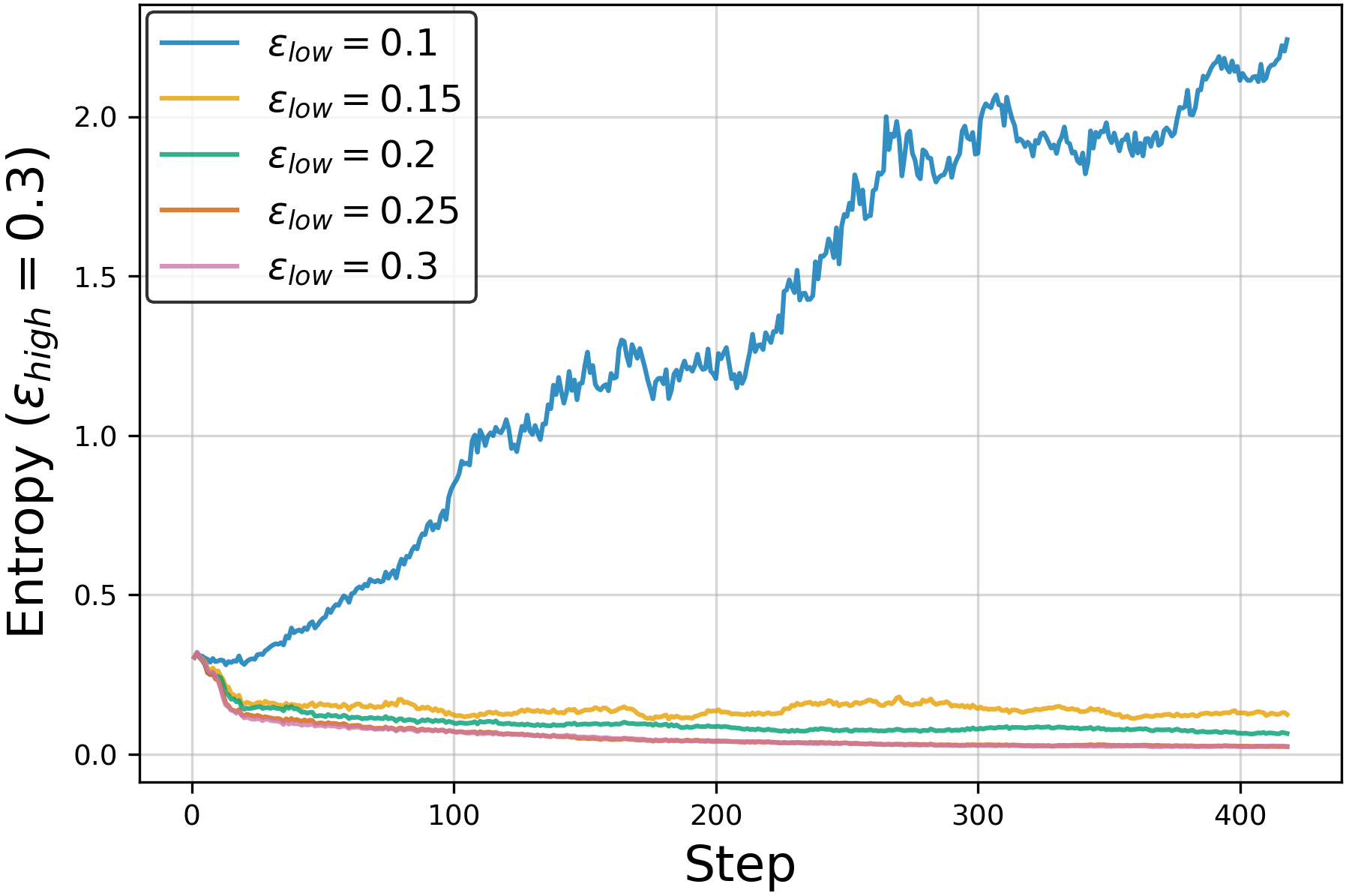}
      \end{minipage}\hfill
      \begin{minipage}[t]{0.32\textwidth}
        \centering
        \includegraphics[width=\linewidth]{\arxivpath 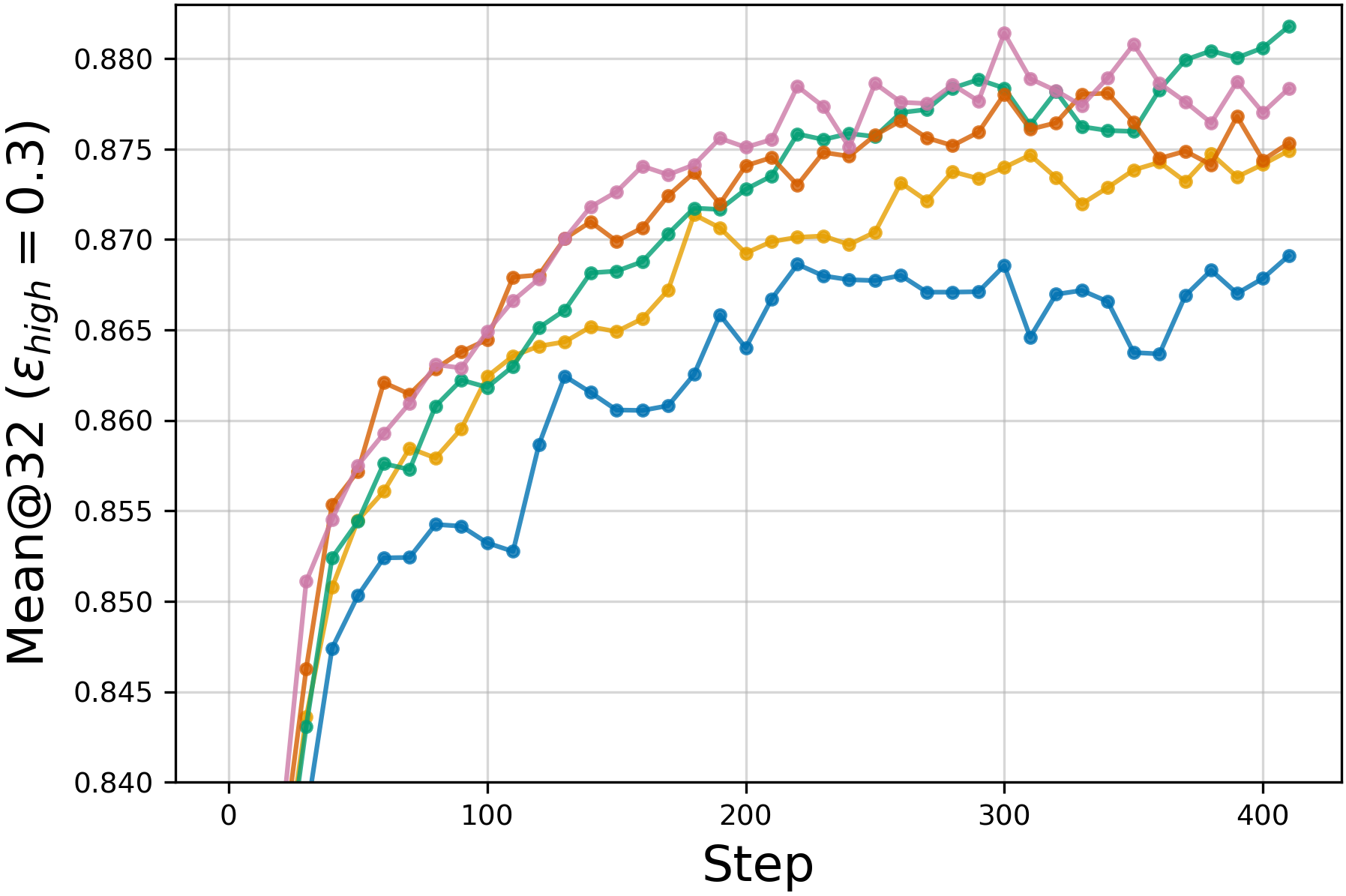}
      \end{minipage}\hfill
      \begin{minipage}[t]{0.32\textwidth}
        \centering
        \includegraphics[width=\linewidth]{\arxivpath 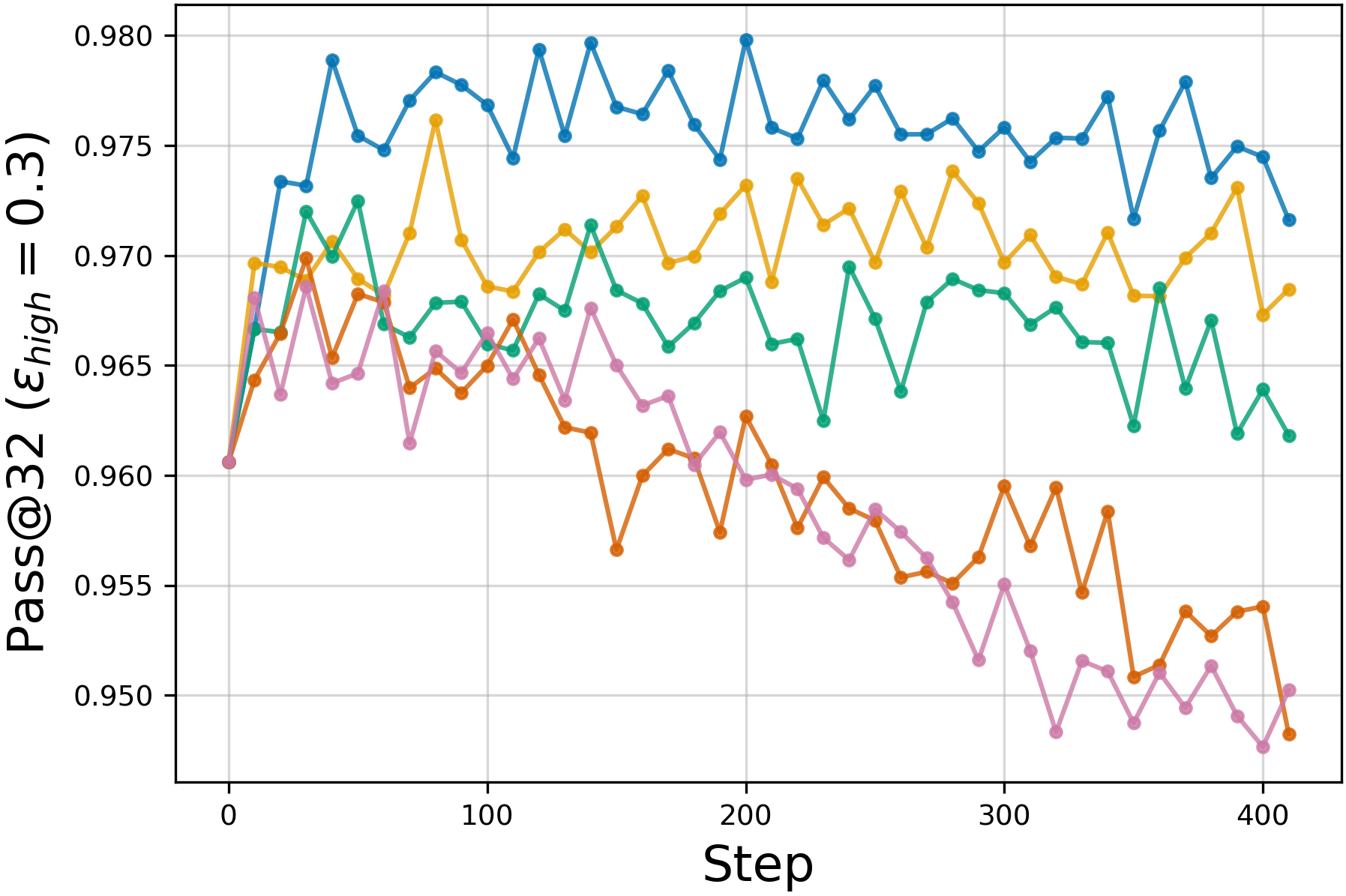}
      \end{minipage}
    \end{subfigure}%
  }
  \caption{Plotting the 
  entropy, 
  \texttt{mean@32}, and \texttt{pass@32} metric throughout the RLVR training for the
  \texttt{Qwen2.5-3B-Instruct} model with various $\epshigh$ and $\epslow$ parameters. The \texttt{mean@32} and \texttt{pass@32} metrics were evaluated on the \texttt{GSM8K} test dataset. 
  We observe that the clipping hyperparameters continuously control the entropy, while the \texttt{pass@32} shows high correlation with it. 
  }
  \label{fig:epsilon_grid_full} 
\end{figure}

In this section, we provide further detail regarding the epsilon hyperparameter sweep experiments. As stated in Section~\ref{subsec:experiments}, 
we used \texttt{Qwen2.5-3B-Instruct} as the base model, while varying $\epshigh \in \{0.2, 0.25, 0.3 \}$ and $\epslow \in \{0.1, 0.15, 0.2, 0.25, 0.3 \}$. 
Here, we provide the full experiment result of the epsilon hyperparameter experiment, which was summarized in Figure~\ref{fig:epsilon_grid}.
Figure~\ref{fig:epsilon_grid_full} shows the \texttt{entropy}, \texttt{mean@32} metric, and \texttt{pass@32} metric validated on the \texttt{GSM8K} dataset with \texttt{Qwen2.5-3B-Instruct} model with various clipping hyperparameter settings.
We plot the $5$ $\epslow$ configuration in one plot, where in each row the $\epshigh$ value is fixed. 
As suggested in Figure~\ref{fig:epsilon_grid}, the two clipping hyperparameters work as a lightweight entropy control knob. 
Interestingly, the entropy explodes when $\epslow = 0.1$ while the $\epslow = 0.15$ does not, for all three $\epshigh$ configurations. 
This might suggest there exists two regimes (stable and unstable) for entropy dynamics during RL training, which could be portrayed as a 2D phase diagram depending on $(\epshigh, \epslow)$.

\section{Full experiment results for clipping ablation study on various math benchmarks } \label{appendix:dapo_experiments}


In this section, we provide further details for the clipping ablation experiments in Table~\ref{tab:clip_ablations_math_benchmarks} for the \texttt{Qwen2.5-7B-Instruct} model trained with the \texttt{DAPO-Math-17k} dataset.
We use three independent matched training runs. Within each run, we average the \texttt{mean@32} and \texttt{pass@32} scores over the last three checkpoints shared by all runs and configurations, at steps 190, 200, and 210. We then report the mean and sample standard deviation across the three runs.

\begin{figure}[hbtp]
  \centering
  {
    \captionsetup[subfigure]{labelformat=empty,justification=centering,singlelinecheck=false}%
    \begin{subfigure}[t]{\textwidth}
      \begin{minipage}[t]{0.46\textwidth}
        \centering
        \includegraphics[width=\linewidth]{\arxivpath 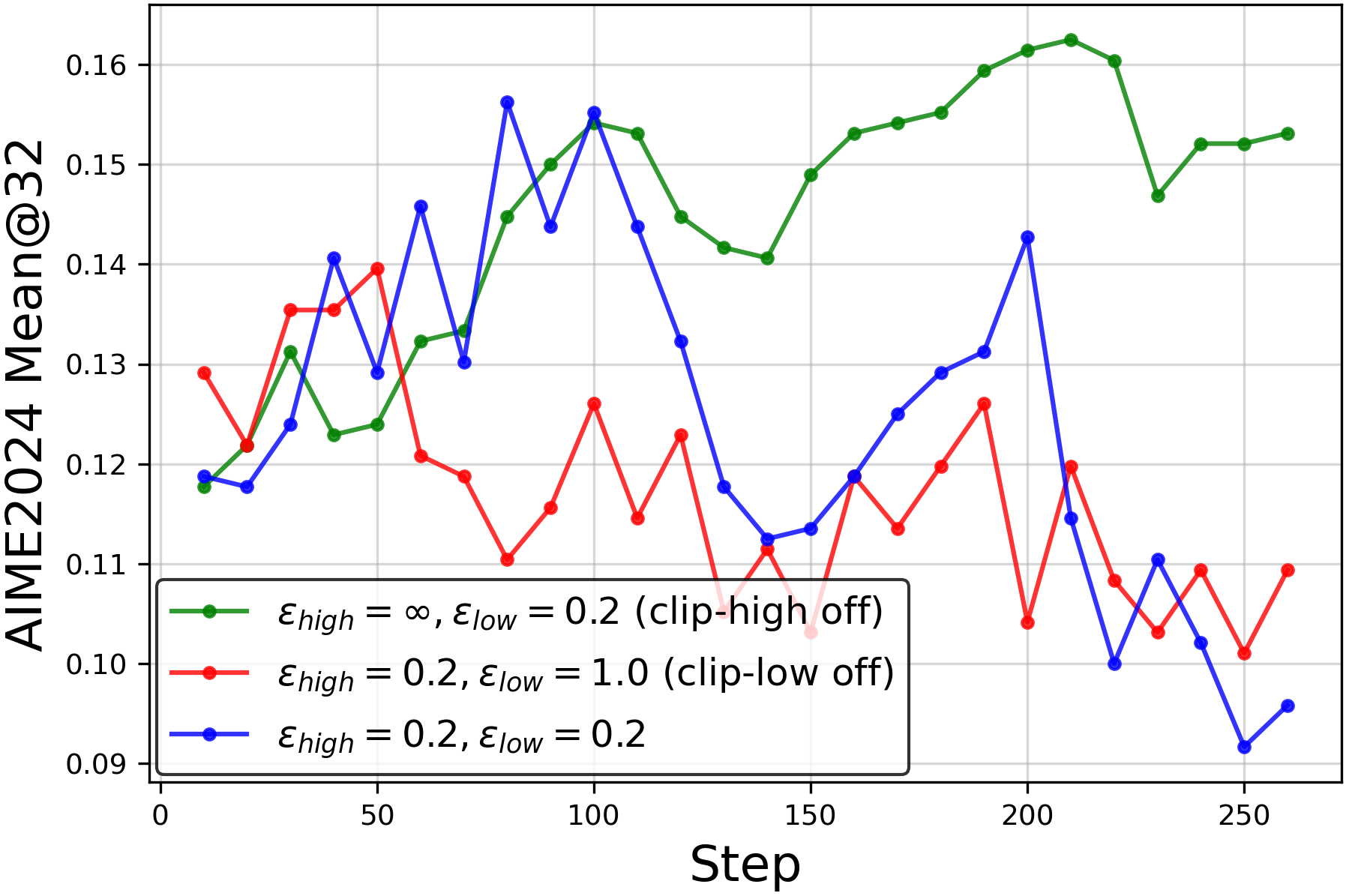}
      \end{minipage}\hfill
      \begin{minipage}[t]{0.46\textwidth}
        \centering
        \includegraphics[width=\linewidth]{\arxivpath 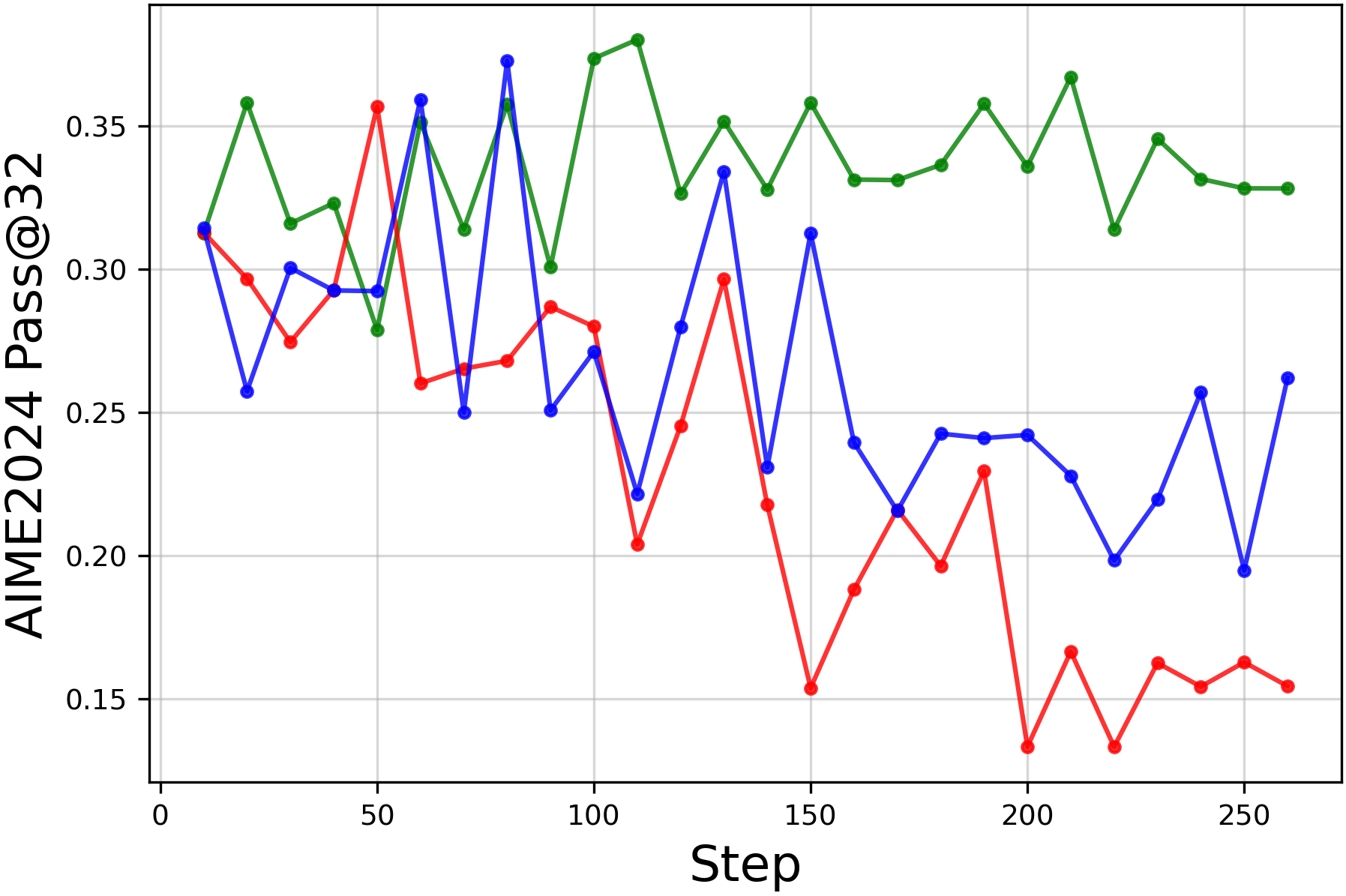}
      \end{minipage}
    \end{subfigure}%
  }
  {
    \captionsetup[subfigure]{labelformat=empty,justification=centering,singlelinecheck=false}%
    \begin{subfigure}[t]{\textwidth}
      \begin{minipage}[t]{0.46\textwidth}
        \centering
        \includegraphics[width=\linewidth]{\arxivpath 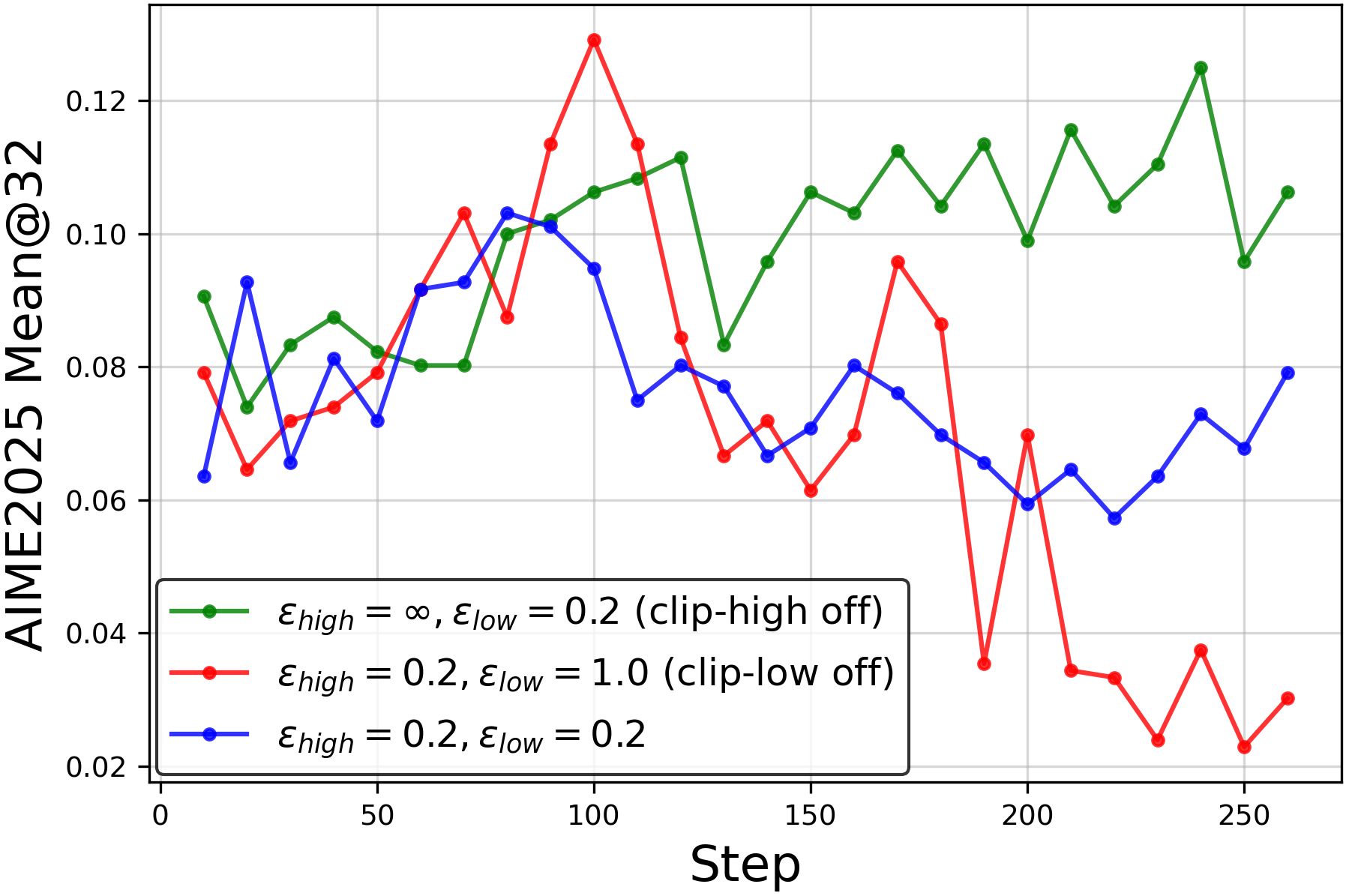}
      \end{minipage}\hfill
      \begin{minipage}[t]{0.46\textwidth}
        \centering
        \includegraphics[width=\linewidth]{\arxivpath 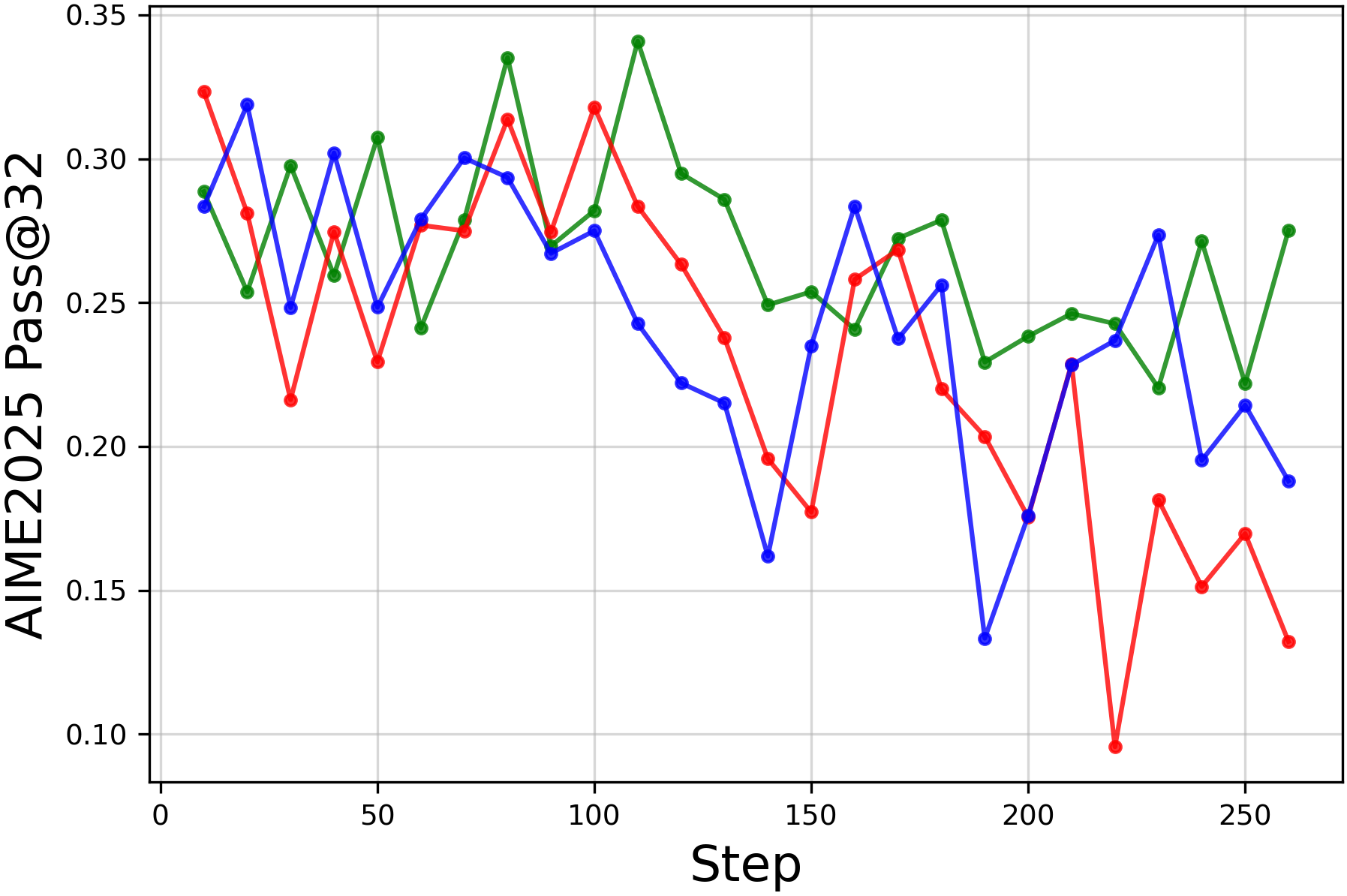}
      \end{minipage}
    \end{subfigure}
}
  {
    \captionsetup[subfigure]{labelformat=empty,justification=centering,singlelinecheck=false}%
    \begin{subfigure}[t]{\textwidth}
      \begin{minipage}[t]{0.46\textwidth}
        \centering
        \includegraphics[width=\linewidth]{\arxivpath 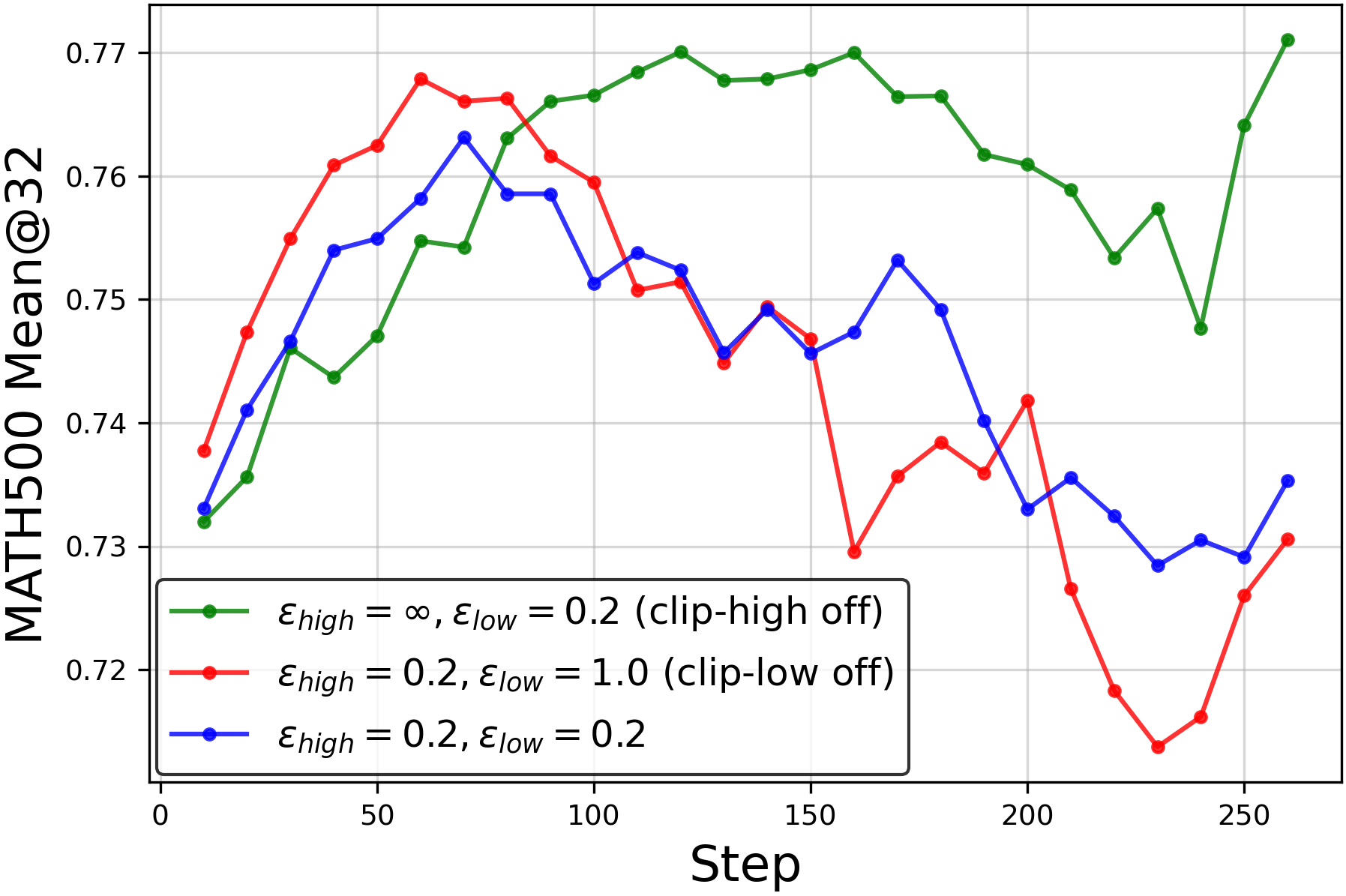}
      \end{minipage}\hfill
      \begin{minipage}[t]{0.46\textwidth}
        \centering
        \includegraphics[width=\linewidth]{\arxivpath 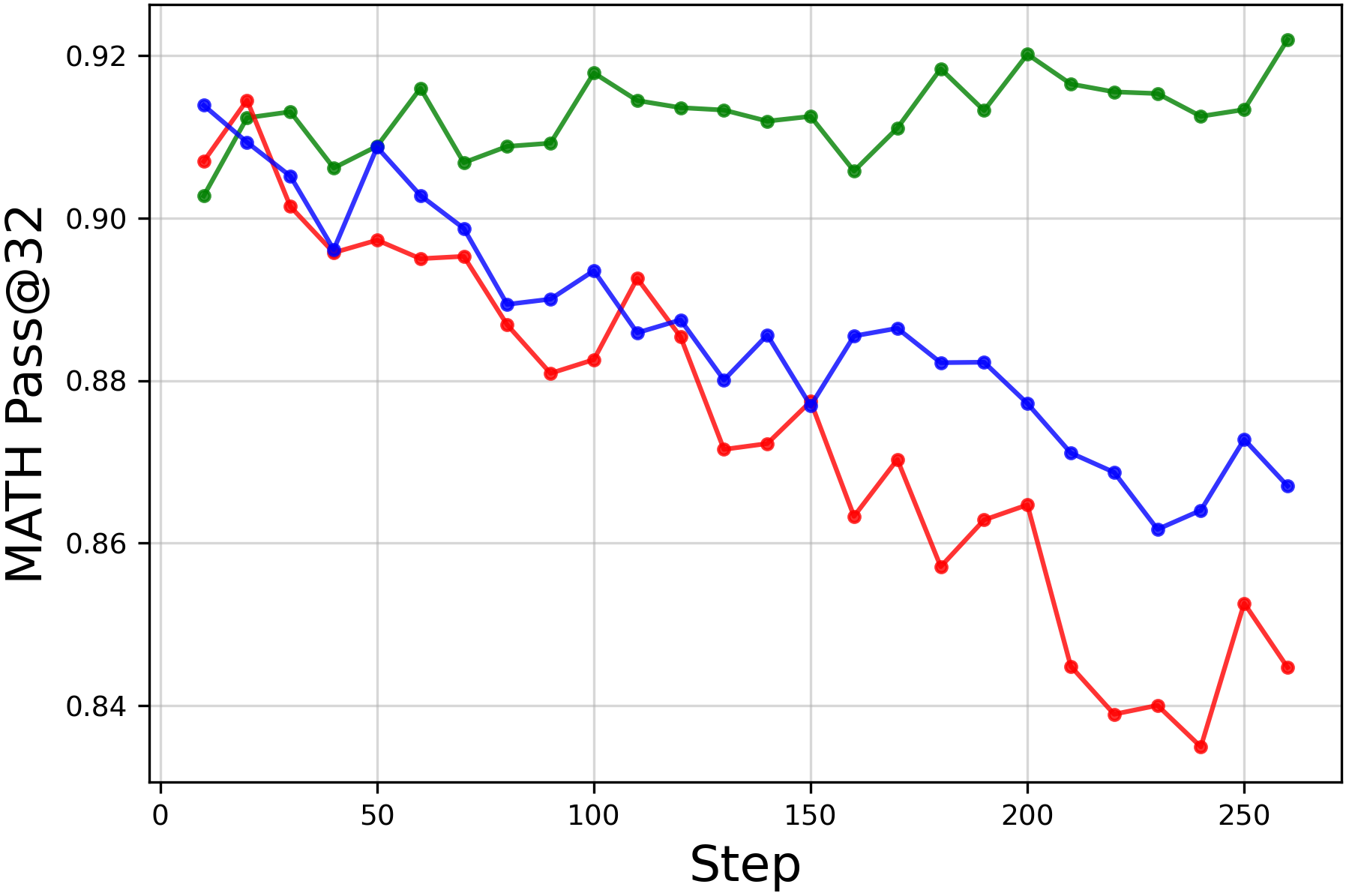}
      \end{minipage}
    \end{subfigure}%
  }
  {
    \captionsetup[subfigure]{labelformat=empty,justification=centering,singlelinecheck=false}%
    \begin{subfigure}[t]{\textwidth}
      \begin{minipage}[t]{0.46\textwidth}
        \centering
        \includegraphics[width=\linewidth]{\arxivpath 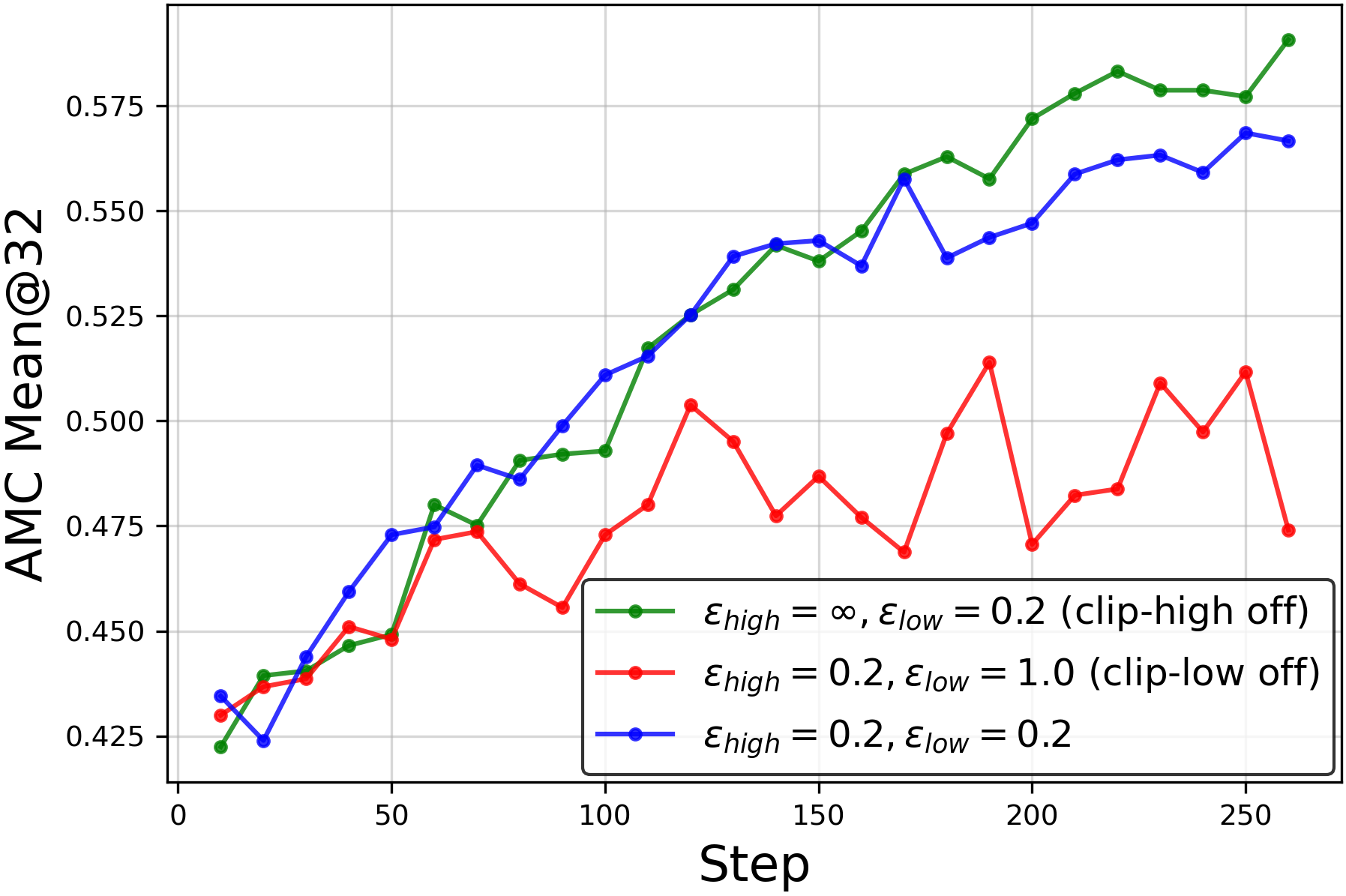}
      \end{minipage}\hfill
      \begin{minipage}[t]{0.46\textwidth}
        \centering
        \includegraphics[width=\linewidth]{\arxivpath 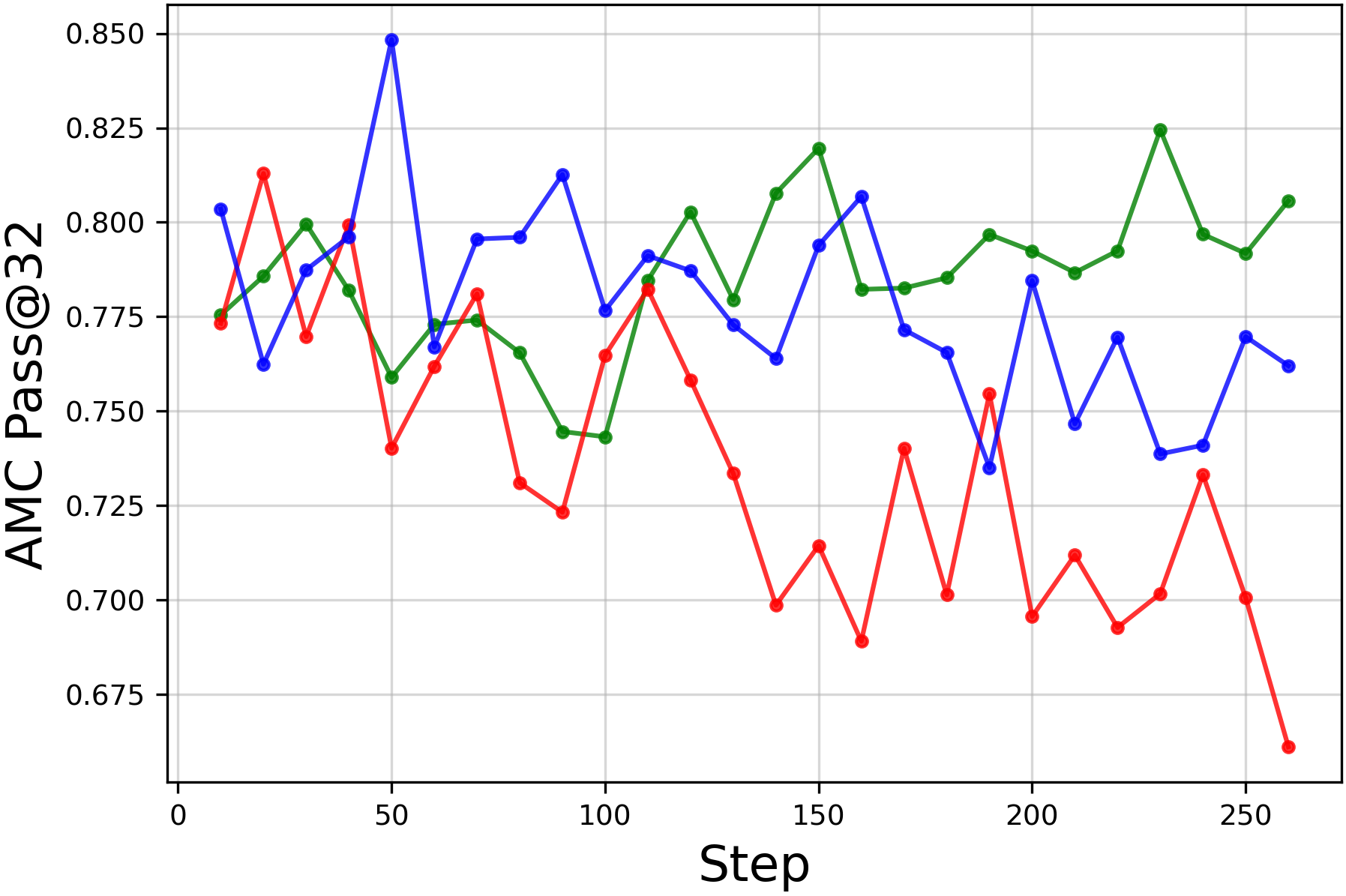}
      \end{minipage}
    \end{subfigure}
}

  \caption{Performance measured by the
  \texttt{mean@32} metric \textbf{(Left)} and \texttt{pass@32} metric \textbf{(Right)} metric during RLVR for the
  \texttt{Qwen2.5-7B-Instruct} model trained with \texttt{DAPO-Math-17k} dataset, evaluated on the \texttt{AIME2024}, \texttt{AIME2025}, \texttt{AMC}, and \texttt{MATH-500} datasets.
  }
  \label{fig:dapo_nonrandom_clipablation_performance} 
\end{figure}

Figure~\ref{fig:dapo_nonrandom_clipablation_performance} shows individual-run learning curves for the \texttt{mean@32} and \texttt{pass@32} metrics throughout RL training. These curves illustrate the training dynamics; the three-run summary in Table~\ref{tab:clip_ablations_math_benchmarks} uses the common-checkpoint protocol described above.
Throughout various reasoning benchmarks, we observe that the \texttt{pass@k} metric has a decreasing trend, but deliberate clipping for increased policy entropy effectively hinders exploration degradation throughout training. The three-run results also show that removing clip-high improves the average \texttt{pass@32} score while keeping the average \texttt{mean@32} score comparable to the symmetric clipping baseline.

\section{Additional experiments for \texttt{Llama} base models}\label{appendix:subsec:llama_reproduction}

\begin{figure}
  \centering

  {%
    \captionsetup[subfigure]{labelformat=empty,justification=raggedright,singlelinecheck=false}%
    \begin{subfigure}[t]{\textwidth}
      \subcaption*{(a)}
      \begin{minipage}[t]{0.46\textwidth}
        \centering
        \includegraphics[width=\linewidth]{\arxivpath 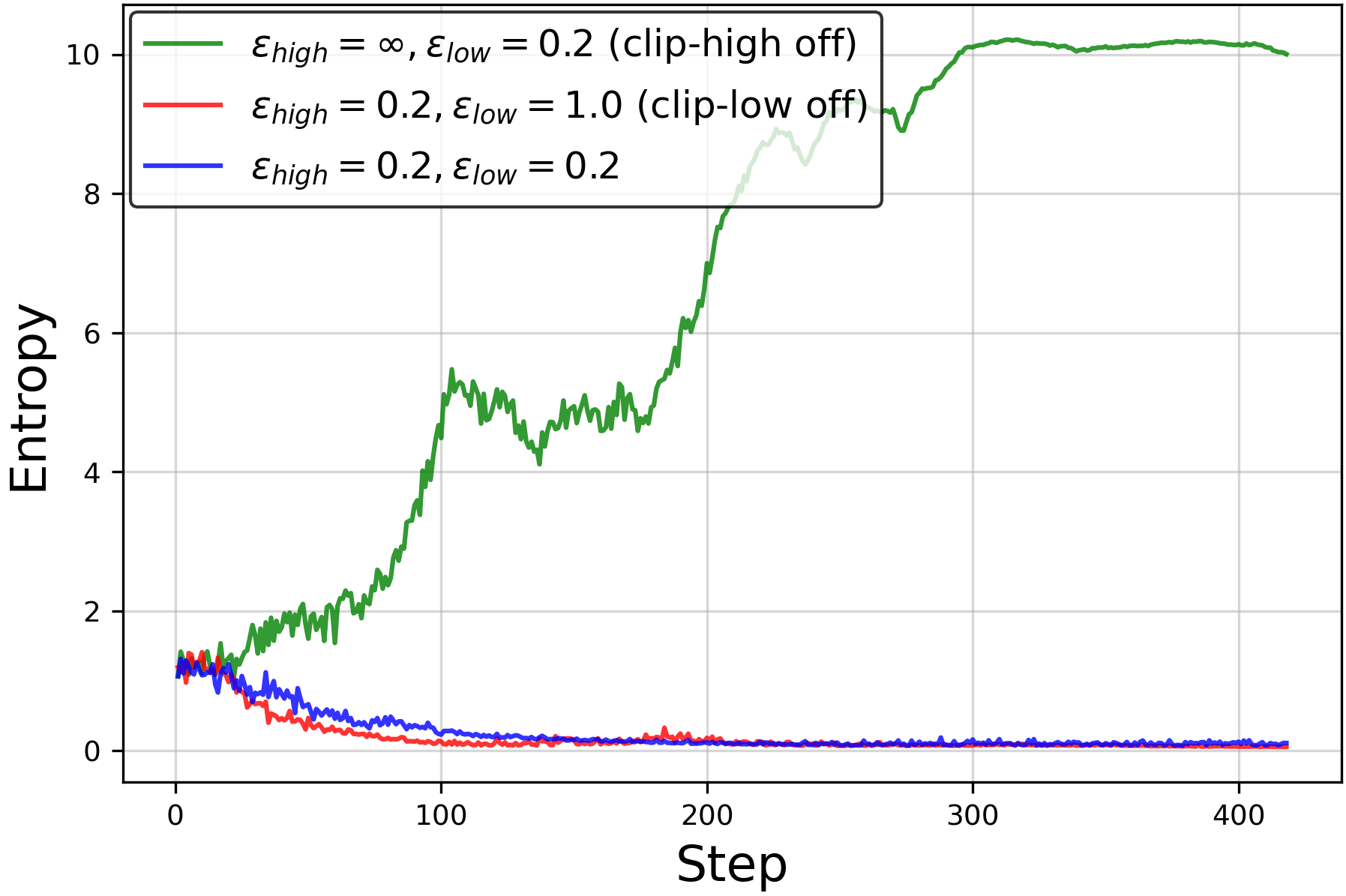}
      \end{minipage}\hfill
      \begin{minipage}[t]{0.46\textwidth}
        \centering
        \includegraphics[width=\linewidth]{\arxivpath 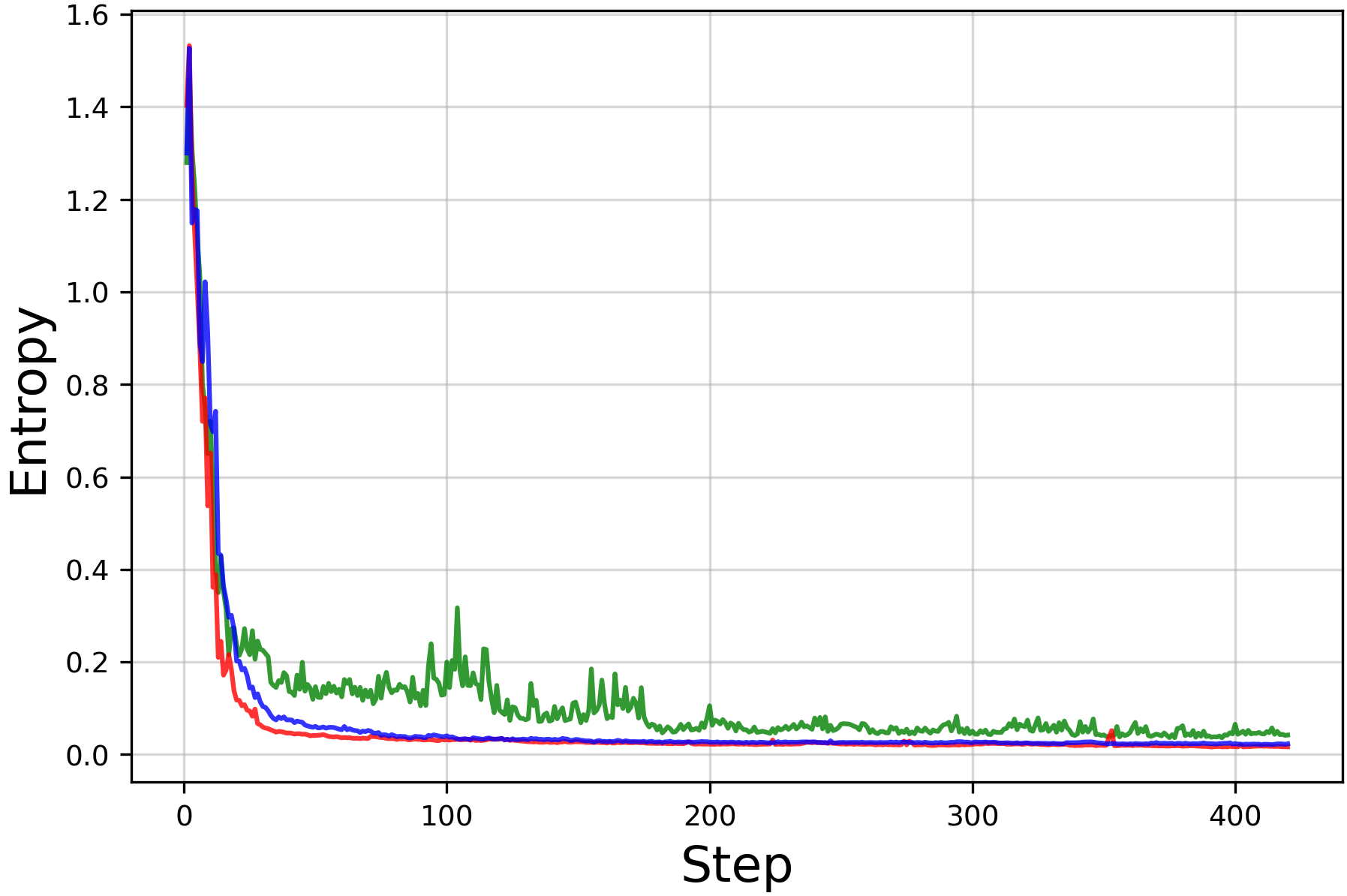}
      \end{minipage}
    \end{subfigure}%
  }


  {%
    \captionsetup[subfigure]{labelformat=empty,justification=raggedright,singlelinecheck=false}%
    \begin{subfigure}[t]{\textwidth}
      \subcaption*{(b)}
      \begin{minipage}[t]{0.46\textwidth}
        \centering
        \includegraphics[width=\linewidth]{\arxivpath 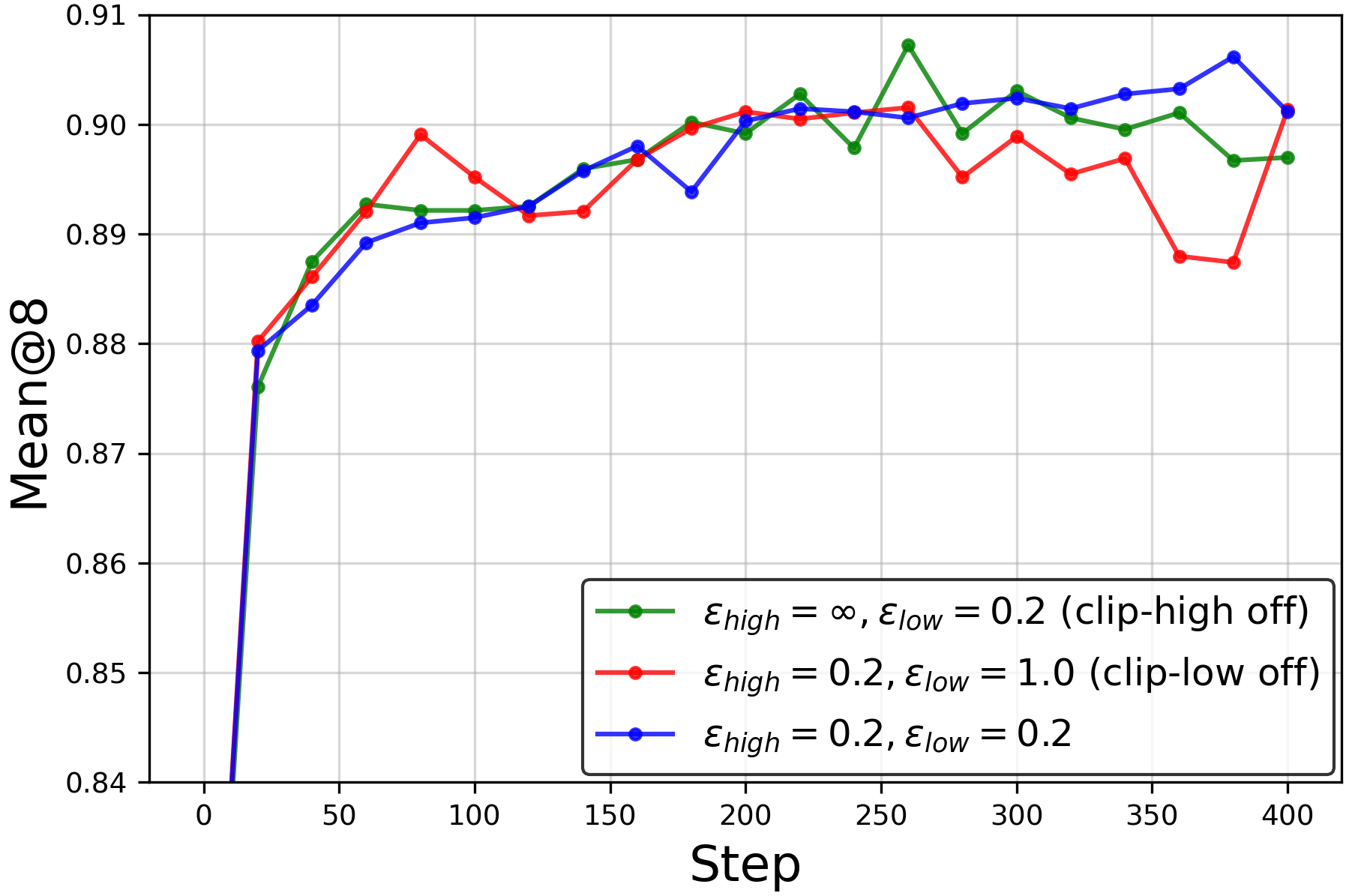}
      \end{minipage}
      \hfill
      \begin{minipage}[t]{0.46\textwidth}
        \centering
        \includegraphics[width=\linewidth]{\arxivpath 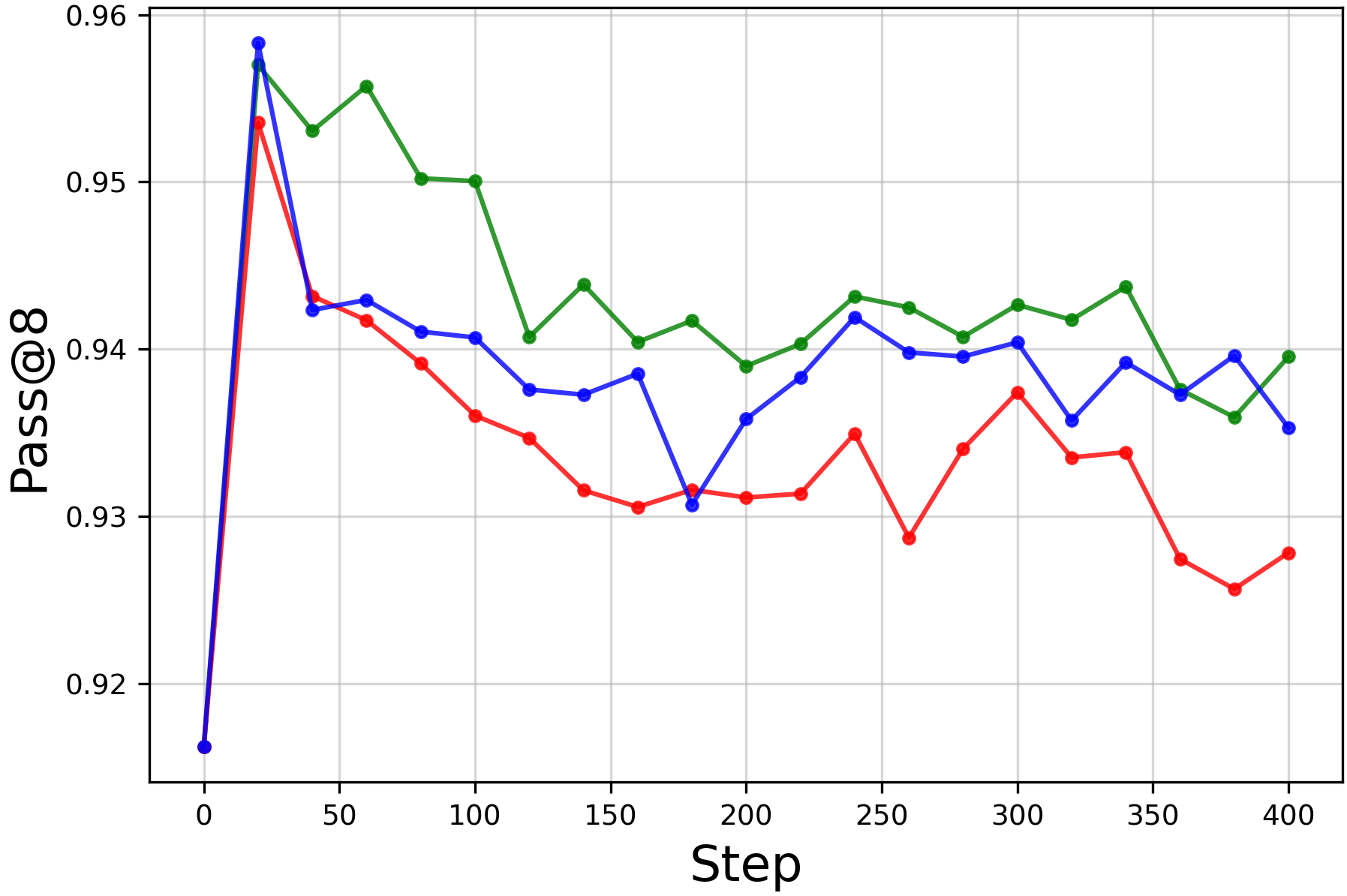}
      \end{minipage}
    \end{subfigure}%
  }
  \caption{Clipping ablation results reproduced with \texttt{Llama} base models. \textbf{(a)} We plot the policy entropy change during RL training trained with \textbf{(Left)} random rewards and \textbf{(Right)} true rewards. We observe that the clipping bias has the same qualitative trend for \texttt{Llama}-family base models as well. \textbf{(b)} We measure the \textbf{(Left)} \texttt{mean@8} metric and \textbf{(Right)} \texttt{pass@8} metric throughout the training. 
  We note that the high-entropy settings improve exploration while preserving the mean accuracy.  
  }
  \label{fig:llama_experiments}
\end{figure}

In this section, 
we reproduce the experimental results
in the paper with \texttt{Llama} base models, to confirm that the empirical results are not model family(\texttt{Qwen})-specific. 
For random rewards settings, we train the \texttt{Llama3.2-1B-Instruct} model, and for true rewards, we use the \texttt{Llama3-8B-Instruct} model. 
In Figure~\ref{fig:llama_experiments} (a), we conduct \textbf{(Left)} random reward experiments with base model \texttt{Llama3.2-1B-Instruct} and \textbf{(Right)} nonrandom reward experiments with base model \texttt{Llama3-8B-Instruct}. As in the case with \texttt{Qwen}-based models, we can clearly observe the opposite effects of upper and lower clip on policy entropy. Figure~\ref{fig:llama_experiments} (b) shows \textbf{(Left)} \texttt{mean@8} and \texttt{pass@8} metric evaluated on the \texttt{GSM8K} dataset with the \texttt{Llama3-8B-Instruct} model. Again, we confirm similar trends that were observed in the \texttt{Qwen} model family. 

\section{Matched Transformer--AdamW intervention}
\label{appendix:adamw_intervention}

We use \texttt{Qwen2.5-3B} checkpoints from three training configurations: the symmetric clipping baseline, clip-high off, and clip-low off. For each configuration, we use checkpoints at steps 120, 240, and 360, with two stored rollout batches per checkpoint. The resulting 18 matched combinations contain 1,152 trajectories and 396,687 valid response tokens. 

For PG and NPG, we evaluate the clip-low and clip-high contributions in Theorems~\ref{thm:entropy_change_clip_general} and~\ref{thm:entropy_change_clip2_general} using the stored advantages and full-vocabulary policy probabilities. 
For the Transformer--AdamW intervention, the first update is common to all four branches. At the start of this update, the current policy equals the rollout policy, so the importance ratio is one and clipping is inactive. We then copy the resulting model parameters, AdamW moments, scheduler state, training batch, minibatch order, and RNG state. Starting from these identical states, we perform the second update with neither clipping mechanism, clip-low only, clip-high only, or both. We evaluate the entropy of all four updated models on the same rollout prefixes.

To measure the clip-low effect with clip-high disabled, we compare the clip-low-only branch with the branch using neither mechanism. With clip-high enabled, we compare the both-clipped branch with the clip-high-only branch. The two clip-high effects are obtained analogously. Table~\ref{tab:adamw_intervention} reports the averaged effects: clip-low increases entropy in both comparisons, while clip-high decreases it in both comparisons.

\end{document}